\documentclass[journal=jcisd8,manuscript=article]{achemso}

\usepackage[final]{changes}
\usepackage{amsmath,graphicx}
\usepackage{amssymb}
\usepackage[T1]{fontenc}  
\usepackage[utf8]{inputenc}
\usepackage[english]{babel}
\usepackage{lipsum}
\usepackage{subfiles}
\usepackage{siunitx}
\usepackage{multirow}
\usepackage{caption}
\usepackage{subcaption}
\usepackage{standalone}

\usepackage{hyperref}

\usepackage{psfrag}

\title{Autoencoding Undirected Molecular Graphs With Neural Networks}

\author{Jeppe Johan Waarkjær Olsen}
\affiliation{Department of Computing, Technical University of Denmark}

\author{Peter Ebert Christensen}
\affiliation{Department of Computing, Technical University of Denmark}

\author{Martin Hangaard Hansen}

\author{Alexander Rosenberg Johansen}
\affiliation{Department of Computing, Technical University of Denmark}
\email{aler@dtu.dk}

\begin{document}
\maketitle
\begin{abstract}
\added{
Discrete structure rules for validating molecular structures are usually limited to fulfilment of the octet rule or similar simple deterministic heuristics.} We propose a model, inspired by language modeling from natural language processing, \replaced{with the ability to learn from a collection of undirected molecular graphs, enabling fitting of any underlying structure rule present in the collection.}{ which can automatically correct molecules in discrete representations using a structure rule learned from a collection of undirected molecular graphs}

We introduce an adaption \replaced{to the popular Transformer model}{ on a modern neural network architecture, the Transformer}, which can learn relationships between atoms and bonds.
\replaced{To our knowledge, the Transformer adaption is the first model that is trained to solve}{ The algorithm thereby
solves} the unsupervised task of recovering partially observed molecules\deleted{  represented as
undirected graphs}.
\added{
In this work, we assess how different degrees of information impacts performance w.r.t. to fitting the QM9 dataset, which conforms to the octet rule, and to fitting the ZINC dataset, which contains hypervalent molecules and ions requiring the model to learn a more complex structure rule.
More specifically, we test a full discrete graph with bond order information, full discrete graph with only connectivity, a \texttt{bag-of-neighbors}, a \texttt{bag-of-atoms}, and a count-based \texttt{unigram} statistics.} \deleted{ This is to our knowledge, the first work that can automatically learn
any discrete molecular structure rule with input exclusively consisting of a training set
of molecules. In this work the neural network successfully approximates the octet rule,
relations in hypervalent molecules and ions when trained on the ZINC and QM9 dataset}

These results provide encouraging evidence that neural networks \replaced{, even when only connectivity is available, can learn arbitrary molecular structure rules specific to a dataset, as the Transformer adaption surpasses a strong octet rule baseline on the ZINC dataset.}{ can learn advanced
molecular structure rules and dataset specific properties, as the transformer surpasses
a strong octet-rule baseline.}

\end{abstract}
\newpage

\section{Introduction}
\label{sec:intro}

In drug discovery,\cite{ertl2000fast,lo2018machine} catalysis,\cite{ulissi2017address,boes2019graph}\added{ and} combustion\added{\cite{van2010accurate,broadbelt2005lexicography}}\deleted{and other areas} the number of possible relevant molecules grows exponentially with size of the molecule or \deleted{size of the} reaction network.
Modeling and exploring large \replaced{datasets}{data sets} of molecules \deleted{therefore} benefit\added{s} from fast coarse grained methods to generate\deleted{, select}, filter\added{, consistency check, validate} and correct molecules \deleted{represented as data}.

\added{
Databases of molecular properties calculated by \textit{ab initio} methods rely on consistency between 3D structures and molecular graphs. This consistency should be highly reliable and avoid any erroneous identifications after structure relaxation. The reliability requirement benefits from several redundant methods, which can flag any possible inconsistencies.

The tasks of validating, correcting, completing, and generating molecules in discrete representations usually rely on a simple heuristic such as the octet rule as the fundamental structure rule to determine the validity of molecules.\cite{fink2005virtual,blum2009970,ruddigkeit2012enumeration,qm9,elton2019deep,li2018multi}
The octet rule is, however, not satisfactory to validate all synthesizable molecules due to the occurrence of hypervalent molecules, ions, and non-integer bond orders such as in aromatic bonds.
}

\deleted{A trade-off exists between accuracy and cost from accurate, but expensive, quantum chemical methods to highly scalable learning algorithms which provide coarse grained predictions on the target properties. In machine learning models for atomic structure data, an important distinction lies between models on continuous real-space representations, such as force fields, as opposed to models on discrete graph based representations, the latter of which is a coarse grained, but far more scalable representation.\cite{hansen2015machine,elton2019deep}}

\added{Machine learning methods working on discrete molecular graphs can work as structure rules, learned from molecular datasets. At the same time, they can fit a great complexity of underlying trends, while being low cost compared to 3D representations and quantum chemical calculations.} Machine learning \deleted{methods} for predicting properties based on discrete \replaced{representations of molecules}{ encodings} have only recently begun to \replaced{use}{ train and predict on} undirected graphs\deleted{,} as opposed to directed linear graphs or sequences such as SMILES.\cite{salakhutdinov2015learning,de2018molgan,you2018graph,gomez2018automatic,Jaeger2018,elton2019deep,zheng2019identifying}
\replaced{Moreover, predictions should be}{ A requirement for any model is that molecules are
considered } invariant under permutation, translation\added{,} and rotation\added{ of the molecular representation}, which calls for undirected graph\replaced{s}{ representations and models that are inherently permutation invariant}.\cite{mater2019deep}\\

\deleted{he tasks of correcting, completing and generating molecules in discrete representations usually rely on the octet rule as the fundamental structure rule. This alone defines the validity, i.e. the stability of molecules.\cite{fink2005virtual,blum2009970,ruddigkeit2012enumeration,qm9,elton2019deep}
Despite the success of the octet rule, it has limitations such as not allowing hyper-valent molecules, ions, and inability to capturing structure phenomena in liquid environments.}

\replaced{We}{ In this paper, we} introduce an unsupervised task, known as masked language modeling or denoising autoencoder, \cite{Bengio_2003,BERT,Vincent2008} over an undirected discrete graph representation of a given molecule.
We define the unsupervised task as \replaced{corrupting }{applying a corruption to} a molecule and learning how to revert such corruption to recover the valid molecule. This objective allows us to learn the underlying structure rule without any hard-coded heuristic\deleted{,} \replaced{by merely}{ but exclusively by} observing valid molecular graphs.\cite{gerratt1997modern}
\added{This can correct molecular graphs directly or cross-validate molecular graphs generated from 3D structures to check for consistency. In addition, the \texttt{binary-transformer} encodings developed for this model are generally of interest for other tasks including generation in context of drug-discovery.}\\

This paper presents several models trained on \replaced{two datasets: QM9, which we use as a benchmark to verify that the models can learn a simple known heuristic defining the dataset -- namely the octet rule -- and ZINC, as a more challenging dataset due to ions and hypervalent molecules }{the QM9 and ZINC} .\cite{qm9,zinc,zincdb} \added{The models are evaluated on several metrics including: perplexity, sample F1, and a new octet F1, which measures if the predictions satisfy the octet rule} \deleted{The ground truth we approximate, and benchmark against, is either that the original atom is correctly predicted (sample F1), or that the prediction is correctly predicted or satisfies the octet rule}.\\

In Natural Language Processing (NLP) the a goal of statistical and probabilistic language modeling is to learn the joint probability mass function of sequences.\cite{Bengio2003}
Historically, this has been accomplished by calculating the probability of observing a word given the sentence that precedes it.\cite{Jurafsky2009}
Methods exploiting the sequential relationship between words in text has been ranging from probabilistic finite automaton\added{,}\cite{Jurafsky2009} to distributed word embeddings \added{,}\cite{Mikolov_2013} and recurrent neural networks (RNN\added{s}).\cite{MikolovKBCK10,ZarembaSV14,MerityKS18}
To cover the most recent development in language modeling, adapted to fit undirected graphs with \replaced{a degree above two}{ high degree}, we test the following methods of increasing complexity:
\begin{itemize}
  \item \texttt{unigram} --- unconditional probabilities of the atoms

  \item \added{\texttt{bag-of-atoms/neighbors} --- neural network that aggregates either all atoms or only neighboring atoms in the molecule}
  
  \item \texttt{binary/bond-transformer} --- neural network architecture with attention using either binary representations of \replaced{connectivity}{ bond types} or full bond type information
\end{itemize}
The \texttt{binary/bond-transformer} \replaced{are}{ is} inspired by a recent trend in NLP, \replaced{known as}{ called} masked language modeling, where the sequential requirement can be relaxed.\cite{BERT}
Most noticeably, we modify masked language modeling to work with molecules by masking atoms and using graph adjacency matrices to model intermolecular relationships.

\section{Methods}
\label{sec:methods}
In this section we present several methods to restore\added{ partially observed} molecules\deleted{that have been corrupted}.
We formally define this as an unsupervised learning task over discrete molecular graphs\added{.
To train the model we apply a a simple corruption function that masks atoms and challenge the model to recover the corruption.
Formal task definition and the five unsupervised models of increasing complexity are defined below, as a baseline we apply the deterministic octet rule.}
\deleted{, where
we use a simple corruption function that masks atoms. We introduce four unsupervised
models with increased modeling complexity: counting, nearest neighbor, all atoms, and the
transformer model; which we compare against the octet-rule.}
\subsection{Unsupervised learning of discrete molecular graphs}
\label{ssec:unsupervised}
%
Autoencoders are a type of neural network\added{s} that \replaced{are trained with}{ use} unsupervised learning.
They create an efficient representation of data by extracting important features.\cite{HintonSalakhutdinov2006b}\replaced{
The denoising autoencoder \cite{Vincent2008} is an autoencoder variant that trains challenges the neural network to revert corruptions of the input.
By reverting corruptions, the neural network has to understand the underlying structure of the data distribution.}
{However, when the autoencoder has a larger hidden than input dimension the neural network risk learning the identity function, which makes the autoencoder useless.
The denoising autoencoder \cite{Vincent2008} alleviates this problem by corrupting the input data on purpose.}\\

In our case the input is a molecule, which we represent as an undirected graph with discrete edges $G=(V, E)$.
Here $V$ is a set of vertices (atoms), such that $(a,i) \in V$ where $a \in A$ is the element\replaced{ and}{ ,} $i \in \mathbb{N}$ is the index.
$E$ is the set of undirected bonds between atoms in the molecule, such that $E \subseteq \{x,y,b\}\mid(x,y)\in V^{2} \wedge (x,y)=(y,x) \wedge x\neq y$, where $b\in \{1,2,3\}$ is the bond type: single, double, or triple.\\

\replaced{We denote the corruption function of the }{ In the}denoising autoencoder \replaced{as}{ we corrupt the input with the corruption function} $\kappa: V \rightarrow \tilde{V}$.

For the experiments we use a corruption function that mask atoms in a molecule with bond type intact. 
This method of corruption is inspired by the masked language model presented in BERT.\cite{BERT}

To apply the corruption function we replace a set of vertices with the \texttt{<MASK>} token as described in equation (\ref{eq:gtilde})
\begin{align}
\tilde{V} &= V - V_{\texttt{subset}} \cup \kappa( V_{\texttt{subset}}),  V_{\texttt{subset}} \subseteq V\label{eq:gtilde}\\
\kappa(a,i) &= (\texttt{<MASK>},i)
\end{align}

Given the corrupted graph, $\tilde{G} = (\tilde{V}, E)$, we want to maximize the probability of recovering the original graph, $G$, which equals maximizing the probability of the masked atoms.
\begin{align}\label{eq:obj-graph}
\max P(G|\tilde{G}) = \max P(V_{\texttt{subset}}|\tilde{G})
\end{align}

In the following subsections we present \replaced{five}{four} models maximizing this objective\replaced{, where each model has an increasing access to graph information and }{ with increasing} modeling complexity.
\subsection{Counting: atomic frequencies}
\label{ssec:counting}
A counting-based model obtains the distribution of atom types by calculating their frequencies over a dataset.
Counting-based models will by intuition have high accuracy when the dataset is biased, which we find the \texttt{QM9} and \texttt{ZINC} are (see Table \ref{tab:unigram_prob}).\\

The count-based model is motivated by the probability chain-rule, where we can model the joint probability of the atoms $v_i = (a, i) \in V$ in a molecule.
\begin{align}\label{eq:conditional}
P(v_1, v_2, \dots, v_n) = P(v_1)P(v_2|v_1)\dots P(v_n| v_1, \dots, v_{n-1})
\end{align}
While equation \ref{eq:conditional} allows us to exactly estimate the conditional atom distribution, the condition grows exponentially with the amount of vertices and becomes infeasible due to the exponential requirement of data and compute.
In NLP the directionality of the sentence allows for clipped, n-gram, versions of equation \ref{eq:conditional} where the prediction of the word distribution is only conditioned on the last $k$ tokens $P(v_n| v_1, \dots, v_{n-1}) = P(v_n| v_{n-k}, \dots, v_{n-1})$.
Using \replaced{n}{N}-grams significantly reduces required computation and data while exploiting\added{ the} locality \replaced{of}{in} language.\cite{Jurafsky2009}\\

In molecules, the degree of vertices and lack of directionality makes such n-gram models cumbersome as each atom can have a tree of recursive n-grams.
Because of such, we limit ourselves to only consider unigram models (1-grams) for the counting case.

A unigram model splits the probability of different terms in a context into a product of individual terms, disregarding the condition of equation \ref{eq:conditional}.
\begin{align}\label{eq:unigram}
P_{\texttt{unigram}}(v_1, v_2, \dots, v_n) &= P(v_1)P(v_2)\dots P(v_n)\\
P(a_j) &= \frac{\texttt{count}(a_j)}{\sum_{a}\texttt{count}(a)}
\end{align}
\replaced{The unigram model has}{Unigram models have} the benefit of being relatively simple to implement and interpret as \replaced{it}{ they} merely count\added{s} the occurrence of elements in the \replaced{training set}{ trainingset}.
The unigram distribution \replaced{of}{ on} the \texttt{QM9} and \texttt{ZINC} \replaced{training sets}{ trainingset} are shown in Table \ref{tab:unigram_prob}.

\begin{table}[H]
\begin{tabular}{|l|l|l|}
\hline
\textbf{Elements} & \textbf{QM9} & \textbf{ZINC} \\ \hline
\textbf{P(H)} & 0.519 & 0.47407 \\ \hline
\textbf{P(C)} & 0.347 & 0.38691 \\ \hline
\textbf{P(O)} & 0.078 & 0.05416 \\ \hline
\textbf{P(N)} & 0.054 & 0.06109 \\ \hline
\textbf{P(F)} & 0.002 & 0.00856 \\ \hline
\textbf{P(P)} & 0 & 0.00001 \\ \hline
\textbf{P(S)} & 0 & 0.00913 \\ \hline
\textbf{P(Cl)} & 0 & 0.00452 \\ \hline
\textbf{P(Br)} & 0 & 0.00144 \\ \hline
\textbf{P(I)} & 0 & 0.00011 \\ \hline
\end{tabular}
\caption{Unigram probabilities for \added{the} QM9 and ZINC \replaced{training sets}{ trainingset}. The unigram probabilities corresponds to the distribution of elements in the dataset.}
\label{tab:unigram_prob}
\end{table}
\replaced{As expected, we observe}{ This indicates} a bias in the \replaced{atoms}{ data} towards the elements $H$ and $C$.
Note that the unigram model will always predict with the same probability distribution of elements for any\deleted{ vertice or} atom as it does not use context.\\

Using our objective from equation \ref{eq:obj-graph} we calculate our unigram probability of the corrupted molecule as
\begin{align}\label{eq:obj-unigram}
\max P(V_{\texttt{subset}}|\tilde{G}) &= \max \prod_{\tilde{v}\in V_{\texttt{subset}}} P(\tilde{v})
\end{align}

\subsection{Bag of vectors: neighbors and atoms}
\label{ssec:bov}
In a \texttt{bag-of-vectors} model a molecule is represented as a multiset of its tokens (elements and/or bonds),\cite{hansen2015machine,hansen2019atomistic} disregarding structure but keeping multiplicity (i.e. multiple occurrences of the same token).
Each token, $x$, is embedded as a trainable vector of real numbers $x\in \mathbb{R}^d$.
By summing the $n$ tokens of a molecule over the $d$ features we obtain the $\texttt{bag-of-vectors}: \mathbb{R}^{n\times d}\rightarrow \mathbb{R}^{d}$ representation (sum is used instead of mean to keep multiplicity).
The \texttt{bag-of-vectors} representation is used as input to a neural network that learns to predict the masked tokens $V_{\texttt{subset}}$.
The token vectors, also known as embeddings, and the neural network are jointly optimised with stochastic gradient descent.\cite{MCCD,Mikolov_2013}
Using eq. \ref{eq:obj-graph} we define two \texttt{bag-of-vector} models for our study: a bag of neighboring atoms (eq. \ref{eq:obj-bon} ) and a bag of all atoms in the corrupted \replaced{atoms}{ molecule} (eq. \ref{eq:obj-boa}).\\

\begin{align}
\max_{\theta}
P_{\theta}^{\texttt{bag-of-neighbors}}(V_{\texttt{subset}}\mid\tilde{G}) &= \max_{\theta}\prod_{v\in V_{\texttt{subset}}} P_{\theta}(v|V_{neighbors})\label{eq:obj-bon} \\
V_{neighbors}&=\{v_j\mid (v_j, v) \in E\}\\
\max_{\theta} P_{\theta}^{\texttt{bag-of-atoms}}(V_{\texttt{subset}}|\tilde{G}) &= \max_{\theta}\prod_{v\in V_{\texttt{subset}}} P_{\theta}(v|\tilde{V})\label{eq:obj-boa}\\
P(x_j|\tilde{X})_{\theta}=\texttt{softmax}(W h_\theta (\tilde{X}))_j &= \frac{\exp((W h_{\theta}(\tilde{X})_j)}{\sum_{i=0}^{|\Sigma|-1}\exp ((W h _{\theta}(\tilde{X}))_i)}\label{eq:softmax}\\
h_{\theta}(\tilde{X}) &=\texttt{NN}(z_{\theta}(\tilde{X}))\label{eq:nn}\\
z_{\theta}(\tilde{X}) &= \sum_{\tilde{x} \in \tilde{X}} \texttt{embedding}(\tilde{x}))\label{eq:bov}
\end{align}
To represent our corrupted tokens in equation \ref{eq:bov}, $\tilde{X}$\added{ being the elements of either $\tilde{V}$ or $V_{neighbors}$}, we use an embedding function.
Embedding functions, $\texttt{embedding}(x)\in R^{d_{emb}}$, are a popular way to represent input tokens in NLP.\cite{Mikolov_2013}
The embedding function uses a dense vector representation for each token class, which allows the embedding function to learn relations between token classes.\added{ The <mask> token is treated as a normal token and thus results in a special mask embedding vector.}
As we want to model all the tokens in the molecule with a neural network we need to have a fixed feature space.
A convenient way to achieve such is the \texttt{bag-of-vectors}, which sums all tokens to achieve a fixed-sized distributed feature representation of $\tilde{X}$.

Given a \texttt{bag-of-vectors} representation, $z_{\theta}$, we want to model the corrupted atoms.
We choose to use a feed forward neural network in equation \ref{eq:nn}, $NN: R^{d_{emb}}\rightarrow R^{d_{nn}}$.
A \replaced{neural network}{ $NN$} is a powerful non-linear function approximator that can learn relations between tokens.

To map the $NN$ output onto probabilities for the element classes we use \added{a} trainable \deleted{a} linear projection, $W\in R^{|\Sigma|\times d_{nn}}$, followed by the \texttt{softmax} function (eq. \ref{eq:softmax}), which squeezes the output to the probability domain.
$|\Sigma|$ denotes the amount of elements we predict over for each atom (e.g in \texttt{QM9} that would be five: H, C, N, O and F).\cite{Jurafsky2009}\\

The bag-of-vector models are trained end-to-end with stochastic gradient descent using a cross-entropy loss function given the set of correctly labelled atoms $V_\texttt{subset}$.\cite{Jurafsky2009}
\begin{align}\label{eq:crossentropy}
L(V_{\texttt{subset}},\tilde{G}) = \sum_{v \in V_{\texttt{subset}}}\log P_{\theta}(v\mid\tilde{G})
\end{align}
Where the conditional probability, $P_{\theta}(v\mid\tilde{G})$, is calculated accordingly to; equation \replaced{\ref{eq:obj-boa}}{ \ref{eq:obj-bon}} for the \texttt{bag-of-\replaced{atoms}{ neighbors}} and equation \replaced{\ref{eq:obj-bon}}{ \ref{eq:obj-boa}} for the \texttt{bag-of-\replaced{neighbors}{ atoms}}.\\

Since these models rely on either pairs of atoms (neighbors) or mere counts (atoms) they can work with a broad family of corruption functions.
However, only including compositional information is a coarse representation of a molecule, e.g. we have several large subsets of molecules in QM9 with fixed element compositions, which have varied structures but identical bag-of-atoms representations.\\
Moreover, in equation \ref{eq:obj-boa} for the all-atom based model we have the same condition for all the predictions, $V_{\texttt{subset}}$.
As such, it will always predict the same distribution for the masked atoms, given the composition.\\

While these models are limited in representational power\deleted{,} they provide a rudimentary baseline for \replaced{comparison to}{  our introduction of } the transformer model on undirected molecular graphs.
\subsection{\replaced{The Transformer}{ Tranformer}: atomic context}
\label{ssec:transformer}
The ideal discrete representation of a molecule must have permutation, translation\added{,} and rotation\added{al} invariance as well as allowing branched and aromatic molecules, in other words, an undirected graph with \replaced{a degree of vertices above two and connectivity description}{ vertices of high degree}.\cite{mater2019deep}\\

\replaced{In this section we present an adaption of}{ We adapt}  the \replaced{Transformer}{ transformer} \cite{Attention} to \replaced{handle such input representation}{ fulfill all the above requirements}.
The \replaced{Transformer}{ transformer} is a neural network architecture that uses repeated adaptive receptive fields (known as attention \cite{bahdanau2014neural}) to model relations between words in a text given their context.
The original \replaced{Transformer}{ transformer}, like many other NLP models, uses sequence information to build context from relative word positioning.
Instead of a sequence representation we\replaced{ represent the molecule by}{ use} an adjacency matrix \deleted{to give a complete description of the molecule}.\cite{relativepos}\\

We test two approaches for encoding bond information: the \texttt{binary-transformer}, where all bonds are binary, and the \texttt{bond-transformer}, where bonds type (1, 2, or 3) is given.\\

Using eq. \ref{eq:obj-graph}, the \replaced{Transformers}{ transformers} take the entire graph representation as input and learn a parameterized function that we train to maximize eq. \ref{eq:obj-transformer}.
\begin{align}\label{eq:obj-transformer}
\max_{\theta}
P_{\theta}^{\texttt{transformer}}(V_{\texttt{subset}}\mid\tilde{G}) &= \max_{\theta}\prod_{v\in V_{\texttt{subset}}} P_{\theta}(v|\Tilde{G})\\
P_{\theta}(v_j|\Tilde{G}) &= \texttt{softmax}(W\ \texttt{transform}_{\theta}(\Tilde{G})^L)_j
\end{align}
Similar to the $\texttt{bag-of-vectors}$ we use a $\texttt{softmax}$ function to learn class (atomic element) probabilities.
The \replaced{Transformers}{ transformers} consist of $L$ $\texttt{transform}_{\theta}$ layers.
Each layer applies a non-linear function to build molecular context. The final layer, $\texttt{transform}_{\theta}(\Tilde{G})^L$, is used for classification.
\replaced{As described in eq. \ref{eq:trans}, each}{ Each} layer consist of an attention mechanism with layer normalization\replaced{;}{,}\cite{layernorm} skip-connections\replaced{;}{,}\cite{resnet,highway} and a feed forward neural network\replaced{,}{;}\cite{rectify} which allows the \replaced{Transformer}{ transformer} to model structures and dependencies for each atom using the entire molecule.\deleted{Where the atomic representation of each layer is defined as $h^l, z^l \in \mathbb{R}^{\mid V\mid \times d_{transform}}$ and $d_{transform}$ is the hidden size of the transformer layers.}
\begin{align}\label{eq:trans}
\texttt{transform}_{\theta}(V, E)^l = h^l &= \texttt{layer-norm}(z^l + \texttt{FFN}(z^l))\\
z^l &= \texttt{layer-norm}(h^{l-1} + \texttt{Attention}(h^{l-1}, E))\\
h^0 &= \texttt{atom-embedding}(V)
\end{align}
\added{Where the atomic representation of each layer is defined as $h^l, z^l \in \mathbb{R}^{\mid V\mid \times d_{transform}}$ and $d_{transform}$ is the hidden size of the transformer layers.}
The $\texttt{atom-embedding}$, $h^0\added{\in \mathbb{R}^{\mid V\mid \times d_{emb}}}$, is identical to the $\texttt{embedding}$ in equation \ref{eq:bov}.
\added{Notice that the size of the distributed representation changes from $d_{emb}$ to $d_{transform}$ in the first transformer layer $h^1$.}
To represent either the full bond type or just the binary edge information we set the adjacency matrix $E_{i,j}\in \{0,1\}$ for the $\texttt{binary-transformer}$ \replaced{and}{ or} $E_{i,j}\in \{0,1,2,3\}$ for the $\texttt{bond-transformer}$.
Like the \texttt{bag-of-vector} models, this is trained with stochastic gradient descent using the cross-entropy loss function (see equation \ref{eq:crossentropy}).
\subsubsection{Attention}
As with the original transformer, we use the key-value lookup \texttt{Attention} function.
This layer can adaptively align information between atoms conditioned on the context of other atoms.\cite{bahdanau2014neural,luong2015effective}
Our  implementation takes a layer of hidden representations, $h^l$, and an adjacency matrix of edges, $E$, as input.
Notice that we have separate trainable $\texttt{bond-embedding}$ functions for the key, $e^K$, and value, $e^V$, edge representations.
\begin{align}\label{eq:selfattn}
    \texttt{Attention}(h, E)_i    &=\sum_{j=1}^{n} \alpha_{i j}\left(h_{j} W^{V} + e_{ij}^V\right)\\
    \alpha_{i j}&=\frac{\exp \phi_{i j}}{\sum_{k=1}^{n} \exp \phi_{i k}}\\
    \phi_{i j}&=\frac{\left(h_{i} W^{Q}\right)\left(h_{j} W^{K} + e_{i,j}^K\right)^{T}}{\sqrt{d_{transform}}}\\
    e &= \texttt{bond-embedding}(E)
\end{align}
\replaced{W}{w}Where $a_{ij}\in [0,1]$ is the attention weights\replaced{;}{,} $n$ is the number of vertices\replaced{; and}{,} $W^Q, W^V,$\deleted{and}$ W^K \in \mathbb{R}^{d_{transform}\times d_{transform}}$ are trainable weights.
The $\texttt{bond-embedding}: E^{|V| \times |V|} \rightarrow \mathbb{R}^{|V| \times |V| \times d_{transform}}$ takes an adjacency matrix and returns a three dimensional tensor with a distributed representation for each edge.\added{ Notice that compared to most graph based models we use information from all the nodes and edges in the graph to calculate the attention weights, at each layer. From our experiments, this improved the performance (see Figure S.3).}\\

\begin{figure}[h]
    \centering
	\includegraphics[scale = 0.5]{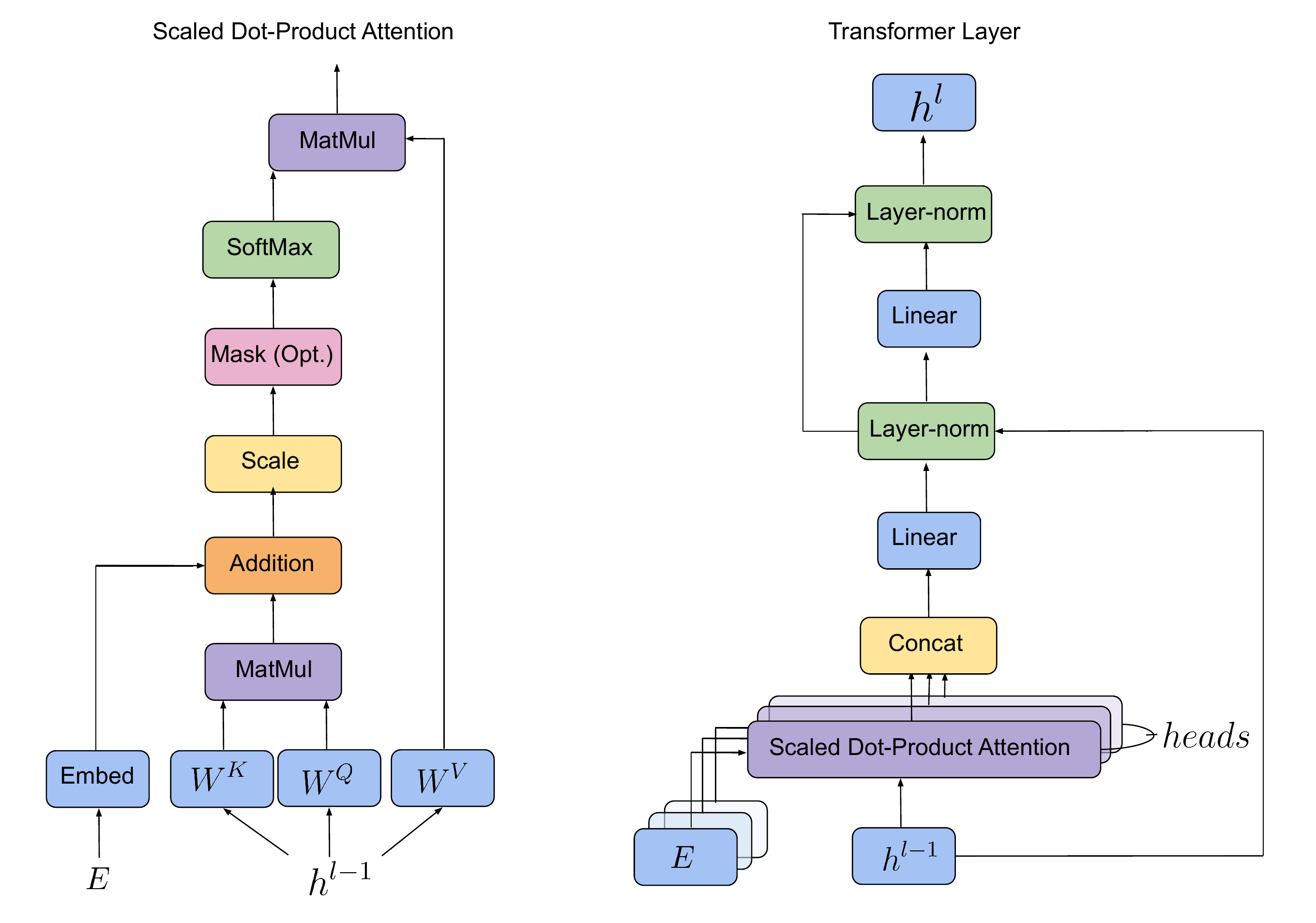} \\
	\caption{(left) Scaled Dot-Product Attention. (right) Multi-Head Attention with multiple layers consisting of several
attention layers running in parallel. Figure modified from \citet{Attention}}.
\label{fig:multihead}
\end{figure}

To have a more expressive attention function we use the multi-head attention mechanism by concatenating $k$ attention layers.
The $k$ attention layers are projected to the hidden size of the network $\mathbb{R}^{d_{transform}\times k} \rightarrow \mathbb{R}^{d_{transform}}$, such that
\begin{align}
    \texttt{Multi-Head-Attention}(h,E)_i 
    &= [\texttt{C}\_1 ,\texttt{C}\_2,\dots \texttt{C}\_k] W_{multi}
\end{align}
where \texttt{C}\_i corresponds to an instance of $\texttt{Attention}$ (eq. \ref{eq:selfattn}) and $W_{multi}\in \mathbb{R}^{(d_{transform}\cdot\ k) \times d_{transform}}$ is a trainable weight.
This is further illustrated in Figure \ref{fig:multihead}

\section{Experimental setup}
\label{sec:setup}
In our experiments we test the described models of the \texttt{unigram}, \texttt{bag-of-neighbors}, \texttt{bag-of-atoms}, \texttt{binary-transformer}, and \texttt{bond-transformer} as denoising autoencoders on the \texttt{QM9} and \texttt{ZINC} datasets.\cite{qm9,zinc}

\subsection{Pre-processing}
\label{ssec:preprocessing}
The \texttt{QM9} dataset has \num{134000} organic molecules with \deleted{of} five types of atoms; $A=\{$H, C, N, O, F\}.
Similarly, the \texttt{ZINC} dataset has \num{250000} drug-like molecules with 10 type\added{s} of atoms, $A=\{$H, C, N, O, F, P, S, Cl, Br, I\}.
The molecules are represented as a SMILE strings \cite{SMILES} corresponding to their discrete graph representations.
We \added{kekulize the molecules -- thus resulting in the dataset only containing single, double and triple bond types -- and }obtain an adjacency matrix for each molecule from the SMILES string using Rdkit.\cite{rdkit}

Since we use the \texttt{QM9} dataset to benchmark our ability to approximate the octet rule, we discard any molecules that contains\added{ atoms with} net charges (1808 molecules). 
In the ZINC dataset, we keep all molecules including molecules with charges and hypervalent molecules.\\

\begin{figure}[H]
    \centering
    \begin{subfigure}[b]{0.49\textwidth}
    \centering
    \includegraphics[width=\textwidth]{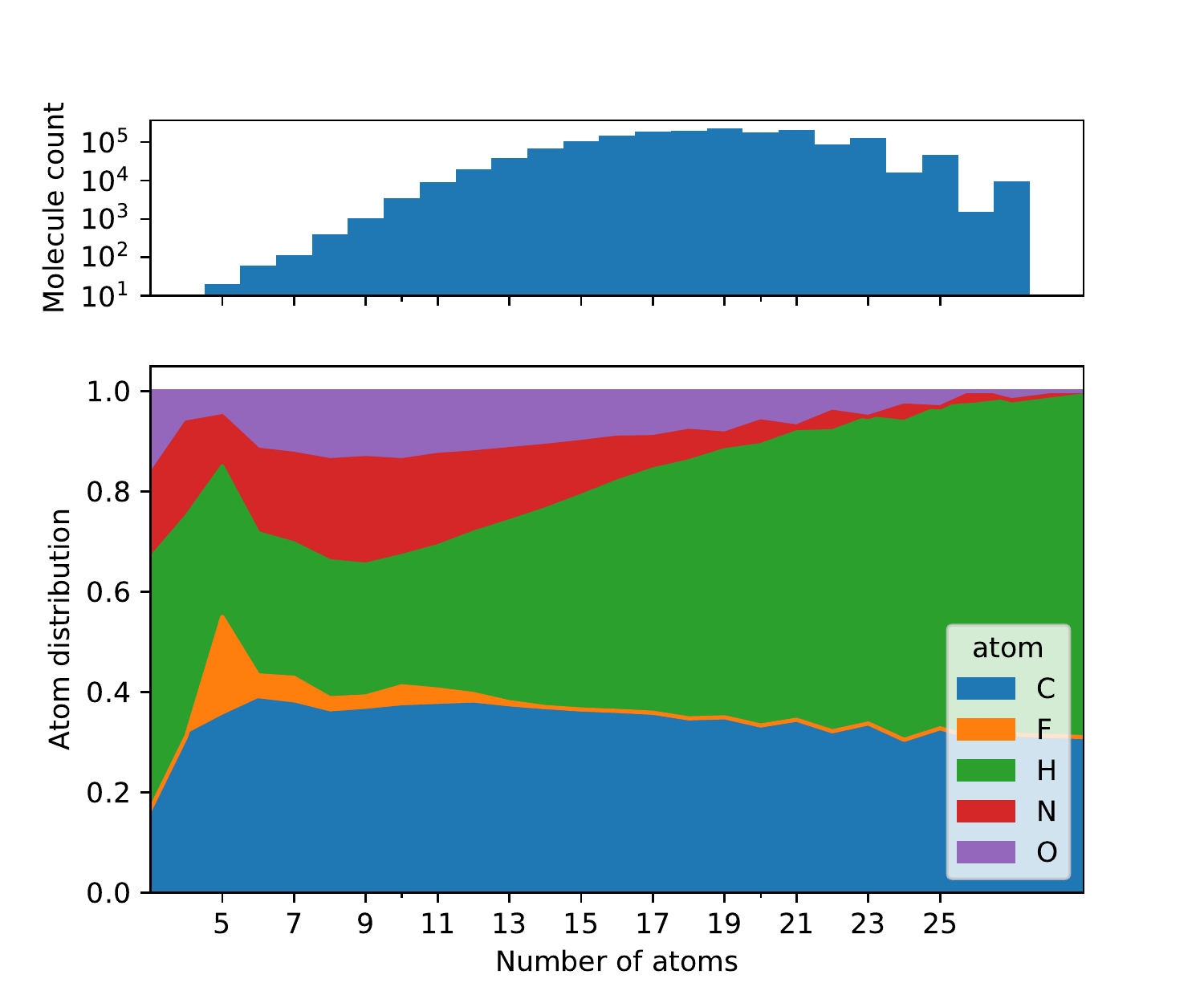}
    \caption{}
    \label{fig:distribution_qm9_train}
    \end{subfigure}
    \hfill
    \begin{subfigure}[b]{0.49\textwidth}
    \centering
    \includegraphics[width=\textwidth]{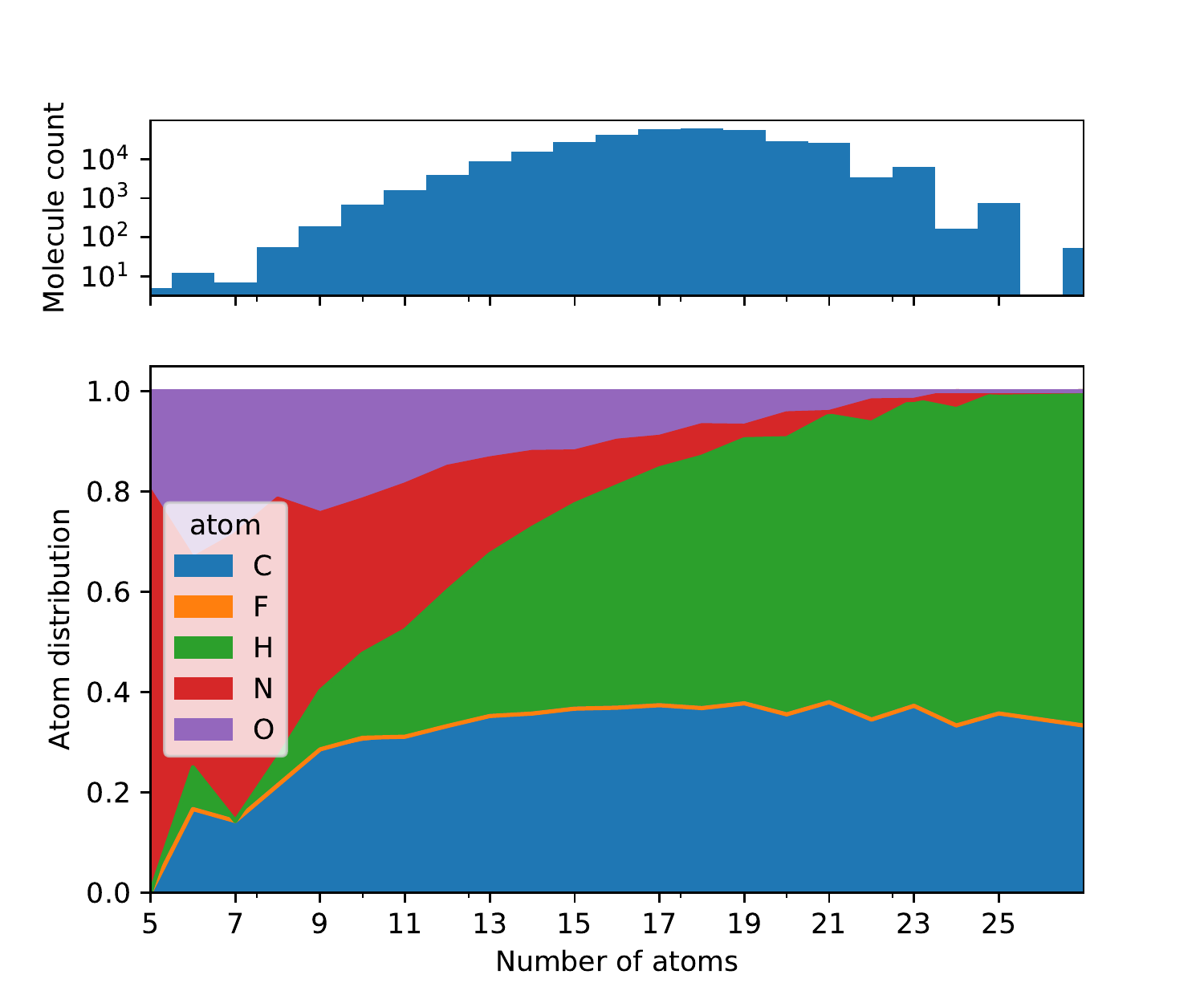}
    \caption{}
    \label{fig:distribution_qm9_test}
    \end{subfigure}

    \centering
    \begin{subfigure}[b]{0.49\textwidth}
    \centering
    \includegraphics[width=\textwidth]{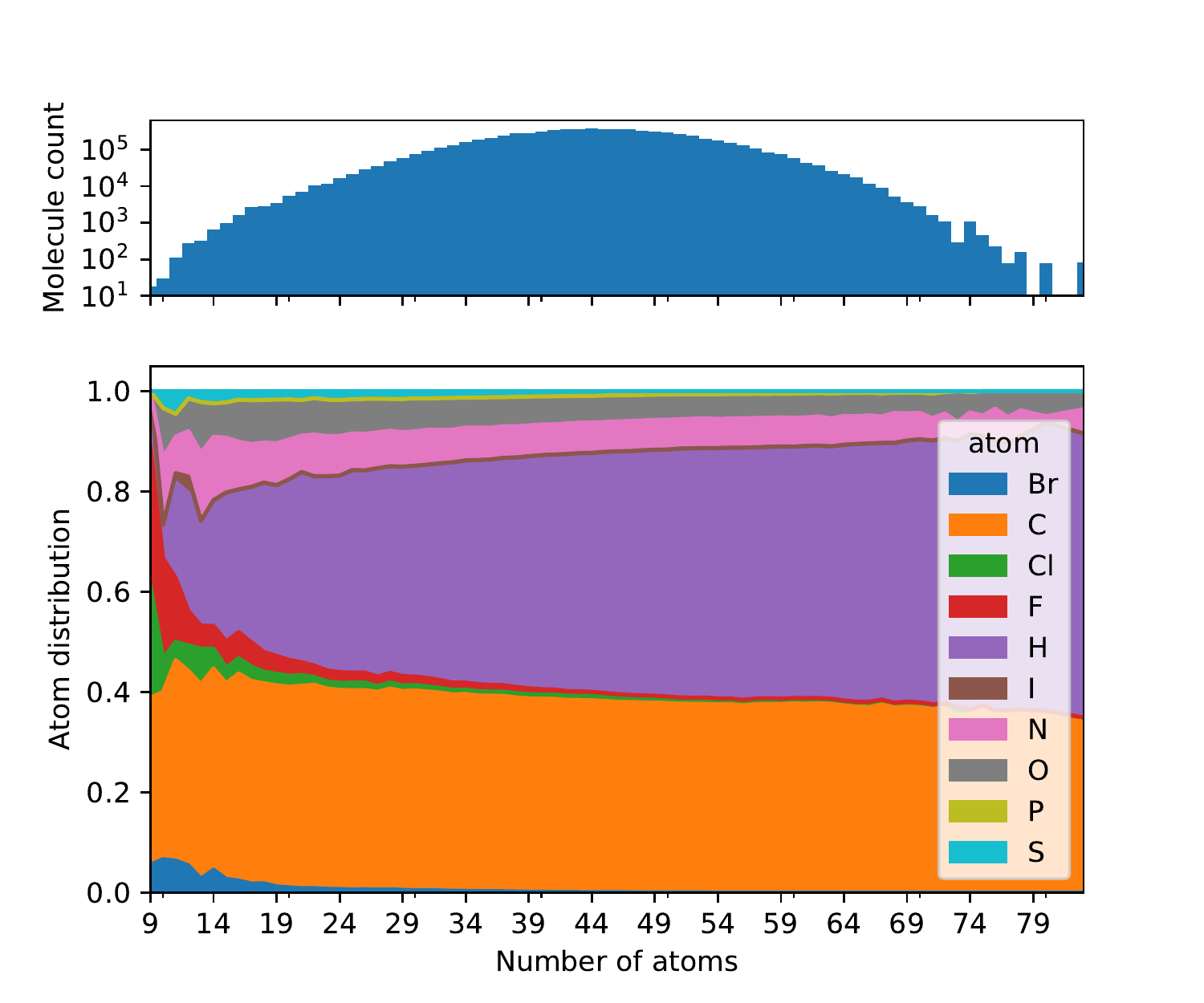}
    \caption{}
    \label{fig:distribution_zinc_train}
    \end{subfigure}
    \hfill
    \begin{subfigure}[b]{0.49\textwidth}
    \centering
    \includegraphics[width=\textwidth]{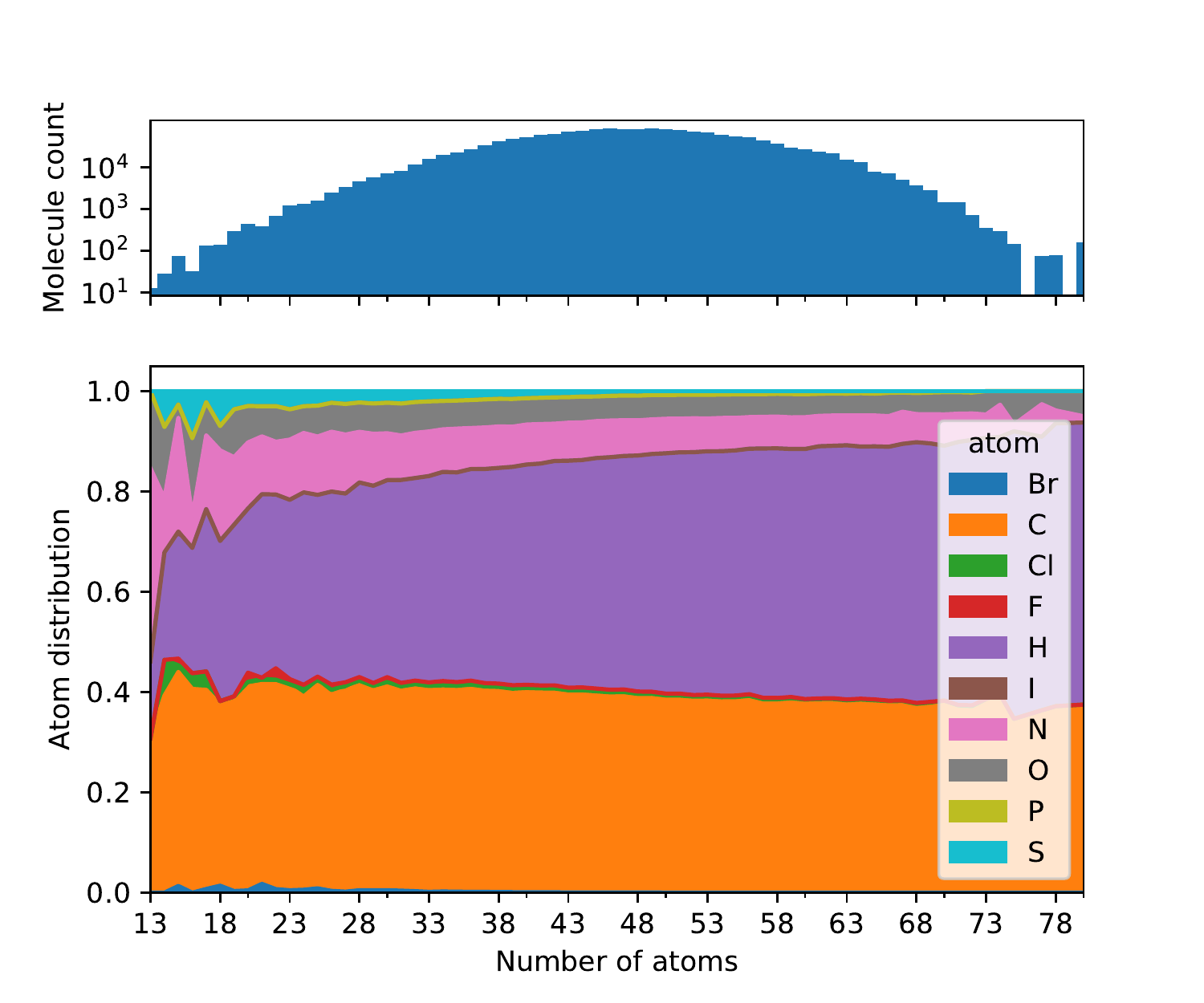}
    \caption{}
    \label{fig:distribution_zinc_test}
    \end{subfigure}
    \caption{Count (top) and distribution of elements per molecule size (number of atoms) for (a) \texttt{QM9} training set, (b) \texttt{QM9} test set, (c) \texttt{ZINC} training set and (d) \texttt{ZINC} test set.}
    \label{fig:distribution_vs_length}
\end{figure}

The resulting set of adjacency matrices are split using scaffolding to homology partition the molecules.
We make a 15\% test, 15\% validation, and 70\% training set split.
In Figure \ref{fig:distribution_vs_length} we show the distribution of elements for different sizes of molecules. Here we see that in both the \texttt{QM9} and \texttt{ZINC} dataset, the size of the molecules are not uniformly distributed, with few small and large molecules. Furthermore, the molecules in \texttt{ZINC} are generally larger than the ones in \texttt{QM9}\added{; up to 80 atoms in \texttt{ZINC} compared to a maximum of around 30 atoms for \texttt{QM9}}. The distribution of different elements depends somewhat on the size of the molecule, especially for smaller molecules.\\

To stress test the models we generate several validation-/ and tests sets with increasing complexity.
For \texttt{ZINC}, the datasets have either 1, 10, 20, 30, 40, 50, 60, 70, or 80 atoms randomly masked in the molecule, denoted by $n_{corrupt}$.

For $n_{corrupt}=1$, we oversample the molecules, by generating five unique different maskings per molecule. This is done to reduce the variance of our estimated performance, especially on molecules with few atoms, since there only \replaced{exist}{exists} few of these in the dataset.\\

\subsection{Training details}
\label{ssec:trainingdetails}
To train the model we optimize the objective for each of the methods (equation \ref{eq:obj-unigram}, \ref{eq:obj-bon}, \ref{eq:obj-boa}, and \ref{eq:obj-transformer}) by corrupting the atoms, with masking, and reversing the corruption.
When increasing the masking we have an exponentially growing combination of corruptions, for which reason we sample the atom modifications in an online manner for training.

To make the model robust towards different levels of corruption we employ an $\epsilon$-greedy corruption scheme.\cite{sutton2018reinforcement}
\begin{align}
\Pr(\text{no. of corruptions} = k) = \left\{\begin{matrix}
1-\epsilon + \frac{\epsilon}{\left | V \right |} & k=n_\texttt{corrupt} \\ 
\frac{\epsilon}{\left | V \right |} & k\neq n_\texttt{corrupt}
\end{matrix}\right.
\end{align}
in the first case, with probability $1-\epsilon$, we corrupt $n_{\texttt{corrupt}}$ atoms and in the second case, with probability $\epsilon-\frac{\epsilon}{\left | V \right |}$, we uniformly corrupt between 1 to $\left | V \right |$ where $\left | V \right |$ is the amount of atoms in the molecule. We use $n_{corrupt}=1$ for training\added{ and found $\epsilon=0.2$ to work well (see Figure S.1)}.
The models are trained\added{ for 100 epochs} on an Nvidia Tesla V100 GPU, using Adam optimization,\cite{ADAM}\replaced{ with}{ and} a learning rate of 0.001\added{ and batch size of 248} for all models.\added{ Not much hyperparameter optimization was done, as these default values performed well. We found that an embedding dimension of 64 and 4 layers worked well for the \texttt{bag-of-atoms} and \texttt{bag-of-neighbors} as more layers caused more overfitting, while 8 layers with 6 attention heads was chosen for the transformers (see Table S.4).} The experiments are implemented in PyTorch\footnote{\url{https://github.com/jeppe742/language_of_molecules}}.\cite{pytorch}

\subsection{Evaluation}
\label{ssec:evaluation}
When predicting the true value of a masked atom in a molecule, several solutions might be equally correct.
In NLP this is often handled by considering sample exact match.
However, for molecular structures we know that multiple elements could exist in the same position.
This is formalized by the octet rule, which allows the prediction of elements with \replaced{the same}{ a similar} number of \added{unpaired} valence electrons. 
We define the union of an exact match and elements that are correct with respect to the octet rule as octet accuracy \deleted{(e.g H and F could both be possible when one bond is present)}.
Given that the \texttt{QM9} dataset is generated by the octet rule, correctly understanding the octet rule would result in 100\% octet accuracy, which is why we use it as our first dataset - to see how difficult it is to learn the octet rule.
The \texttt{ZINC} dataset on the other hand does not conform to the octet rule as it contains \deleted{ions and} \replaced{hypervalent molecules}{ hypervalency}.
This tests our models ability to go beyond the octet rule and learn other underlying structure rules of molecules present in \texttt{ZINC}.\\

Since the distribution of atoms in the data is heavily biased, we use the F1-micro and F1-macro scores, which are a weighted average of the precision and recall.\cite{sasaki2007truth}\\

While the octet rule becomes increasingly \replaced{ambiguous}{ ambigious} when more elements are allowed, understanding what underlying structures are more common, exact match is of interest.
This is important to evaluate if we can fit the specific distribution of a dataset.
We define exact match as sample accuracy and F1.
Moreover, we supply sample perplexity measures, which is a more fine grained way of assessing certainty in model prediction.
\begin{equation}
    \text{Perplexity} = \exp{\left( - \frac{1}{|V_{subset}|}  \sum_{v \in V_{subset}} \log P(v|\tilde{G}) \right)}
    \label{eq:pp}
\end{equation}

We benchmark our proposed models against an octet rule model.
The octet rule model counts the number of covalent bonds of the masked atom and predicts the unigram probabilities of the elements of the corresponding group in the periodic table. We denote this model as the \texttt{octet-rule-unigram}.
When predicting elements with ambiguity (e.g hydrogen and \replaced{fluorine}{fluor} in the \texttt{QM9} dataset) the \texttt{octet-rule-unigram} will therefore not obtain perfect perplexity.
As no predictions \replaced{exist}{ exsists} for hypervalent elements (five and six covalent bonds), the \texttt{octet-rule-unigram} predicts uniform probability.
Notice that as opposed to using a unigram model, this actually gives better perplexity as S is underrepresented in the dataset (see Table \ref{tab:unigram_prob}).

\section{Results}
\label{sec:results}
We test all proposed models on octet and sample accuracy, F1, and perplexity.
First, we evaluate the models on the \texttt{QM9} dataset, where the purpose is to learn an \replaced{approximation to the octet rule}{ Octet-rule approximation}.
Next, we measure the models on the \texttt{ZINC} dataset and attempt to extend the octet approximation with hypervalent molecules and ions.
Finally, we provide a qualitative insight into model prediction by analyzing six different samples (three correct, three incorrect) from the \texttt{binary-transformer}.

\section{QM9 - approximating Octet rule}
\label{ssec:qm9}

In Table \ref{tab:qm9_results_nmask=1}, we evaluate our models on octet rule accuracy, octet rule F1-(micro/macro) and sample perplexity.

As expected, the \texttt{bond-transformer} achieves almost perfect performance (\textbf{99.99\%} octet accuracy), since the task becomes a matter of counting covalent bonds, once you include the order of the bonds.
The \texttt{binary-transformer} also achieves excellent performance (\textbf{99.73\%} octet accuracy), even though it is not given any information about bond types. With 1 masked atom, the problem of recovering the corrupted atom, without any bond types, can be seen as a combinatorial problem. This suggests that the \texttt{binary-transformer} is able approximately solve this problem by inferring the bond orders from the remaining molecule.\\

By only using neighborhood information, the \texttt{Bag-of-neighbors} model gets 90\%, which serves as a very strong baseline, but without the full structural context, the model cannot approximate the \replaced{octet rule}{ octet-rule}.
Similar, by only providing compositional information, the \texttt{Bag-of-Atoms} model, performs significantly worse, showing that structural and neighboring information is important.\\

Finally, the \texttt{Unigram}, relies purely on the frequency of occurrence of elements in the dataset, thus always guessing the masked atom is hydrogen and performs poorly.\\

We provide extended results on masking multiple atoms, transformer model sizes, and accuracy by length in Supporting information.

\begin{table}[H]
\begin{tabular}{|l|l|l|l|l|}
\hline
Model &  Octet Accuracy & Octet F1 (micro/macro) & Perplexity\\ \hline

\texttt{bond-transformer}
 & \textbf{99.99}$\pm$0.01 & \textbf{99.99}$\pm$0.01 / \textbf{99.99}$\pm$0.01 & \textbf{1.002}$\pm$0.001 \\ \hline
 
\texttt{binary-transformer}
 & 99.73 $\pm$0.01 & 99.73 $\pm$0.01 / 93.44 $\pm$4.20 & 1.009 $\pm$0.002 \\ \hline
 
\texttt{bag-of-neighbors}
 & 90.67 $\pm$0.01  & 90.67 $\pm$0.01 / 77.18 $\pm$0.01 & 1.281 $\pm$0.004 \\ \hline
 
\texttt{bag-of-atoms}
 & 65.77 $\pm$4.48 & 65.77 $\pm$4.48 / 44.30 $\pm$4.92 & 3.310 $\pm$0.478  \\ \hline
 
\texttt{Unigram}
 & 47.32  & 47.32 \ \ \ \ \ \ \ \ \ / 32.85 & 3.104 \\ \hline
 
 \texttt{octet-rule-unigram}
 & 100   & 100 \ \ \ \ \ \ \ \ \ \ \ / 100 & 1.002 \\ \hline
\end{tabular}
\caption{Performance of our models for 1 masked atoms per molecule. The uncertainty corresponds to the standard deviation of ten models, trained with different start seed.}
\label{tab:qm9_results_nmask=1}
\end{table}

\section{ZINC - going beyond the octet rule}
\label{ssec:zinc}

We consider the \texttt{ZINC} dataset as it cannot be fully explained by the octet rule and has \replaced{ a larger quantity of ambiguous elements }{ a more of ambiguous elements and a larger quantity of them} than \texttt{QM9}. E.g. with $n_{corrupt}=1$, our \texttt{ZINC} \replaced{test set}{testset} contains \num{924} \replaced{fluorine atoms}{ Fluors} to be predicted as opposed to \num{9} \replaced{fluorine atoms}{ Fluors} in \texttt{QM9}.

Given some elements, namely ions and hypervalent molecules, cannot be predicted by the octet rule we add k-smoothing \cite{Jurafsky2009} to the \texttt{octet-rule-unigram} model.
This avoids the case of 0 probability, which would result in infinite perplexity loss.
We optimize k on the validation set and found the optimum at k=1842.\added{(see Figure S2)}\\

From Table \ref{tab:results_zinc_mask1} we see that the \texttt{octet-rule-unigram} model no longer has 100\% octet F1, which emphasizes to what extend that the dataset cannot be fully explained by the octet rule, due to molecules with charges and hypervalency.
Both our transformer models \replaced{perform}{ performs} similar or better than the \texttt{octet-rule-unigram}, when evaluated on Octet F1, sample F1 and sample perplexity.
This is especially the case with with F1 macro, that puts more emphasis on the underrepresented cases, which in our case are the most interesting. This indicates that the transformer models also have learned to discriminate between elements that should be equally likely from the perspective of the octet rule, but might have higher likelihood under a given structure.

\begin{table}[H]
\begin{tabular}{|l|l|l|l|l|}
\hline
Model &  Octet F1 (micro/macro)& Sample F1 (micro/macro) & Perplexity\\ \hline

\texttt{bond-transformer}
 & \textbf{99.52}$\pm$0.04 / \textbf{97.97}$\pm$3.17 & \textbf{98.64}$\pm$0.03 / \textbf{62.67}$\pm$3.19 & \textbf{1.047}$\pm$0.001 \\ \hline
 
\texttt{binary-transformer}
 & 99.13 $\pm$0.05 / 91.38 $\pm$4.94 & 98.18 $\pm$0.06 / 55.76 $\pm$4.89 & 1.063 $\pm$0.002 \\ \hline
 
\texttt{octet-rule-unigram}
 & 99.17 \ \ \ \ \ \ \ \ \ / 88.65  & 97.22 \ \ \ \ \ \ \ \ \ / 38.48 & 1.164 \\ \hline
 
\texttt{bag-of-neighbors}
 & 90.73 $\pm$0.03 / 76.50 $\pm$0.45 & 89.00 $\pm$ 0.03 / 29.47 $\pm$0.37 & 1.412 $\pm$0.004 \\ \hline
 
\texttt{bag-of-atoms}
 & 50.84 $\pm$0.30 / 56.75 $\pm$0.58 & 49.06 $\pm$0.32 / 9.84 \ $\pm$0.53 & 3.135 $\pm$0.073  \\ \hline
 
\texttt{Unigram}
 & 48.05 \ \ \ \ \ \ \ \ \ / 56.40 & 46.10 \ \ \ \ \ \ \ \ \ / 6.31 & 3.221 \\ \hline
\end{tabular}
\caption{Performance of our models for 1 masked atoms per molecule. The uncertainty corresponds to the standard deviation of ten models, trained with different start seed.}
\label{tab:results_zinc_mask1}
\end{table}

\begin{figure}[H]
    \centering
    \includegraphics[width=0.8\textwidth]{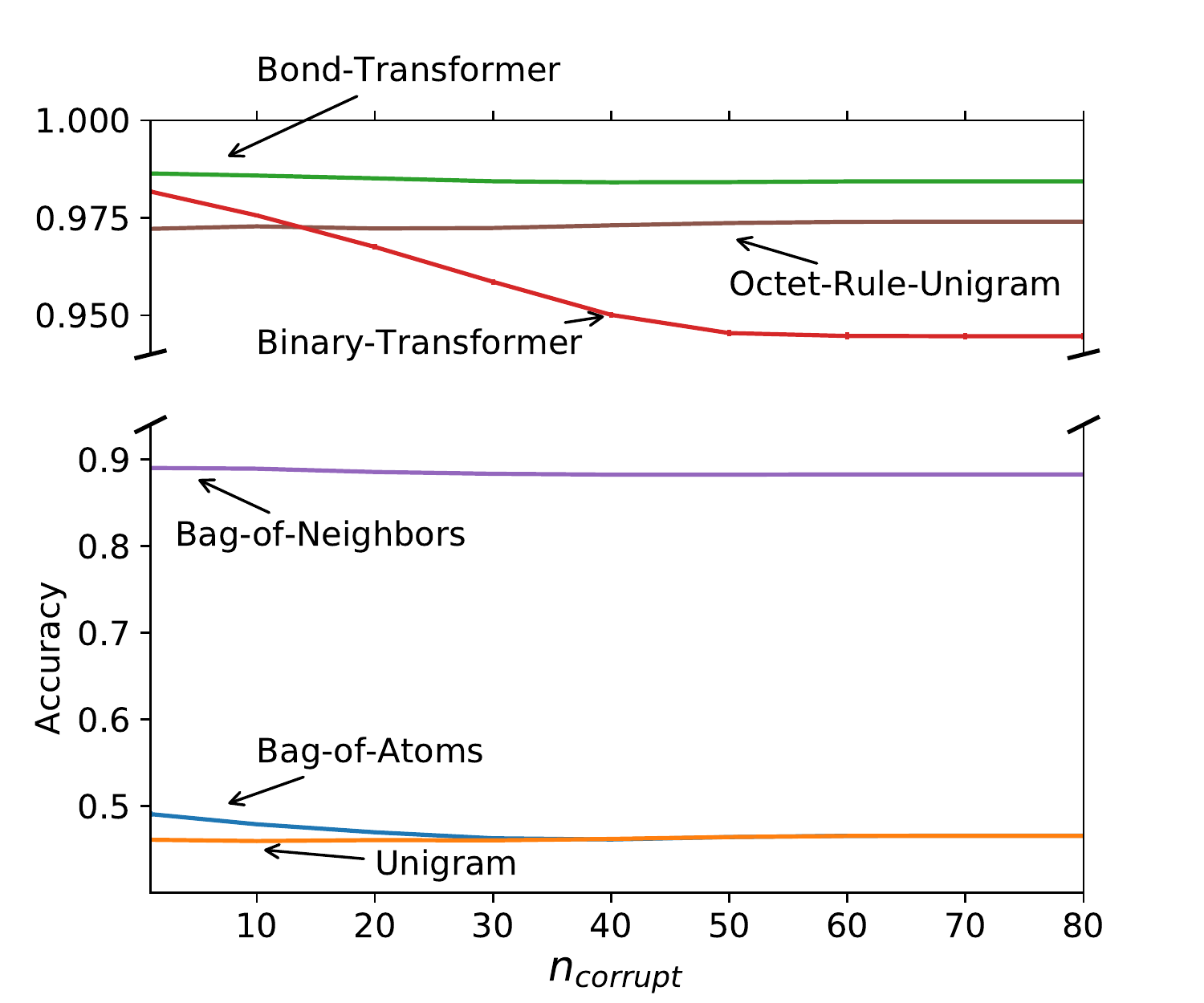}
    \caption{Sample accuracy of the models, evaluated by different number of masked atoms on ZINC dataset. Errors bar corresponds to standard deviation of 10 models trained with different start seed.}
    \label{fig:acc_vs_nmask_zinc}
\end{figure}

Since our model has the ability to corrupt multiple atoms in a molecule, we investigate how the amount of corruption affects the performance. This is shown in Figure \ref{fig:acc_vs_nmask_zinc} (see Supporting information for F1 metrics, and accuracy/F1 by number of atoms). Here we see that the accuracy of \texttt{Bond-Transformer} barely is affected, even when all the atoms in the molecule are masked. This suggests that the model primarily uses the structural information (bond type and connections).
The \texttt{Binary-Transformer} however drops slightly in accuracy as the molecule is corrupted. This makes sense, as without bond type information, the model can use the label of the remaining atoms to infer the bond types, but as we corrupt more, we limit the available information in the molecule.
The same is the case for the \texttt{Bag-of-neighbors}.
In the case of \texttt{Bag-of-atoms}, the model seem to converge to the \texttt{Unigram}.\\

\begin{figure}[H]
    \centering
    \includegraphics[width=\textwidth]{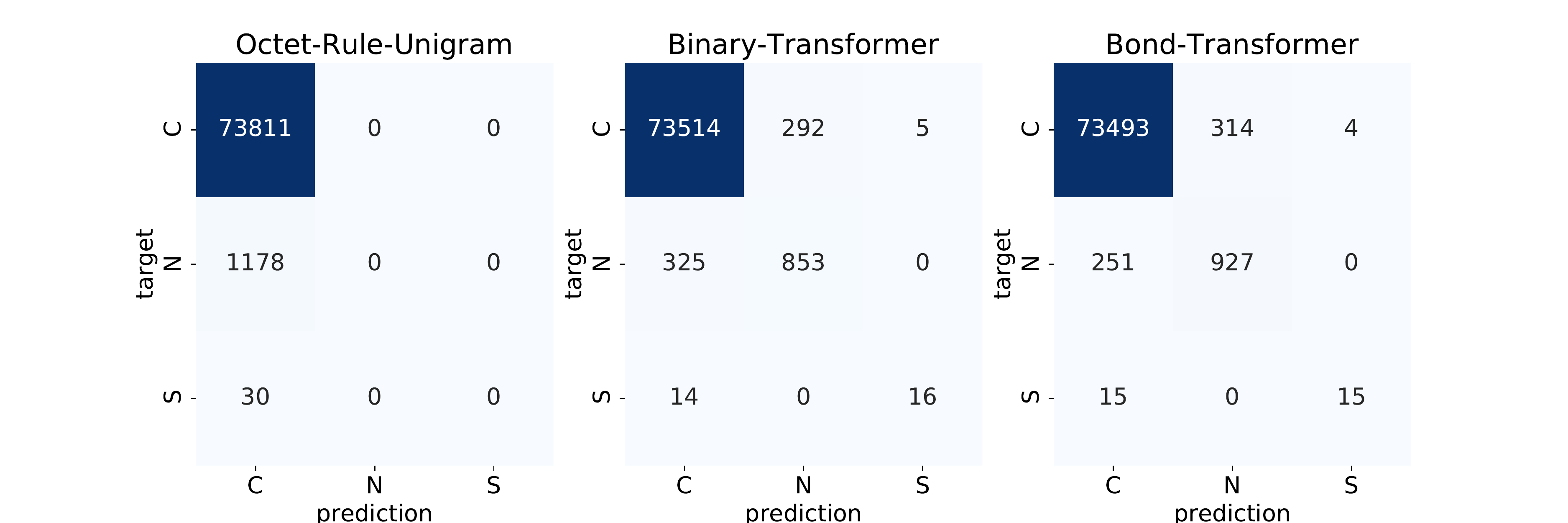}
    \caption{Confusion matrix for the \replaced{test set}{testset}, with $n_{\texttt{corrupt}}=1$, where the masked atom has four covalent bonds. We provide this matrix for the \texttt{octet-rule-unigram}, \texttt{binary-transformer}, and \texttt{bond-transformer}.}
    \label{fig:confusion4_zinc}
\end{figure}

To investigate if our model can understand ions we have visualized the confusion matrix for atoms with four covalent bonds in Figure \ref{fig:confusion4_zinc} (other bond order confusion matrices can be found in Supporting information). \replaced{For a masked atom with four covalent bonds the possible classes in the dataset are a C, a $\text{N}^+$ ion or a hypervalent S}{In this case, our dataset contains examples of C, but also $\text{N}^+$ ions and hypervalent S}.
The confusion matrix shows that while our \texttt{Octet-rule-unigram} model only predicts C, both the \texttt{Binary-transformer} and \texttt{Bond-transformer} has learned, that both S and N can have four covalent bonds and how to discriminate between them. Thus the models seems to have successfully learned a more complex structure rule, than the octet rule.\\

To better understand the models success in predicting hypervalent elements we visualize the confusion matrix for five and six covalent bonds in Figure S\replaced{8}{1} and S\replaced{9}{2} (see Supporting information).
With five covalent bonds we only have one occurrence of P, which is correctly predicted by the \texttt{bond-transformer}.
For six covalent bonds, both transformers correctly predict all elements with S.\\

To assess the models ability for predicting ambiguous elements we visualize the confusion matrix for one covalent bond in Figure S2.
In particular, we find that both transformer models (\texttt{binary-transformer}/\texttt{bond-transformer}) can successfully predict a large number of F molecules (279/270) while only misclassifying a small amount of H (23/21) as F.\\

For future investigations, we find that the \texttt{QM9} and \texttt{ZINC} datasets are heavily biased towards H and C.
This might make training difficult due to dataset imbalances and could be improved by oversampling rare elements.\cite{Buda2018ASS}

\section{Qualitative results}
\label{sec:qualitativeresults}
\begin{figure}[H]
    \centering
    \begin{subfigure}[b]{0.45\textwidth}
    \centering
    \includegraphics[width=\textwidth]{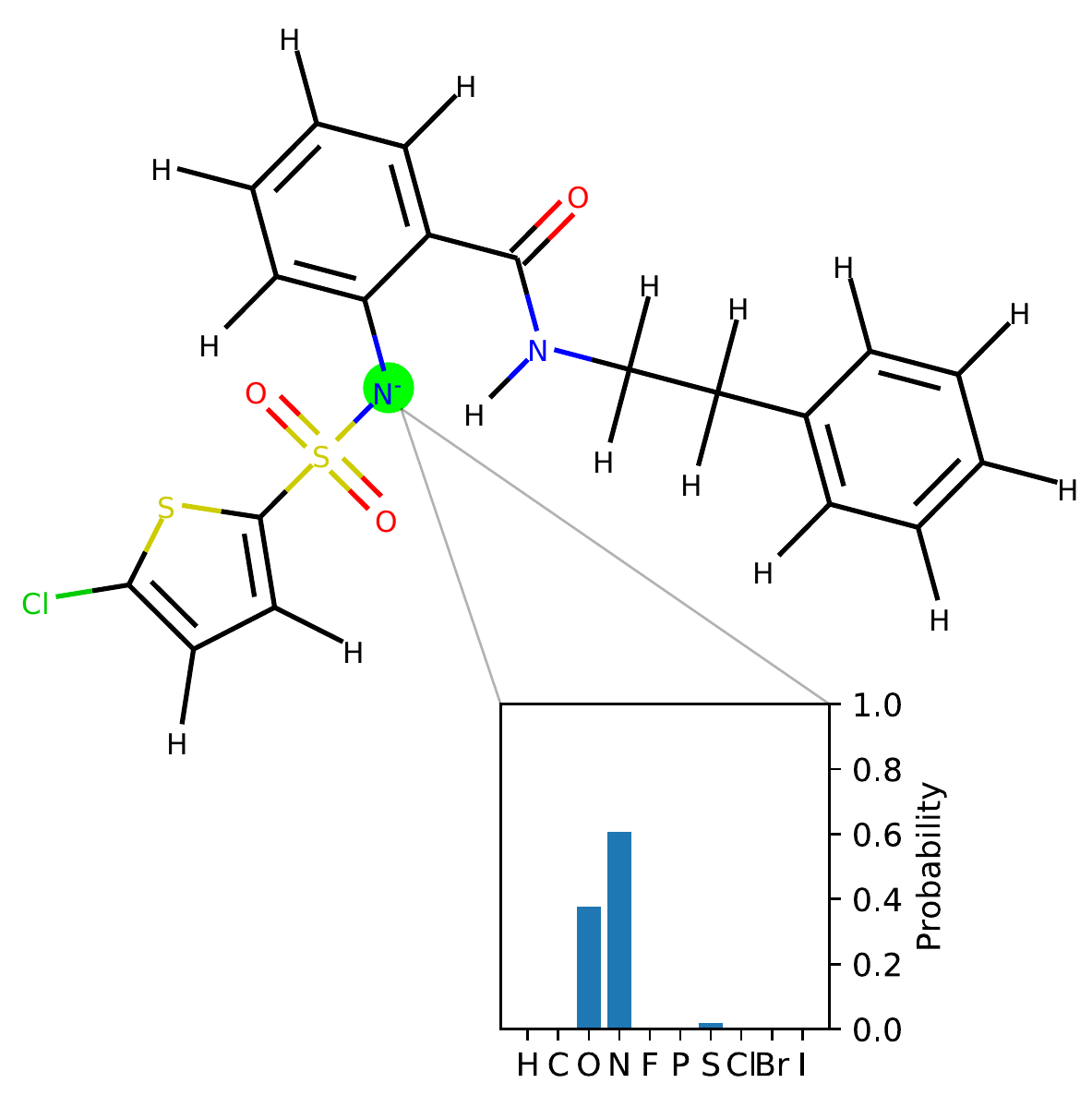}
    \caption{}
    \label{fig:predictions_good1}
    \end{subfigure}
    \hfill
    \begin{subfigure}[b]{0.45\textwidth}
    \centering
    \includegraphics[width=\textwidth]{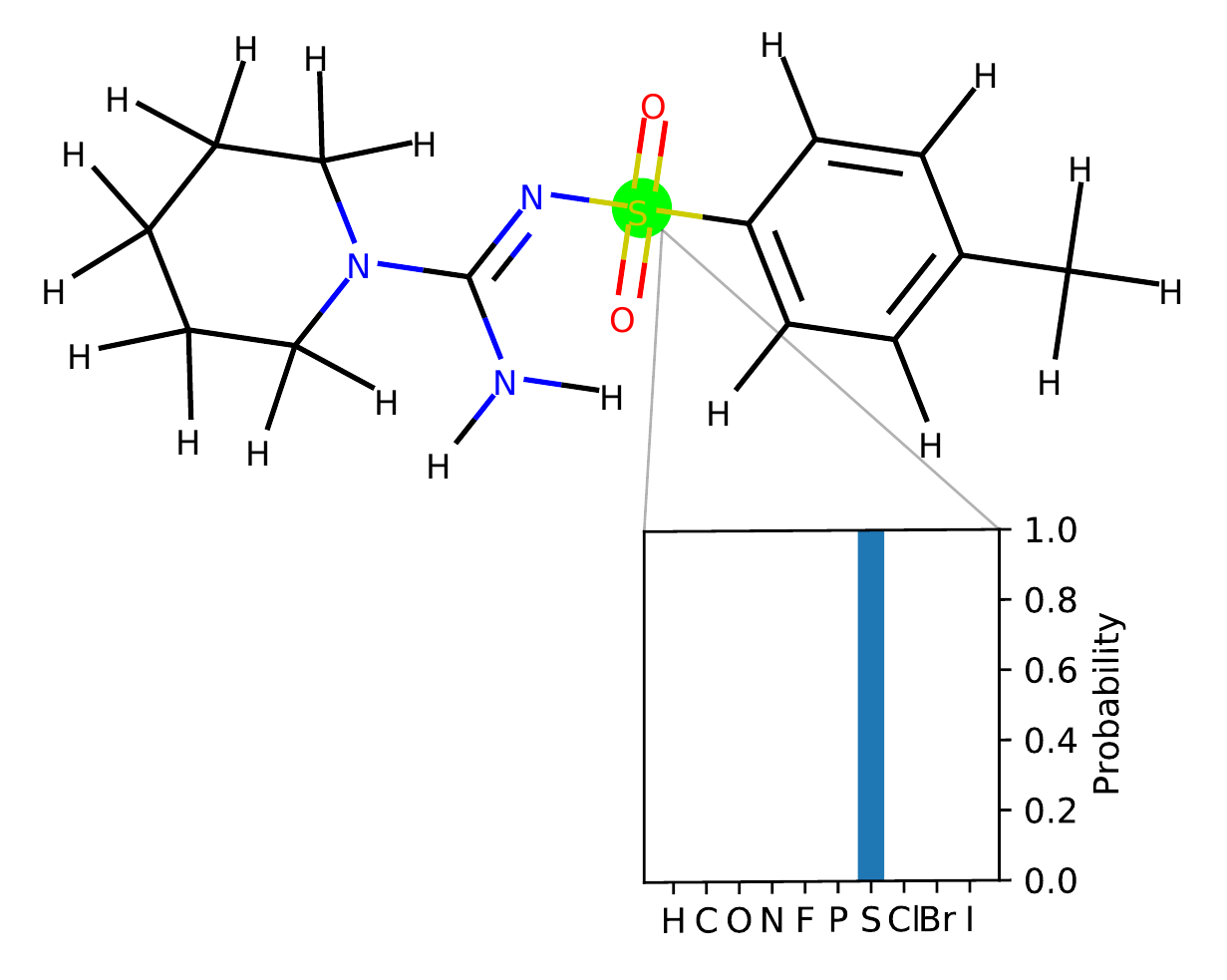}
    \caption{}
    \label{fig:predictions_good2}
    \end{subfigure}
    \hfill
    \begin{subfigure}[b]{0.45\textwidth}
    \centering
    \includegraphics[width=\textwidth]{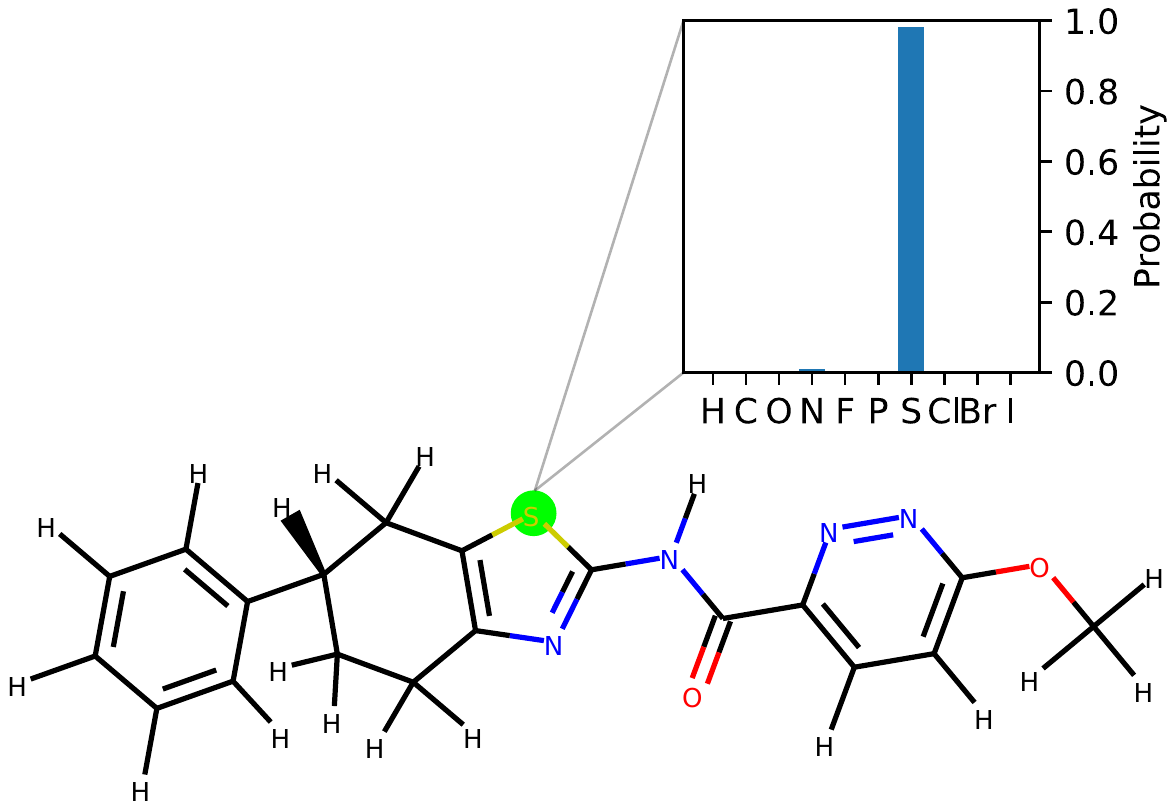}
    \caption{}
    \label{fig:predictions_good3}
    \end{subfigure}
    \hfill
    \begin{subfigure}[b]{0.45\textwidth}
    \centering
    \includegraphics[width=\textwidth]{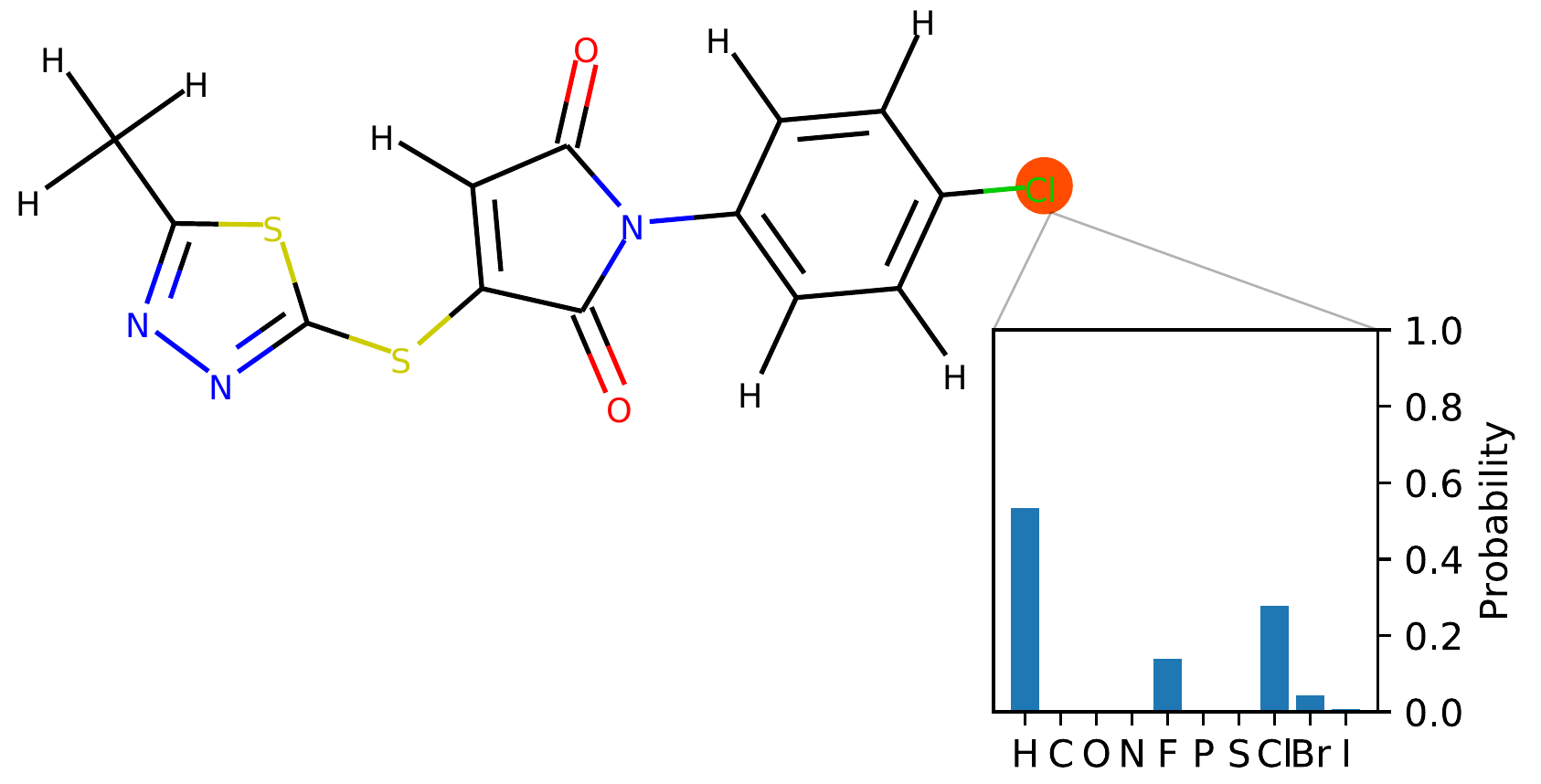}
    \caption{}
    \label{fig:predictions_bad1}
    \end{subfigure}
    \hfill
    \begin{subfigure}[b]{0.45\textwidth}
    \centering
    \includegraphics[width=\textwidth]{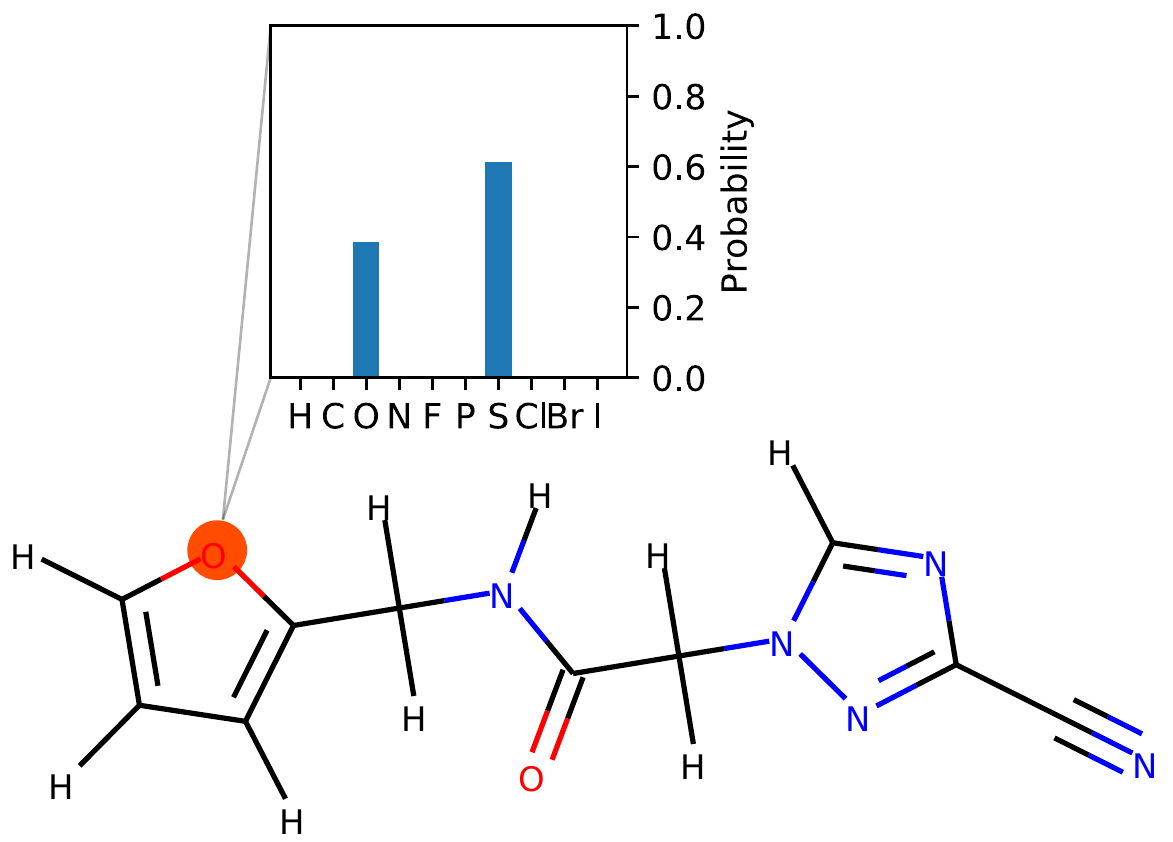}
    \caption{}
    \label{fig:predictions_bad2}
    \end{subfigure}
        \hfill
    \begin{subfigure}[b]{0.45\textwidth}
    \centering
    \includegraphics[width=\textwidth]{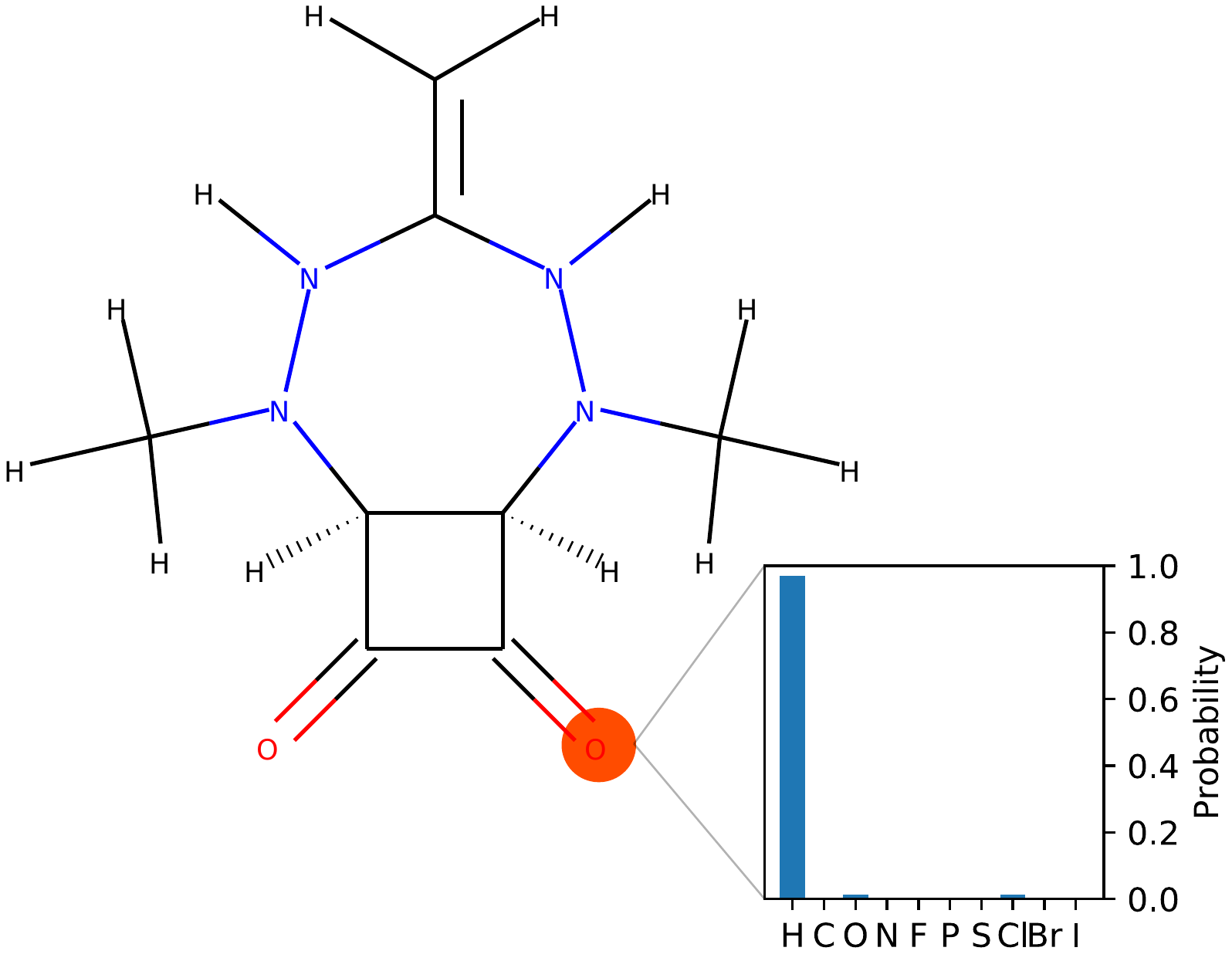}
    \caption{}
    \label{fig:predictions_bad3}
    \end{subfigure}
    \caption{Predicted atom probabilities. The molecule corresponds to the true molecule, where the colored atom is the target we want to predict. Green corresponds to correct, and red to wrong predictions.}
    \label{fig:predictions_zinc}
\end{figure}

To investigate the \texttt{binary-transformer} corrections of atoms in a molecule, we inspect a few interesting predictions on the \texttt{ZINC} dataset.
We show the molecules with the predicted conditional probabilities of the possible element labels on the masked atoms.
Figure \ref{fig:predictions_good1} illustrates an example where the model correctly predicts N, even though $\text{N}^-$ ions are very rare in the dataset. It also puts a reasonable amount of probability of the target being O, which could be a valid guess assuming the octet rule applies.
In Figure \ref{fig:predictions_good2} we see an example of a hypervalent S, which our model correctly predicts, with a very high certainty. The hypervalent S often appears in the dataset with the two double bonded O, which might be a giveaway for the model.
The example in Figure \ref{fig:predictions_good3} would however most likely not have a immediate explanation, but the model is very certain of it prediction, which is also correct.

The context of the elements with one covalent bond is expected to be identical, under the octet rule, since both have one neighbor to any of the other elements in the data, but since the data is heavily biased towards hydrogen it is worth checking if the predicted probabilities are also biased.
From Figure \ref{fig:predictions_bad1}, we see that even though the model incorrectly predicts H, the second most likely guess of Cl is correct, even though F appears twice as often in the dataset. A similar case can be seen in  \ref{fig:predictions_bad2} where the model is in doubt between two elements, that both could be considered correct under the octet rule.
Finally, in Figure \ref{fig:predictions_bad3} we have an example where the model is very certain, but makes a completely wrong prediction.

\section{Conclusion}
\label{sec:conclusion}
In this work we have introduced the \texttt{binary-transformer} and \texttt{bond-transformer} models, and evaluated their ability to recover masked atoms in an undirected molecular graph with discrete representations of bonds.
The models achieves $99.73 \pm 0.01$ \% and $99.99 \pm 0.01$ octet F1-micro on the \texttt{QM9} dataset, while masking 1 atom per molecule, suggesting that the model is capable of learning the octet rule, which is the underlying selection criteria for the QM9 \replaced{dataset}{ data set}.\\
When evaluated on the \texttt{ZINC} dataset, which contains more complex structure rules, our transformer models outperforms the \texttt{octet-rule-unigram} model in all metrics, including achieving $99.52 \pm 0.04$ and $99.13 \pm 0.05$ octet F1-micro, when masking 1 atom per molecule. When paired with the analysis of the confusion matrix, this indicates that the models has learned rules that exceed the octet rule, like ions and hypervalent molecules.

Deep learning models are extremely flexible and we have shown that the transformer architecture, which makes no assumption of the amount of atoms or bonds in a molecule, and could in theory be able to model a wide variety of molecular rules.
With the high accuracy on the \texttt{QM9} and \texttt{ZINC} datasets we hypothesize that the transformer models, both the bond and binary based versions, could be well suited for learning other molecular rules, such as structure rules related to properties.
As inference with the transformer is cheap, correcting billions of molecules is therefore possible.\\

The transformer model and embeddings made from undirected molecular graphs may furthermore be useful in chemical discovery tasks such as automatically generating and enumerating new molecules.\\

Moreover, years of progress in language modeling for NLP has given rise to strong contextual vectors of information that is now the defacto standard for state-of-the-art models in close to every popular dataset for benchmarking neural network performance. \cite{ELMO,BERT,mtdnn}
In particularly, these pretrained language models works surprisingly well for areas of limited labeled data, something that is fairly prevalent in many molecular chemistry tasks as data might be expensive to gather.
\section{Acknowledgments}
This research is funded by the Innovation Foundation Denmark through the DABAI project

\bibliography{refs}

\providecommand{\latin}[1]{#1}
\makeatletter
\providecommand{\doi}
  {\begingroup\let\do\@makeother\dospecials
  \catcode`\{=1 \catcode`\}=2 \doi@aux}
\providecommand{\doi@aux}[1]{\endgroup\texttt{#1}}
\makeatother
\providecommand*\mcitethebibliography{\thebibliography}
\csname @ifundefined\endcsname{endmcitethebibliography}
  {\let\endmcitethebibliography\endthebibliography}{}
\begin{mcitethebibliography}{53}
\providecommand*\natexlab[1]{#1}
\providecommand*\mciteSetBstSublistMode[1]{}
\providecommand*\mciteSetBstMaxWidthForm[2]{}
\providecommand*\mciteBstWouldAddEndPuncttrue
  {\def\EndOfBibitem{\unskip.}}
\providecommand*\mciteBstWouldAddEndPunctfalse
  {\let\EndOfBibitem\relax}
\providecommand*\mciteSetBstMidEndSepPunct[3]{}
\providecommand*\mciteSetBstSublistLabelBeginEnd[3]{}
\providecommand*\EndOfBibitem{}
\mciteSetBstSublistMode{f}
\mciteSetBstMaxWidthForm{subitem}{(\alph{mcitesubitemcount})}
\mciteSetBstSublistLabelBeginEnd
  {\mcitemaxwidthsubitemform\space}
  {\relax}
  {\relax}

\bibitem[Ertl \latin{et~al.}(2000)Ertl, Rohde, and Selzer]{ertl2000fast}
Ertl,~P.; Rohde,~B.; Selzer,~P. Fast calculation of molecular polar surface
  area as a sum of fragment-based contributions and its application to the
  prediction of drug transport properties. \emph{Journal of medicinal
  chemistry} \textbf{2000}, \emph{43}, 3714--3717\relax
\mciteBstWouldAddEndPuncttrue
\mciteSetBstMidEndSepPunct{\mcitedefaultmidpunct}
{\mcitedefaultendpunct}{\mcitedefaultseppunct}\relax
\EndOfBibitem
\bibitem[Lo \latin{et~al.}(2018)Lo, Rensi, Torng, and Altman]{lo2018machine}
Lo,~Y.-C.; Rensi,~S.~E.; Torng,~W.; Altman,~R.~B. Machine learning in
  chemoinformatics and drug discovery. \emph{Drug discovery today}
  \textbf{2018}, \emph{23}, 1538--1546\relax
\mciteBstWouldAddEndPuncttrue
\mciteSetBstMidEndSepPunct{\mcitedefaultmidpunct}
{\mcitedefaultendpunct}{\mcitedefaultseppunct}\relax
\EndOfBibitem
\bibitem[Ulissi \latin{et~al.}(2017)Ulissi, Medford, Bligaard, and
  N{\o}rskov]{ulissi2017address}
Ulissi,~Z.~W.; Medford,~A.~J.; Bligaard,~T.; N{\o}rskov,~J.~K. To address
  surface reaction network complexity using scaling relations machine learning
  and DFT calculations. \emph{Nature communications} \textbf{2017}, \emph{8},
  14621\relax
\mciteBstWouldAddEndPuncttrue
\mciteSetBstMidEndSepPunct{\mcitedefaultmidpunct}
{\mcitedefaultendpunct}{\mcitedefaultseppunct}\relax
\EndOfBibitem
\bibitem[Boes \latin{et~al.}(2019)Boes, Mamun, Winther, and
  Bligaard]{boes2019graph}
Boes,~J.~R.; Mamun,~O.; Winther,~K.; Bligaard,~T. Graph Theory Approach to
  High-Throughput Surface Adsorption Structure Generation. \emph{The Journal of
  Physical Chemistry A} \textbf{2019}, \relax
\mciteBstWouldAddEndPunctfalse
\mciteSetBstMidEndSepPunct{\mcitedefaultmidpunct}
{}{\mcitedefaultseppunct}\relax
\EndOfBibitem
\bibitem[Van~Geem \latin{et~al.}(2010)Van~Geem, Pyl, Marin, Harper, and
  Green]{van2010accurate}
Van~Geem,~K.~M.; Pyl,~S.~P.; Marin,~G.~B.; Harper,~M.~R.; Green,~W.~H. Accurate
  high-temperature reaction networks for alternative fuels: butanol isomers.
  \emph{Industrial \& engineering chemistry research} \textbf{2010}, \emph{49},
  10399--10420\relax
\mciteBstWouldAddEndPuncttrue
\mciteSetBstMidEndSepPunct{\mcitedefaultmidpunct}
{\mcitedefaultendpunct}{\mcitedefaultseppunct}\relax
\EndOfBibitem
\bibitem[Broadbelt and Pfaendtner(2005)Broadbelt, and
  Pfaendtner]{broadbelt2005lexicography}
Broadbelt,~L.~J.; Pfaendtner,~J. Lexicography of kinetic modeling of complex
  reaction networks. \emph{AIChE journal} \textbf{2005}, \emph{51},
  2112--2121\relax
\mciteBstWouldAddEndPuncttrue
\mciteSetBstMidEndSepPunct{\mcitedefaultmidpunct}
{\mcitedefaultendpunct}{\mcitedefaultseppunct}\relax
\EndOfBibitem
\bibitem[Fink \latin{et~al.}(2005)Fink, Bruggesser, and
  Reymond]{fink2005virtual}
Fink,~T.; Bruggesser,~H.; Reymond,~J.-L. Virtual exploration of the
  small-molecule chemical universe below 160 daltons. \emph{Angewandte Chemie
  International Edition} \textbf{2005}, \emph{44}, 1504--1508\relax
\mciteBstWouldAddEndPuncttrue
\mciteSetBstMidEndSepPunct{\mcitedefaultmidpunct}
{\mcitedefaultendpunct}{\mcitedefaultseppunct}\relax
\EndOfBibitem
\bibitem[Blum and Reymond(2009)Blum, and Reymond]{blum2009970}
Blum,~L.~C.; Reymond,~J.-L. 970 million druglike small molecules for virtual
  screening in the chemical universe database GDB-13. \emph{Journal of the
  American Chemical Society} \textbf{2009}, \emph{131}, 8732--8733\relax
\mciteBstWouldAddEndPuncttrue
\mciteSetBstMidEndSepPunct{\mcitedefaultmidpunct}
{\mcitedefaultendpunct}{\mcitedefaultseppunct}\relax
\EndOfBibitem
\bibitem[Ruddigkeit \latin{et~al.}(2012)Ruddigkeit, Van~Deursen, Blum, and
  Reymond]{ruddigkeit2012enumeration}
Ruddigkeit,~L.; Van~Deursen,~R.; Blum,~L.~C.; Reymond,~J.-L. Enumeration of 166
  billion organic small molecules in the chemical universe database GDB-17.
  \emph{Journal of chemical information and modeling} \textbf{2012}, \emph{52},
  2864--2875\relax
\mciteBstWouldAddEndPuncttrue
\mciteSetBstMidEndSepPunct{\mcitedefaultmidpunct}
{\mcitedefaultendpunct}{\mcitedefaultseppunct}\relax
\EndOfBibitem
\bibitem[Ramakrishnan \latin{et~al.}(2014)Ramakrishnan, Dral, Rupp, and
  Von~Lilienfeld]{qm9}
Ramakrishnan,~R.; Dral,~P.~O.; Rupp,~M.; Von~Lilienfeld,~O.~A. Quantum
  chemistry structures and properties of 134 kilo molecules. \emph{Scientific
  data} \textbf{2014}, \emph{1}, 140022\relax
\mciteBstWouldAddEndPuncttrue
\mciteSetBstMidEndSepPunct{\mcitedefaultmidpunct}
{\mcitedefaultendpunct}{\mcitedefaultseppunct}\relax
\EndOfBibitem
\bibitem[Elton \latin{et~al.}(2019)Elton, Boukouvalas, Fuge, and
  Chung]{elton2019deep}
Elton,~D.~C.; Boukouvalas,~Z.; Fuge,~M.~D.; Chung,~P.~W. Deep learning for
  molecular generation and optimization-a review of the state of the art.
  \emph{arXiv preprint arXiv:1903.04388} \textbf{2019}, \relax
\mciteBstWouldAddEndPunctfalse
\mciteSetBstMidEndSepPunct{\mcitedefaultmidpunct}
{}{\mcitedefaultseppunct}\relax
\EndOfBibitem
\bibitem[Li \latin{et~al.}(2018)Li, Zhang, and Liu]{li2018multi}
Li,~Y.; Zhang,~L.; Liu,~Z. Multi-objective de novo drug design with conditional
  graph generative model. \emph{Journal of cheminformatics} \textbf{2018},
  \emph{10}, 33\relax
\mciteBstWouldAddEndPuncttrue
\mciteSetBstMidEndSepPunct{\mcitedefaultmidpunct}
{\mcitedefaultendpunct}{\mcitedefaultseppunct}\relax
\EndOfBibitem
\bibitem[Salakhutdinov(2015)]{salakhutdinov2015learning}
Salakhutdinov,~R. Learning deep generative models. \emph{Annual Review of
  Statistics and Its Application} \textbf{2015}, \emph{2}, 361--385\relax
\mciteBstWouldAddEndPuncttrue
\mciteSetBstMidEndSepPunct{\mcitedefaultmidpunct}
{\mcitedefaultendpunct}{\mcitedefaultseppunct}\relax
\EndOfBibitem
\bibitem[De~Cao and Kipf(2018)De~Cao, and Kipf]{de2018molgan}
De~Cao,~N.; Kipf,~T. MolGAN: An implicit generative model for small molecular
  graphs. \emph{arXiv preprint arXiv:1805.11973} \textbf{2018}, \relax
\mciteBstWouldAddEndPunctfalse
\mciteSetBstMidEndSepPunct{\mcitedefaultmidpunct}
{}{\mcitedefaultseppunct}\relax
\EndOfBibitem
\bibitem[You \latin{et~al.}(2018)You, Liu, Ying, Pande, and
  Leskovec]{you2018graph}
You,~J.; Liu,~B.; Ying,~Z.; Pande,~V.; Leskovec,~J. Graph convolutional policy
  network for goal-directed molecular graph generation. Advances in Neural
  Information Processing Systems. 2018; pp 6410--6421\relax
\mciteBstWouldAddEndPuncttrue
\mciteSetBstMidEndSepPunct{\mcitedefaultmidpunct}
{\mcitedefaultendpunct}{\mcitedefaultseppunct}\relax
\EndOfBibitem
\bibitem[G{\'o}mez-Bombarelli \latin{et~al.}(2018)G{\'o}mez-Bombarelli, Wei,
  Duvenaud, Hern{\'a}ndez-Lobato, S{\'a}nchez-Lengeling, Sheberla,
  Aguilera-Iparraguirre, Hirzel, Adams, and Aspuru-Guzik]{gomez2018automatic}
G{\'o}mez-Bombarelli,~R.; Wei,~J.~N.; Duvenaud,~D.;
  Hern{\'a}ndez-Lobato,~J.~M.; S{\'a}nchez-Lengeling,~B.; Sheberla,~D.;
  Aguilera-Iparraguirre,~J.; Hirzel,~T.~D.; Adams,~R.~P.; Aspuru-Guzik,~A.
  Automatic chemical design using a data-driven continuous representation of
  molecules. \emph{ACS central science} \textbf{2018}, \emph{4}, 268--276\relax
\mciteBstWouldAddEndPuncttrue
\mciteSetBstMidEndSepPunct{\mcitedefaultmidpunct}
{\mcitedefaultendpunct}{\mcitedefaultseppunct}\relax
\EndOfBibitem
\bibitem[Jaeger \latin{et~al.}(2018)Jaeger, Fulle, and Turk]{Jaeger2018}
Jaeger,~S.; Fulle,~S.; Turk,~S. Mol2vec: Unsupervised Machine Learning Approach
  with Chemical Intuition. \emph{Journal of Chemical Information and Modeling}
  \textbf{2018}, \emph{58}, 27--35, PMID: 29268609\relax
\mciteBstWouldAddEndPuncttrue
\mciteSetBstMidEndSepPunct{\mcitedefaultmidpunct}
{\mcitedefaultendpunct}{\mcitedefaultseppunct}\relax
\EndOfBibitem
\bibitem[Zheng \latin{et~al.}(2019)Zheng, Yan, Yang, and
  Xu]{zheng2019identifying}
Zheng,~S.; Yan,~X.; Yang,~Y.; Xu,~J. Identifying Structure--Property
  Relationships through SMILES Syntax Analysis with Self-Attention Mechanism.
  \emph{Journal of chemical information and modeling} \textbf{2019}, \emph{59},
  914--923\relax
\mciteBstWouldAddEndPuncttrue
\mciteSetBstMidEndSepPunct{\mcitedefaultmidpunct}
{\mcitedefaultendpunct}{\mcitedefaultseppunct}\relax
\EndOfBibitem
\bibitem[Mater and Coote(2019)Mater, and Coote]{mater2019deep}
Mater,~A.~C.; Coote,~M.~L. Deep Learning in Chemistry. \emph{Journal of
  Chemical Information and Modeling} \textbf{2019}, \relax
\mciteBstWouldAddEndPunctfalse
\mciteSetBstMidEndSepPunct{\mcitedefaultmidpunct}
{}{\mcitedefaultseppunct}\relax
\EndOfBibitem
\bibitem[Bengio \latin{et~al.}(2003)Bengio, Ducharme, Vincent, and
  Janvin]{Bengio_2003}
Bengio,~Y.; Ducharme,~R.; Vincent,~P.; Janvin,~C. A Neural Probabilistic
  Language Model. \emph{J. Mach. Learn. Res.} \textbf{2003}, \emph{3},
  1137--1155\relax
\mciteBstWouldAddEndPuncttrue
\mciteSetBstMidEndSepPunct{\mcitedefaultmidpunct}
{\mcitedefaultendpunct}{\mcitedefaultseppunct}\relax
\EndOfBibitem
\bibitem[Devlin \latin{et~al.}(2018)Devlin, Chang, Lee, and Toutanova]{BERT}
Devlin,~J.; Chang,~M.-W.; Lee,~K.; Toutanova,~K. Bert: Pre-training of deep
  bidirectional transformers for language understanding. \emph{arXiv preprint
  arXiv:1810.04805} \textbf{2018}, \relax
\mciteBstWouldAddEndPunctfalse
\mciteSetBstMidEndSepPunct{\mcitedefaultmidpunct}
{}{\mcitedefaultseppunct}\relax
\EndOfBibitem
\bibitem[Vincent \latin{et~al.}(2008)Vincent, Larochelle, Bengio, and
  Manzagol]{Vincent2008}
Vincent,~P.; Larochelle,~H.; Bengio,~Y.; Manzagol,~P.-A. Extracting and
  Composing Robust Features with Denoising Autoencoders. Proceedings of the
  25th International Conference on Machine Learning. New York, NY, USA, 2008;
  pp 1096--1103\relax
\mciteBstWouldAddEndPuncttrue
\mciteSetBstMidEndSepPunct{\mcitedefaultmidpunct}
{\mcitedefaultendpunct}{\mcitedefaultseppunct}\relax
\EndOfBibitem
\bibitem[Gerratt \latin{et~al.}(1997)Gerratt, Cooper, Karadakov, and
  Raimondi]{gerratt1997modern}
Gerratt,~J.; Cooper,~D.; Karadakov,~P.~a.; Raimondi,~M. Modern valence bond
  theory. \emph{Chemical Society Reviews} \textbf{1997}, \emph{26},
  87--100\relax
\mciteBstWouldAddEndPuncttrue
\mciteSetBstMidEndSepPunct{\mcitedefaultmidpunct}
{\mcitedefaultendpunct}{\mcitedefaultseppunct}\relax
\EndOfBibitem
\bibitem[G{\'o}mez-Bombarelli \latin{et~al.}(2018)G{\'o}mez-Bombarelli, Wei,
  Duvenaud, Hern{\'a}ndez-Lobato, S{\'a}nchez-Lengeling, Sheberla,
  Aguilera-Iparraguirre, Hirzel, Adams, and Aspuru-Guzik]{zinc}
G{\'o}mez-Bombarelli,~R.; Wei,~J.~N.; Duvenaud,~D.;
  Hern{\'a}ndez-Lobato,~J.~M.; S{\'a}nchez-Lengeling,~B.; Sheberla,~D.;
  Aguilera-Iparraguirre,~J.; Hirzel,~T.~D.; Adams,~R.~P.; Aspuru-Guzik,~A.
  Automatic chemical design using a data-driven continuous representation of
  molecules. \emph{ACS central science} \textbf{2018}, \emph{4}, 268--276\relax
\mciteBstWouldAddEndPuncttrue
\mciteSetBstMidEndSepPunct{\mcitedefaultmidpunct}
{\mcitedefaultendpunct}{\mcitedefaultseppunct}\relax
\EndOfBibitem
\bibitem[Irwin \latin{et~al.}(2012)Irwin, Sterling, Mysinger, Bolstad, and
  Coleman]{zincdb}
Irwin,~J.~J.; Sterling,~T.; Mysinger,~M.~M.; Bolstad,~E.~S.; Coleman,~R.~G.
  ZINC: a free tool to discover chemistry for biology. \emph{Journal of
  chemical information and modeling} \textbf{2012}, \emph{52}, 1757--1768\relax
\mciteBstWouldAddEndPuncttrue
\mciteSetBstMidEndSepPunct{\mcitedefaultmidpunct}
{\mcitedefaultendpunct}{\mcitedefaultseppunct}\relax
\EndOfBibitem
\bibitem[Bengio \latin{et~al.}(2003)Bengio, Ducharme, Vincent, and
  Janvin]{Bengio2003}
Bengio,~Y.; Ducharme,~R.; Vincent,~P.; Janvin,~C. A Neural Probabilistic
  Language Model. \emph{J. Mach. Learn. Res.} \textbf{2003}, \emph{3},
  1137--1155\relax
\mciteBstWouldAddEndPuncttrue
\mciteSetBstMidEndSepPunct{\mcitedefaultmidpunct}
{\mcitedefaultendpunct}{\mcitedefaultseppunct}\relax
\EndOfBibitem
\bibitem[Jurafsky and Martin(2009)Jurafsky, and Martin]{Jurafsky2009}
Jurafsky,~D.; Martin,~J.~H. \emph{Speech and Language Processing (2Nd
  Edition)}; Prentice-Hall, Inc.: Upper Saddle River, NJ, USA, 2009\relax
\mciteBstWouldAddEndPuncttrue
\mciteSetBstMidEndSepPunct{\mcitedefaultmidpunct}
{\mcitedefaultendpunct}{\mcitedefaultseppunct}\relax
\EndOfBibitem
\bibitem[Mikolov \latin{et~al.}(2013)Mikolov, Sutskever, Chen, Corrado, and
  Dean]{Mikolov_2013}
Mikolov,~T.; Sutskever,~I.; Chen,~K.; Corrado,~G.; Dean,~J. Distributed
  Representations of Words and Phrases and Their Compositionality. Proceedings
  of the 26th International Conference on Neural Information Processing Systems
  - Volume 2. USA, 2013; pp 3111--3119\relax
\mciteBstWouldAddEndPuncttrue
\mciteSetBstMidEndSepPunct{\mcitedefaultmidpunct}
{\mcitedefaultendpunct}{\mcitedefaultseppunct}\relax
\EndOfBibitem
\bibitem[Mikolov \latin{et~al.}(2010)Mikolov, Karafiát, Burget, Cernocký, and
  Khudanpur]{MikolovKBCK10}
Mikolov,~T.; Karafiát,~M.; Burget,~L.; Cernocký,~J.; Khudanpur,~S. Recurrent
  neural network based language model. INTERSPEECH. 2010; pp 1045--1048\relax
\mciteBstWouldAddEndPuncttrue
\mciteSetBstMidEndSepPunct{\mcitedefaultmidpunct}
{\mcitedefaultendpunct}{\mcitedefaultseppunct}\relax
\EndOfBibitem
\bibitem[Zaremba \latin{et~al.}(2014)Zaremba, Sutskever, and
  Vinyals]{ZarembaSV14}
Zaremba,~W.; Sutskever,~I.; Vinyals,~O. Recurrent Neural Network
  Regularization. \emph{CoRR} \textbf{2014}, \emph{abs/1409.2329}\relax
\mciteBstWouldAddEndPuncttrue
\mciteSetBstMidEndSepPunct{\mcitedefaultmidpunct}
{\mcitedefaultendpunct}{\mcitedefaultseppunct}\relax
\EndOfBibitem
\bibitem[Merity \latin{et~al.}(2018)Merity, Keskar, and Socher]{MerityKS18}
Merity,~S.; Keskar,~N.~S.; Socher,~R. Regularizing and Optimizing {LSTM}
  Language Models. 6th International Conference on Learning Representations,
  {ICLR} 2018, Vancouver, BC, Canada, April 30 - May 3, 2018, Conference Track
  Proceedings. 2018\relax
\mciteBstWouldAddEndPuncttrue
\mciteSetBstMidEndSepPunct{\mcitedefaultmidpunct}
{\mcitedefaultendpunct}{\mcitedefaultseppunct}\relax
\EndOfBibitem
\bibitem[Hinton and Salakhutdinov(2006)Hinton, and
  Salakhutdinov]{HintonSalakhutdinov2006b}
Hinton,~G.~E.; Salakhutdinov,~R.~R. Reducing the dimensionality of data with
  neural networks. \emph{Science} \textbf{2006}, \emph{313}, 504--507\relax
\mciteBstWouldAddEndPuncttrue
\mciteSetBstMidEndSepPunct{\mcitedefaultmidpunct}
{\mcitedefaultendpunct}{\mcitedefaultseppunct}\relax
\EndOfBibitem
\bibitem[Hansen \latin{et~al.}(2015)Hansen, Biegler, Ramakrishnan, Pronobis,
  Von~Lilienfeld, M\"{u}ller, and Tkatchenko]{hansen2015machine}
Hansen,~K.; Biegler,~F.; Ramakrishnan,~R.; Pronobis,~W.; Von~Lilienfeld,~O.~A.;
  M\"{u}ller,~K.-R.; Tkatchenko,~A. Machine learning predictions of molecular
  properties: Accurate many-body potentials and nonlocality in chemical space.
  \emph{The journal of physical chemistry letters} \textbf{2015}, \emph{6},
  2326--2331\relax
\mciteBstWouldAddEndPuncttrue
\mciteSetBstMidEndSepPunct{\mcitedefaultmidpunct}
{\mcitedefaultendpunct}{\mcitedefaultseppunct}\relax
\EndOfBibitem
\bibitem[Hansen \latin{et~al.}(2019)Hansen, Torres, Jennings, Wang, Boes,
  Mamun, and Bligaard]{hansen2019atomistic}
Hansen,~M.~H.; Torres,~J. A.~G.; Jennings,~P.~C.; Wang,~Z.; Boes,~J.~R.;
  Mamun,~O.~G.; Bligaard,~T. An Atomistic Machine Learning Package for Surface
  Science and Catalysis. \emph{arXiv preprint arXiv:1904.00904} \textbf{2019},
  \relax
\mciteBstWouldAddEndPunctfalse
\mciteSetBstMidEndSepPunct{\mcitedefaultmidpunct}
{}{\mcitedefaultseppunct}\relax
\EndOfBibitem
\bibitem[Mikolov \latin{et~al.}(2013)Mikolov, Chen, Corrado, and Dean]{MCCD}
Mikolov,~T.; Chen,~K.; Corrado,~G.; Dean,~J. Efficient Estimation of Word
  Representations in Vector Space. 1st International Conference on Learning
  Representations, {ICLR} 2013, Scottsdale, Arizona, USA, May 2-4, 2013,
  Workshop Track Proceedings. 2013\relax
\mciteBstWouldAddEndPuncttrue
\mciteSetBstMidEndSepPunct{\mcitedefaultmidpunct}
{\mcitedefaultendpunct}{\mcitedefaultseppunct}\relax
\EndOfBibitem
\bibitem[Vaswani \latin{et~al.}(2017)Vaswani, Shazeer, Parmar, Uszkoreit,
  Jones, Gomez, Kaiser, and Polosukhin]{Attention}
Vaswani,~A.; Shazeer,~N.; Parmar,~N.; Uszkoreit,~J.; Jones,~L.; Gomez,~A.~N.;
  Kaiser,~{\L}.; Polosukhin,~I. Attention is all you need. Advances in neural
  information processing systems. 2017; pp 5998--6008\relax
\mciteBstWouldAddEndPuncttrue
\mciteSetBstMidEndSepPunct{\mcitedefaultmidpunct}
{\mcitedefaultendpunct}{\mcitedefaultseppunct}\relax
\EndOfBibitem
\bibitem[Bahdanau \latin{et~al.}(2014)Bahdanau, Cho, and
  Bengio]{bahdanau2014neural}
Bahdanau,~D.; Cho,~K.; Bengio,~Y. Neural machine translation by jointly
  learning to align and translate. \emph{arXiv preprint arXiv:1409.0473}
  \textbf{2014}, \relax
\mciteBstWouldAddEndPunctfalse
\mciteSetBstMidEndSepPunct{\mcitedefaultmidpunct}
{}{\mcitedefaultseppunct}\relax
\EndOfBibitem
\bibitem[Shaw \latin{et~al.}(2018)Shaw, Uszkoreit, and Vaswani]{relativepos}
Shaw,~P.; Uszkoreit,~J.; Vaswani,~A. Self-Attention with Relative Position
  Representations. \emph{CoRR} \textbf{2018}, \emph{abs/1803.02155}\relax
\mciteBstWouldAddEndPuncttrue
\mciteSetBstMidEndSepPunct{\mcitedefaultmidpunct}
{\mcitedefaultendpunct}{\mcitedefaultseppunct}\relax
\EndOfBibitem
\bibitem[Ba \latin{et~al.}(2016)Ba, Kiros, and Hinton]{layernorm}
Ba,~L.~J.; Kiros,~R.; Hinton,~G.~E. Layer Normalization. \emph{CoRR}
  \textbf{2016}, \emph{abs/1607.06450}\relax
\mciteBstWouldAddEndPuncttrue
\mciteSetBstMidEndSepPunct{\mcitedefaultmidpunct}
{\mcitedefaultendpunct}{\mcitedefaultseppunct}\relax
\EndOfBibitem
\bibitem[He \latin{et~al.}(2015)He, Zhang, Ren, and Sun]{resnet}
He,~K.; Zhang,~X.; Ren,~S.; Sun,~J. Deep Residual Learning for Image
  Recognition. \emph{CoRR} \textbf{2015}, \emph{abs/1512.03385}\relax
\mciteBstWouldAddEndPuncttrue
\mciteSetBstMidEndSepPunct{\mcitedefaultmidpunct}
{\mcitedefaultendpunct}{\mcitedefaultseppunct}\relax
\EndOfBibitem
\bibitem[Srivastava \latin{et~al.}(2015)Srivastava, Greff, and
  Schmidhuber]{highway}
Srivastava,~R.~K.; Greff,~K.; Schmidhuber,~J. Highway Networks. \emph{CoRR}
  \textbf{2015}, \emph{abs/1505.00387}\relax
\mciteBstWouldAddEndPuncttrue
\mciteSetBstMidEndSepPunct{\mcitedefaultmidpunct}
{\mcitedefaultendpunct}{\mcitedefaultseppunct}\relax
\EndOfBibitem
\bibitem[Glorot \latin{et~al.}(2011)Glorot, Bordes, and Bengio]{rectify}
Glorot,~X.; Bordes,~A.; Bengio,~Y. Deep Sparse Rectifier Neural Networks.
  Proceedings of the Fourteenth International Conference on Artificial
  Intelligence and Statistics. Fort Lauderdale, FL, USA, 2011; pp
  315--323\relax
\mciteBstWouldAddEndPuncttrue
\mciteSetBstMidEndSepPunct{\mcitedefaultmidpunct}
{\mcitedefaultendpunct}{\mcitedefaultseppunct}\relax
\EndOfBibitem
\bibitem[Luong \latin{et~al.}(2015)Luong, Pham, and
  Manning]{luong2015effective}
Luong,~M.-T.; Pham,~H.; Manning,~C.~D. Effective approaches to attention-based
  neural machine translation. \emph{arXiv preprint arXiv:1508.04025}
  \textbf{2015}, \relax
\mciteBstWouldAddEndPunctfalse
\mciteSetBstMidEndSepPunct{\mcitedefaultmidpunct}
{}{\mcitedefaultseppunct}\relax
\EndOfBibitem
\bibitem[Weininger(1988)]{SMILES}
Weininger,~D. SMILES, a chemical language and information system. 1.
  Introduction to methodology and encoding rules. \emph{Journal of Chemical
  Information and Computer Sciences} \textbf{1988}, \emph{28}, 31--36\relax
\mciteBstWouldAddEndPuncttrue
\mciteSetBstMidEndSepPunct{\mcitedefaultmidpunct}
{\mcitedefaultendpunct}{\mcitedefaultseppunct}\relax
\EndOfBibitem
\bibitem[Landrum()]{rdkit}
Landrum,~G. RDKit: Open-source cheminformatics.
  \url{http://www.rdkit.org}\relax
\mciteBstWouldAddEndPuncttrue
\mciteSetBstMidEndSepPunct{\mcitedefaultmidpunct}
{\mcitedefaultendpunct}{\mcitedefaultseppunct}\relax
\EndOfBibitem
\bibitem[Sutton and Barto(2018)Sutton, and Barto]{sutton2018reinforcement}
Sutton,~R.~S.; Barto,~A.~G. \emph{Reinforcement learning: An introduction};
  2018\relax
\mciteBstWouldAddEndPuncttrue
\mciteSetBstMidEndSepPunct{\mcitedefaultmidpunct}
{\mcitedefaultendpunct}{\mcitedefaultseppunct}\relax
\EndOfBibitem
\bibitem[Kingma and Ba(2014)Kingma, and Ba]{ADAM}
Kingma,~D.~P.; Ba,~J. Adam: A method for stochastic optimization. \emph{arXiv
  preprint arXiv:1412.6980} \textbf{2014}, \relax
\mciteBstWouldAddEndPunctfalse
\mciteSetBstMidEndSepPunct{\mcitedefaultmidpunct}
{}{\mcitedefaultseppunct}\relax
\EndOfBibitem
\bibitem[Paszke \latin{et~al.}(2017)Paszke, Gross, Chintala, Chanan, Yang,
  DeVito, Lin, Desmaison, Antiga, and Lerer]{pytorch}
Paszke,~A.; Gross,~S.; Chintala,~S.; Chanan,~G.; Yang,~E.; DeVito,~Z.; Lin,~Z.;
  Desmaison,~A.; Antiga,~L.; Lerer,~A. Automatic Differentiation in {PyTorch}.
  NIPS Autodiff Workshop. 2017\relax
\mciteBstWouldAddEndPuncttrue
\mciteSetBstMidEndSepPunct{\mcitedefaultmidpunct}
{\mcitedefaultendpunct}{\mcitedefaultseppunct}\relax
\EndOfBibitem
\bibitem[Yutaka(2007)]{sasaki2007truth}
Yutaka,~S. The truth of the F-measure. \emph{Teach Tutor mater} \textbf{2007},
  \emph{1}, 1--5\relax
\mciteBstWouldAddEndPuncttrue
\mciteSetBstMidEndSepPunct{\mcitedefaultmidpunct}
{\mcitedefaultendpunct}{\mcitedefaultseppunct}\relax
\EndOfBibitem
\bibitem[Buda \latin{et~al.}(2018)Buda, Maki, and Mazurowski]{Buda2018ASS}
Buda,~M.; Maki,~A.; Mazurowski,~M.~A. A systematic study of the class imbalance
  problem in convolutional neural networks. \emph{Neural networks : the
  official journal of the International Neural Network Society} \textbf{2018},
  \emph{106}, 249--259\relax
\mciteBstWouldAddEndPuncttrue
\mciteSetBstMidEndSepPunct{\mcitedefaultmidpunct}
{\mcitedefaultendpunct}{\mcitedefaultseppunct}\relax
\EndOfBibitem
\bibitem[Peters \latin{et~al.}(2018)Peters, Neumann, Iyyer, Gardner, Clark,
  Lee, and Zettlemoyer]{ELMO}
Peters,~M.~E.; Neumann,~M.; Iyyer,~M.; Gardner,~M.; Clark,~C.; Lee,~K.;
  Zettlemoyer,~L. Deep contextualized word representations. \emph{arXiv
  preprint arXiv:1802.05365} \textbf{2018}, \relax
\mciteBstWouldAddEndPunctfalse
\mciteSetBstMidEndSepPunct{\mcitedefaultmidpunct}
{}{\mcitedefaultseppunct}\relax
\EndOfBibitem
\bibitem[Liu \latin{et~al.}(2019)Liu, He, Chen, and Gao]{mtdnn}
Liu,~X.; He,~P.; Chen,~W.; Gao,~J. Multi-Task Deep Neural Networks for Natural
  Language Understanding. \emph{CoRR} \textbf{2019},
  \emph{abs/1901.11504}\relax
\mciteBstWouldAddEndPuncttrue
\mciteSetBstMidEndSepPunct{\mcitedefaultmidpunct}
{\mcitedefaultendpunct}{\mcitedefaultseppunct}\relax
\EndOfBibitem
\end{mcitethebibliography}
\newpage

\appendix
\renewcommand{\thetable}{S.\arabic{table}}
\renewcommand{\thefigure}{S.\arabic{figure}}
\setcounter{figure}{0}
\setcounter{table}{0}
\section{Appendix}

\maketitle

\begin{table}[H]
    \centering
    \begin{tabular}{c|l}
        Variable & Description\\ \hline
        $G$ & Graph, defined as a set of nodes and edges (V,E) \\
        $V$ & Set of nodes (atoms) in the graph\\
        $V_\texttt{subset}$ & set of masked atoms. $|V_\texttt{subset}|=n_{corrupt}$\\
        $n_{corrupt}$ & Number of atoms corrupted per molecule\\
        $E$ & Adjacency matrix ($E_{ij}\in \{0,1,2,3\}$) \\
        $\widetilde{G}$ & Corrupted graph, with $V_\texttt{subset}$ replaced with a \texttt{<MASK>} token. \\
        $v_i$ & I'th atom in the graph. $v_i=(a,i)$\\
        $a$ & Element of an atom. $a\in \{H,C,O,N,F,P,S,Cl,Br,I\}$\\
        $x$ & Token represented as a vector. $x\in\mathbb{R}^d$\\
        $\texttt{embedding}(x)$ & Embedding of a token. $\texttt{embedding}(x)\in \mathbb{R}^{d_{emb}}$\\
        $z_\theta$ & Intermediate representation in the BoW model. $z_\theta\in \mathbb{R}^{d_{emb}}$ \\
        $h_\theta$ & Hidden representation in the BoW model. $h_\theta \in \mathbb{R}^{d_{nn}}$\\
        $h^0$ & Embedding of the nodes in the graph. $h^0\in \mathbb{R}^{|V|\times d_{emb}}$\\
        $z^l, h^l$ & Intermediate and hidden representation of the l'th layer. $z^l, h^l \in \mathbb{R}^{|V|\times d_{transform}}$\\
        $e^V, e^K$ & Embeddings of the bonds. $e^V, e^K\in \mathbb{R}^{|V|\times|V|\times d_{transform}}$\\
        $W^Q,W^K,W^V$ & Trainable weights. $W^Q,W^K,W^V\in \mathbb{R}^{d_{transform}\times d_{transform}}$\\
        $\alpha_{ij}$ & Attention weights between i'th and j'th atom. $\alpha_{ij}\in [0,1], \sum_j \alpha_{ij}=1$\\
        $\phi_{ij}$ & Unnormalized attention weights. $\phi_{ij}\in \mathbb{R}$ \\
        $W_{multi}$ & Trainable weight. $W_{multi}\in \mathbb{R}^{(d_{transform}\cdot k)\times d_{transform}}$\\
        $\texttt{C\_i}$ & I'th head of attention function for an atom. $\texttt{C\_i}\in \mathbb{R}^{d_{transform}}$

        \end{tabular}
    \caption{Describtion of variables used.}
    \label{tab:my_label}
\end{table}

\begin{table}[H]
    \centering
    \begin{tabular}{|c|c|c|}
    \hline
        Model & Training time (min) & Dataset \\\hline
        \texttt{binary-transformer} & 110 & QM9 \\ \hline
        \texttt{binary-transformer} & 482 & ZINC \\ \hline
        \texttt{bond-transformer} & 112 & QM9 \\ \hline
        \texttt{bond-transformer} & 484 & ZINC \\ \hline
        \texttt{bag-of-atoms} & 71 & QM9 \\ \hline
        \texttt{bag-of-atoms} & 158 & ZINC \\ \hline
        \texttt{bag-of-neighbors} & 72 & QM9 \\ \hline
        \texttt{bag-of-neighbors} & 144 & ZINC \\ \hline

    \end{tabular}
    \caption{Training time of our different models, on the QM9 and ZINC datasets.}
    \label{tab:my_label}
\end{table}

\section{Epsilon-greedy}
\begin{figure}[H]
    \centering
    \begin{subfigure}[b]{0.45\textwidth}
        \includegraphics[width=\textwidth]{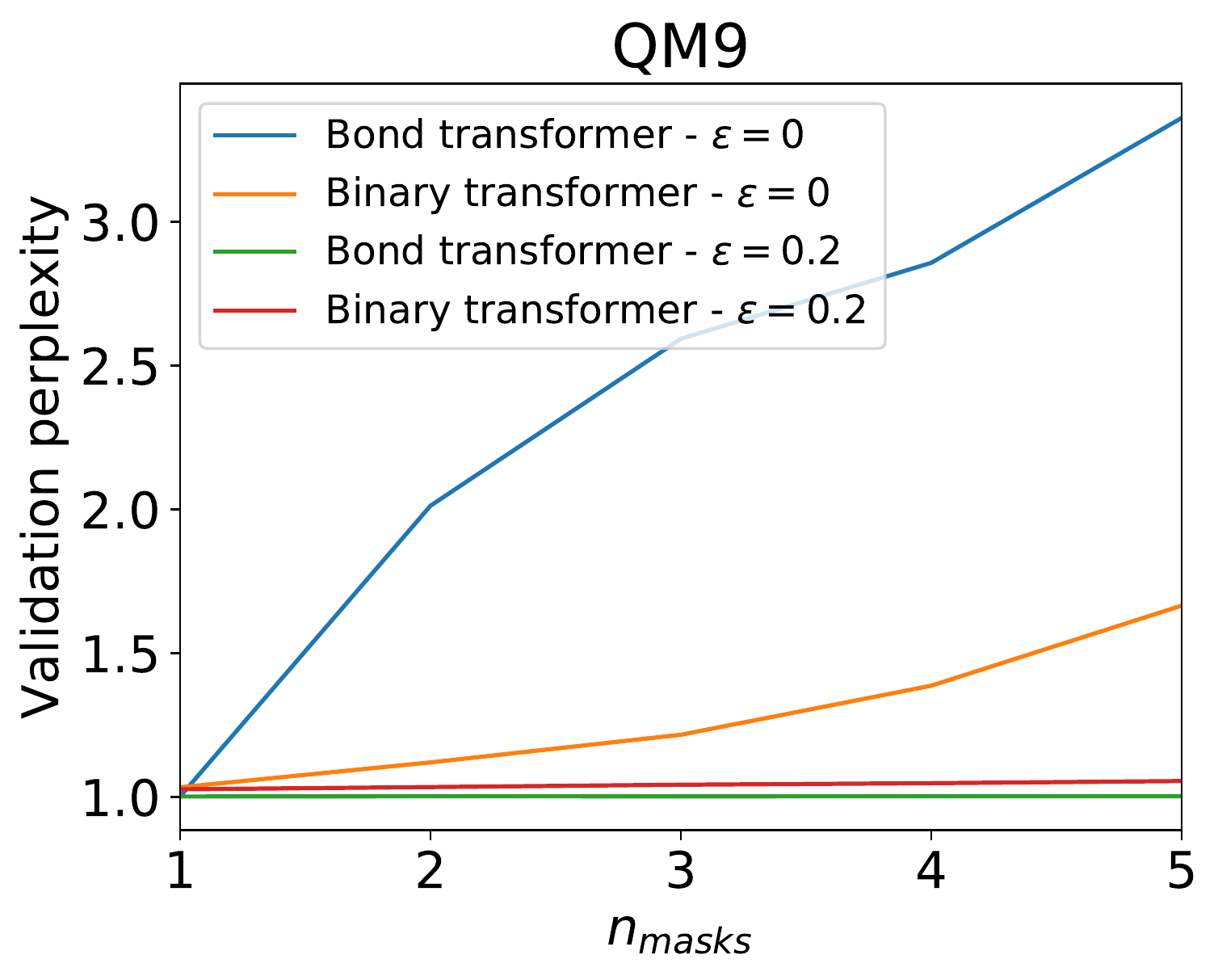}
    \end{subfigure}
    \begin{subfigure}[b]{0.45\textwidth}
        \includegraphics[width=\textwidth]{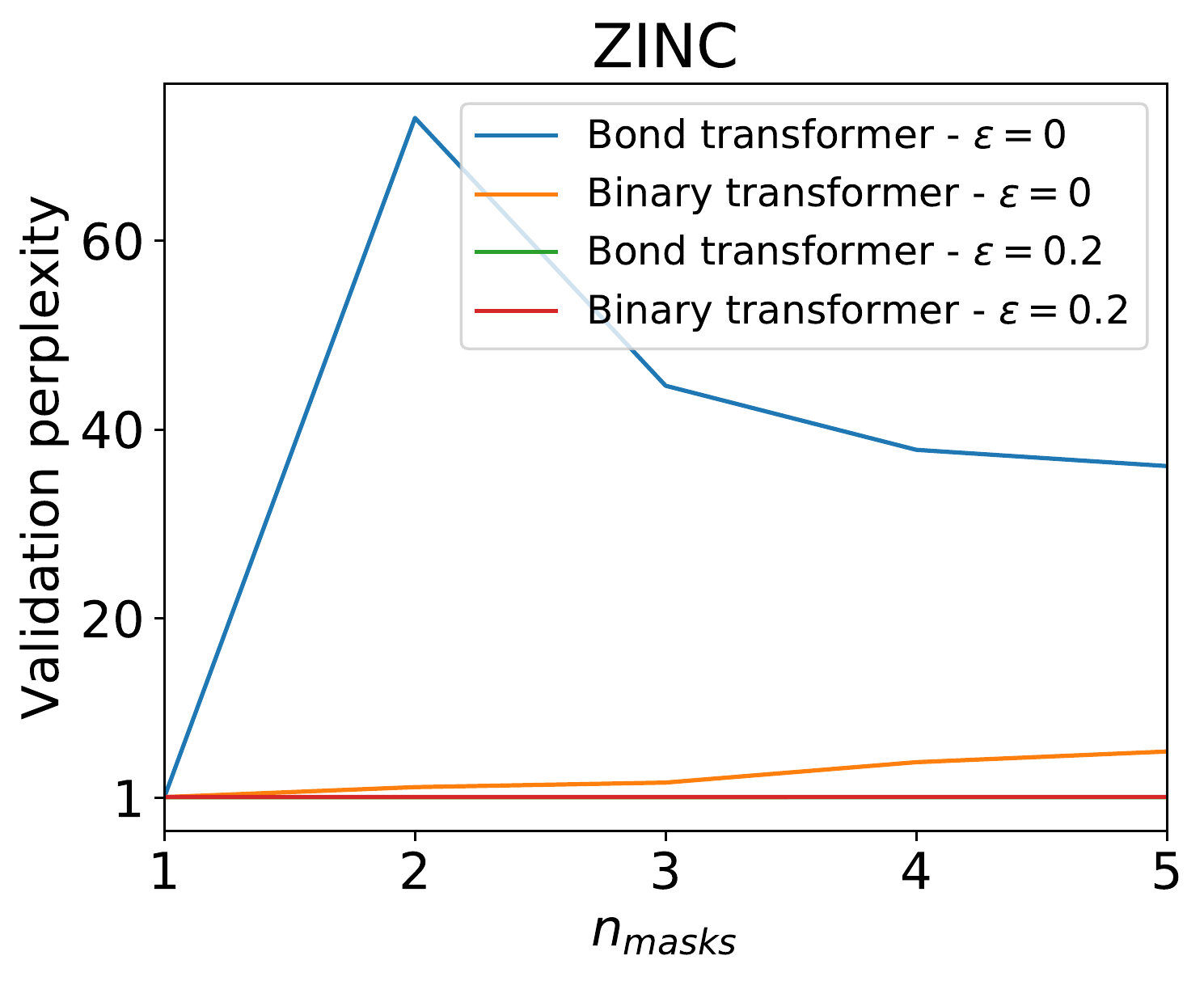}
    \end{subfigure}
    \caption{Validation perplexity of binary and bond transformer -- with and without $\epsilon$-greedy masking strategy -- with different number of masked atoms. (a) is on the QM9 dataset and (b) is on the ZINC dataset }
    \label{fig:epsilon_greedy}
\end{figure}

\section{K-smoothing}
\begin{figure}[H]
    \centering
    \includegraphics[width=\textwidth]{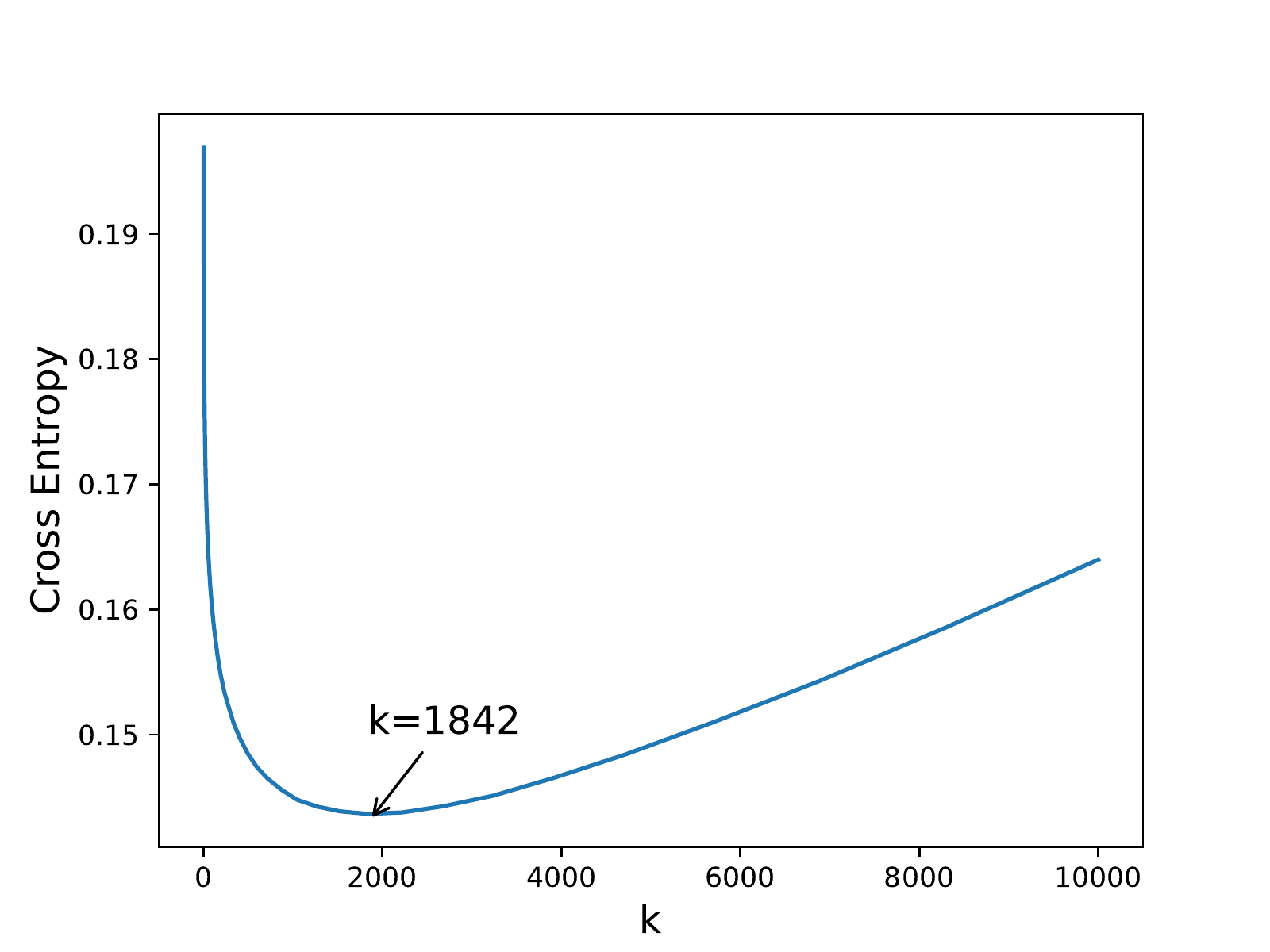}
    \caption{Cross entropy as a function of k-smoothing evaluated on the ZINC validation dataset.}
    \label{fig:my_label}
\end{figure}

\section{Graph Attention}

\begin{figure}[H]
    \centering
    \begin{subfigure}[b]{0.45\textwidth}
        \includegraphics[width=\textwidth]{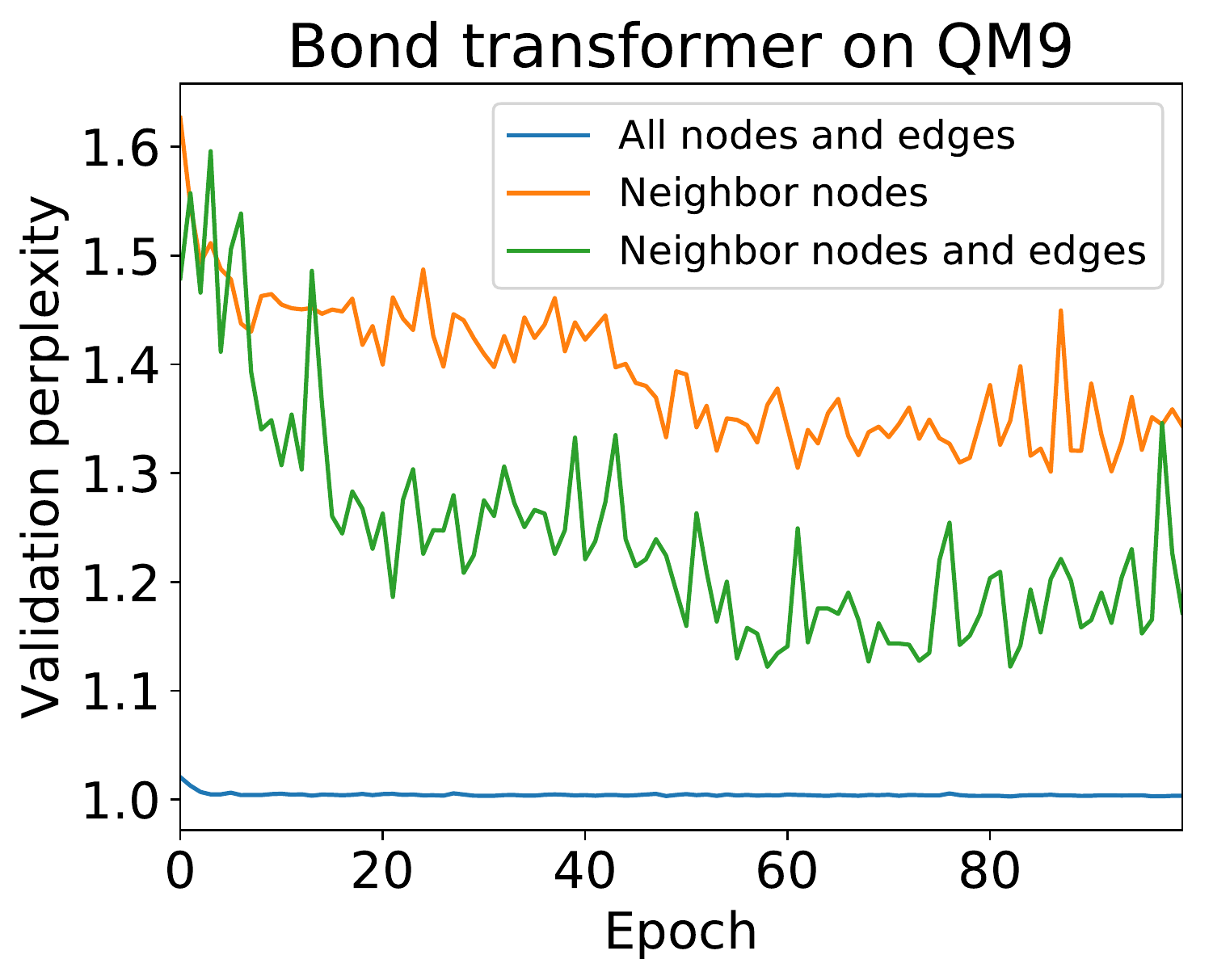}
    \end{subfigure}
    \begin{subfigure}[b]{0.45\textwidth}
        \includegraphics[width=\textwidth]{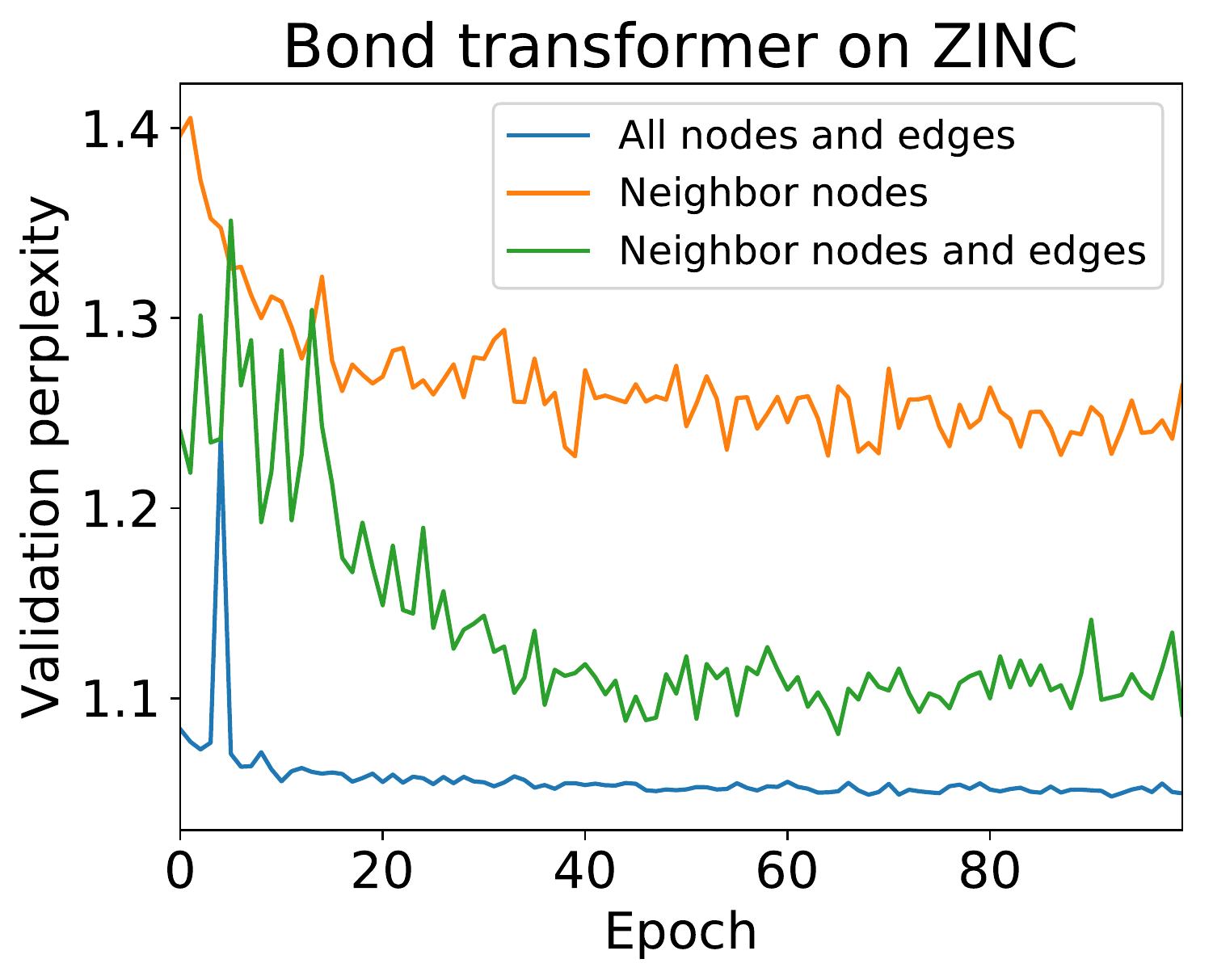}
    \end{subfigure}
        \begin{subfigure}[b]{0.45\textwidth}
        \includegraphics[width=\textwidth]{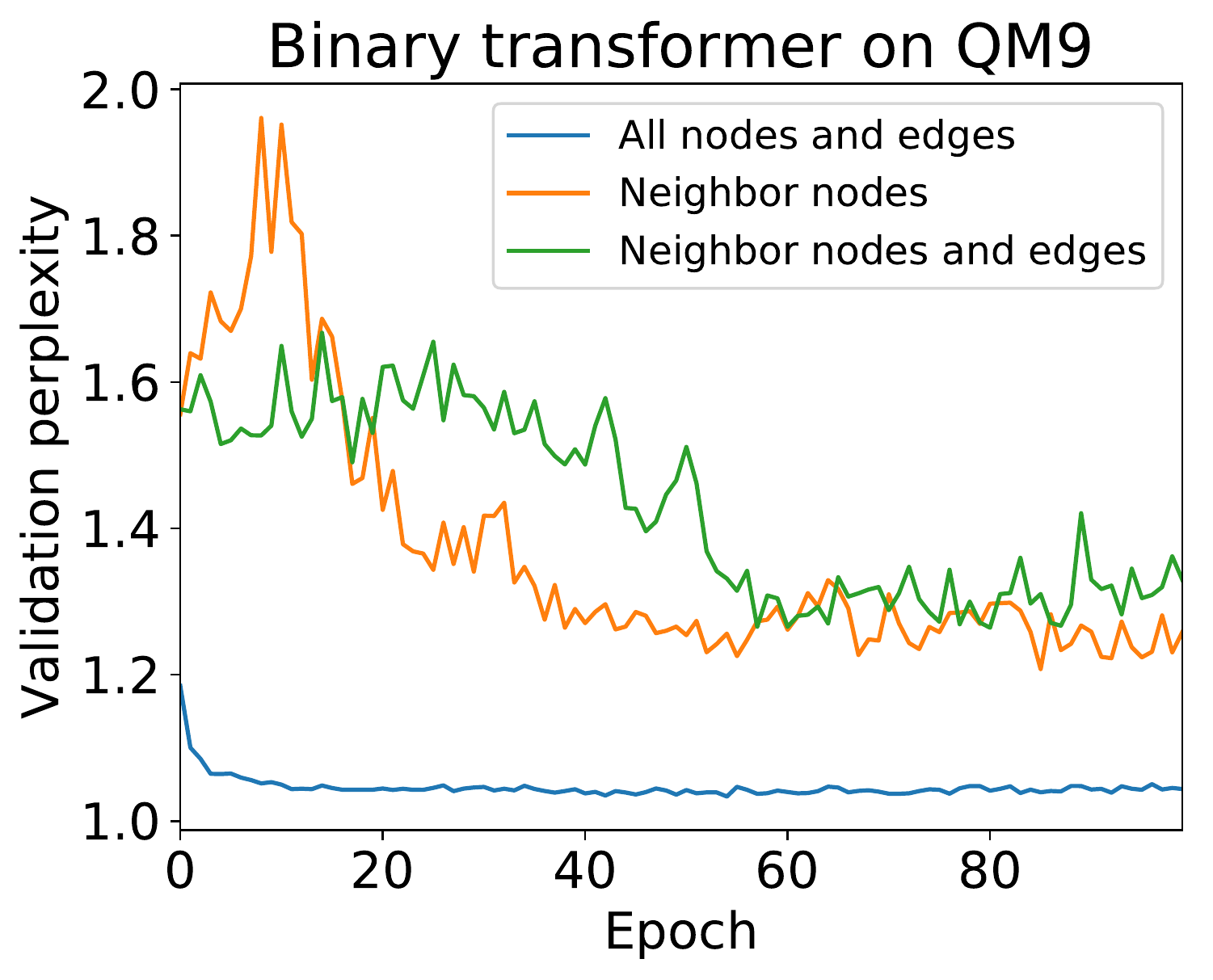}
    \end{subfigure}
    \begin{subfigure}[b]{0.45\textwidth}
        \includegraphics[width=\textwidth]{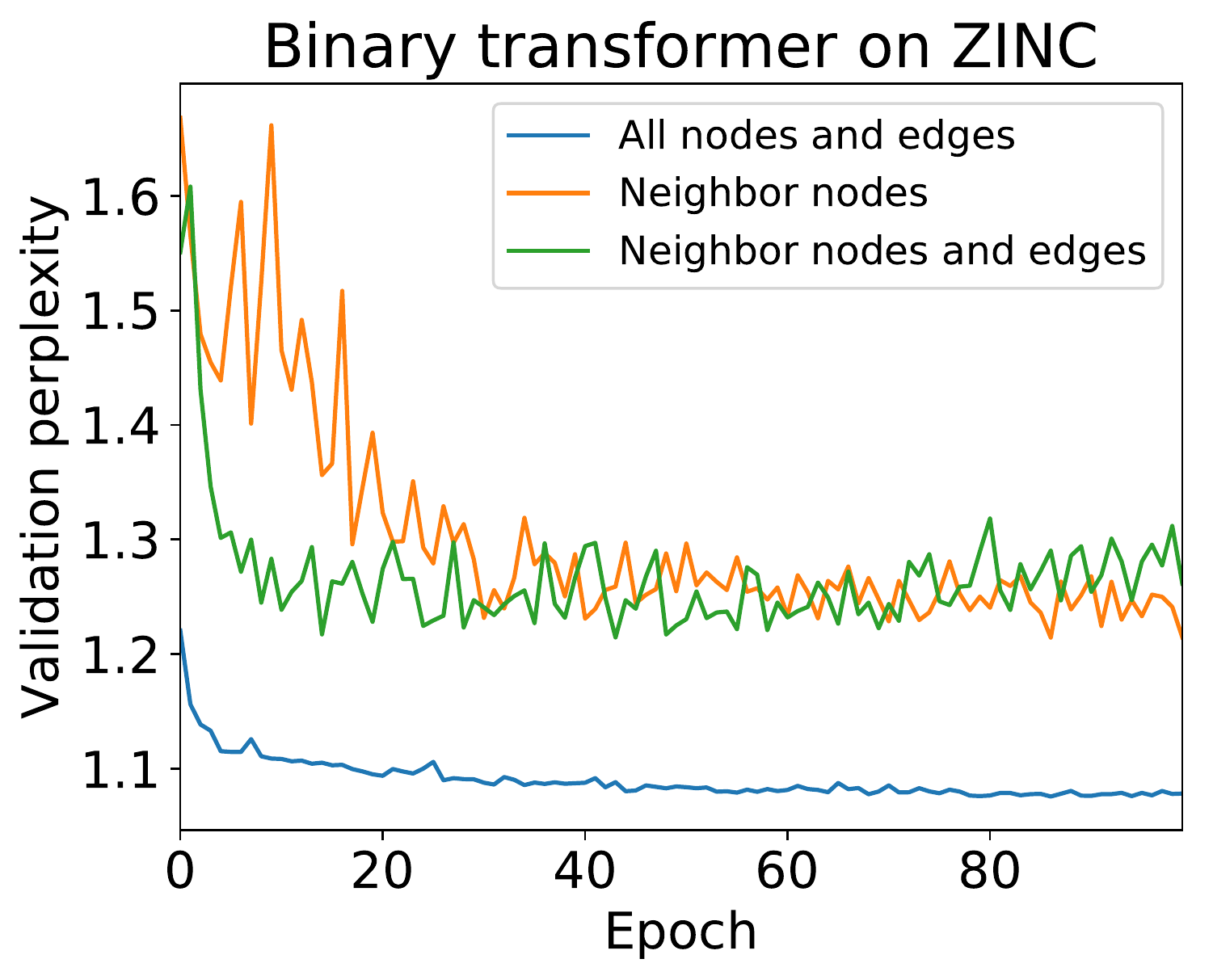}
    \end{subfigure}
    \caption{Perplexity on validation dataset – with one atom masked per molecule – for each epoch of training. (a) is a bond transformer trained on QM9, (b) is a bond transformer trained on ZINC, (c) is a binary transformer trained on QM9 and (d) is a binary transformer trained on ZINC }
    \label{fig:attentions}
\end{figure}

\section{QM9 extended results}
\label{sec:appendix:qm9}

\begin{figure}[H]
    \centering
    \begin{subfigure}[b]{0.45\textwidth}
    \centering
    \includegraphics[width=\textwidth]{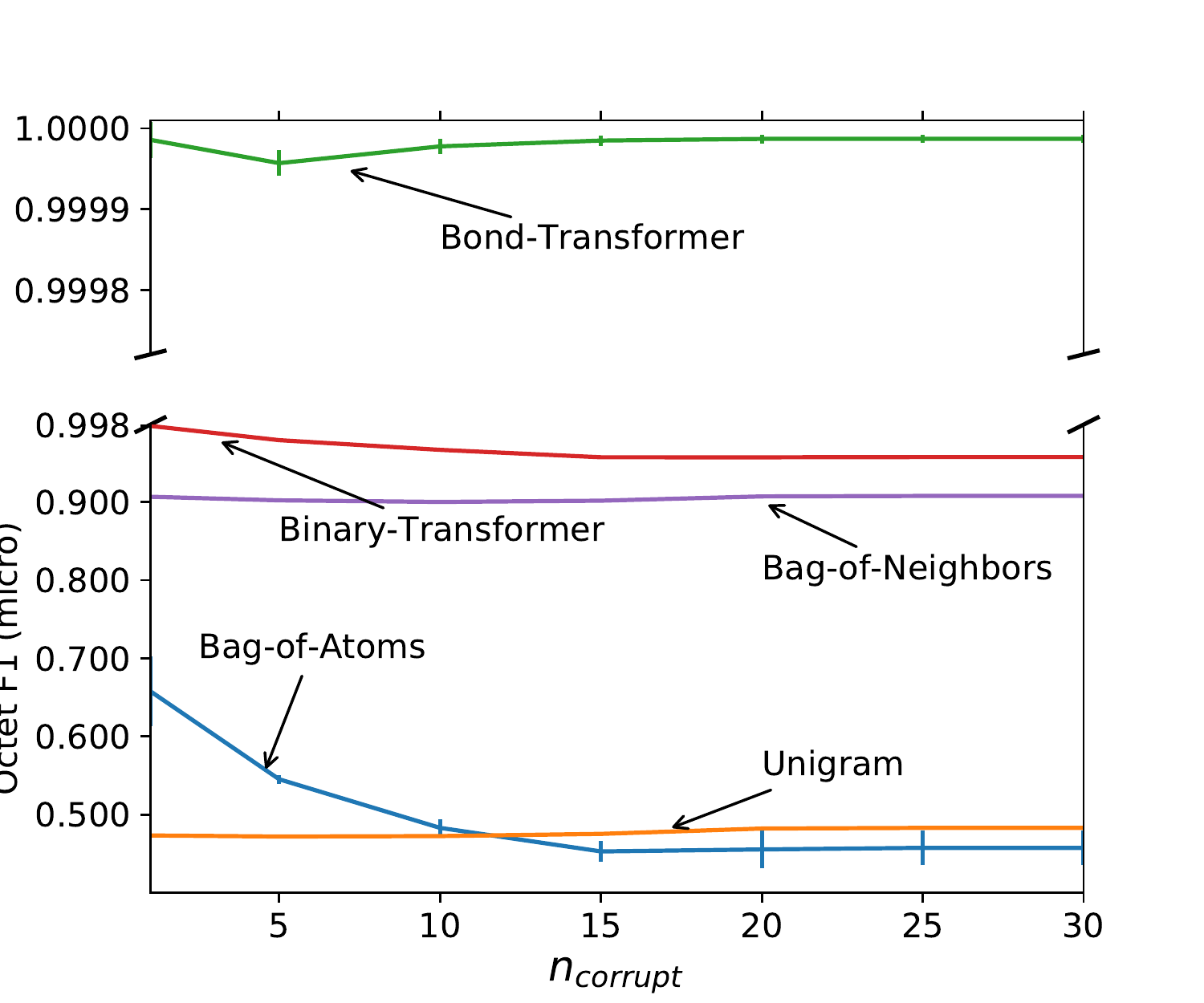}
    \caption{}
    \label{fig:f1_micro_vs_nmask_qm9}
    \end{subfigure}
    \hfill
    \begin{subfigure}[b]{0.45\textwidth}
    \centering
    \includegraphics[width=\textwidth]{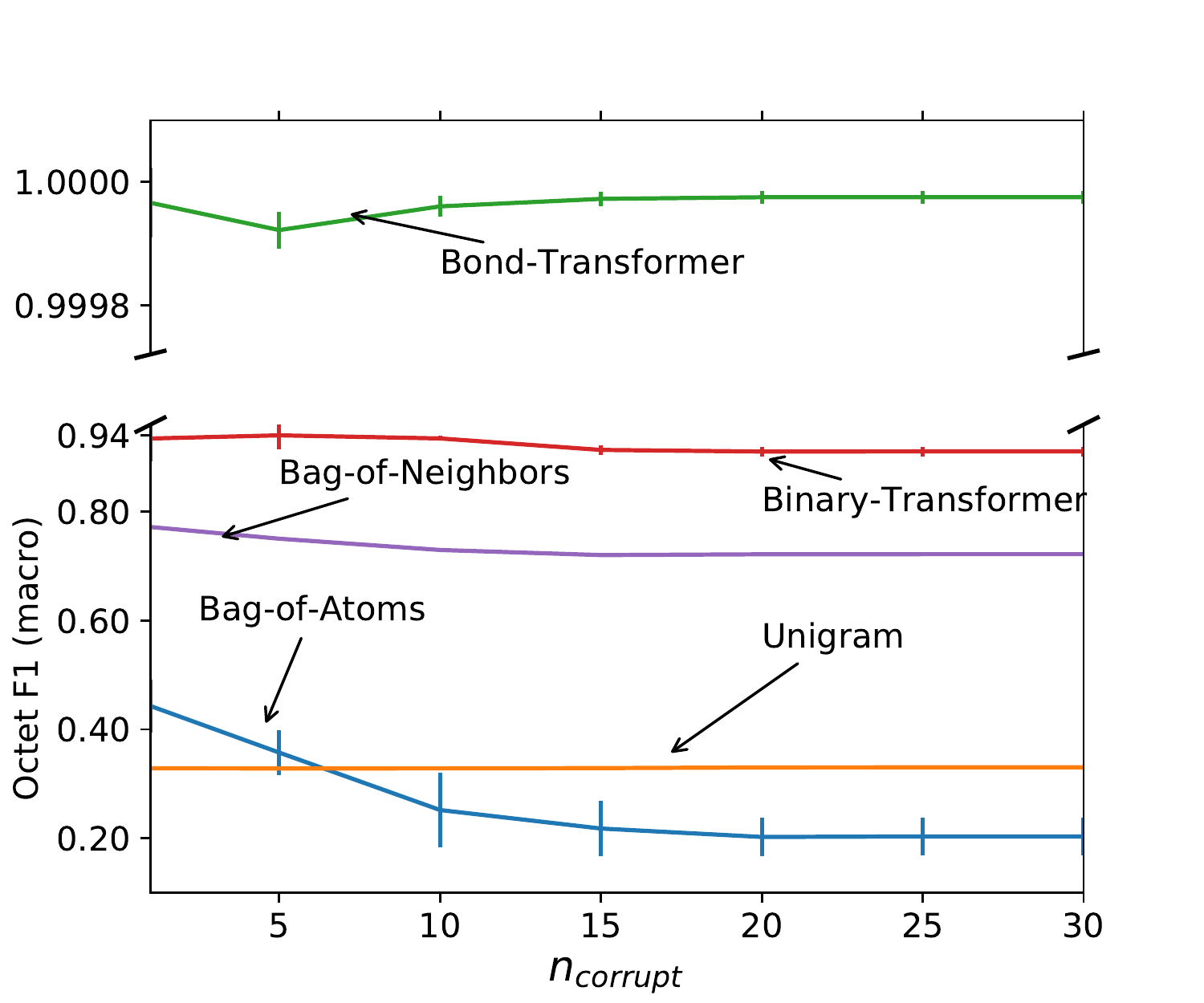}
    \caption{}
    \label{fig:f1_macro_vs_nmask_qm9}
    \end{subfigure}
     \centering
    \begin{subfigure}[b]{0.45\textwidth}
    \centering
    \includegraphics[width=\textwidth]{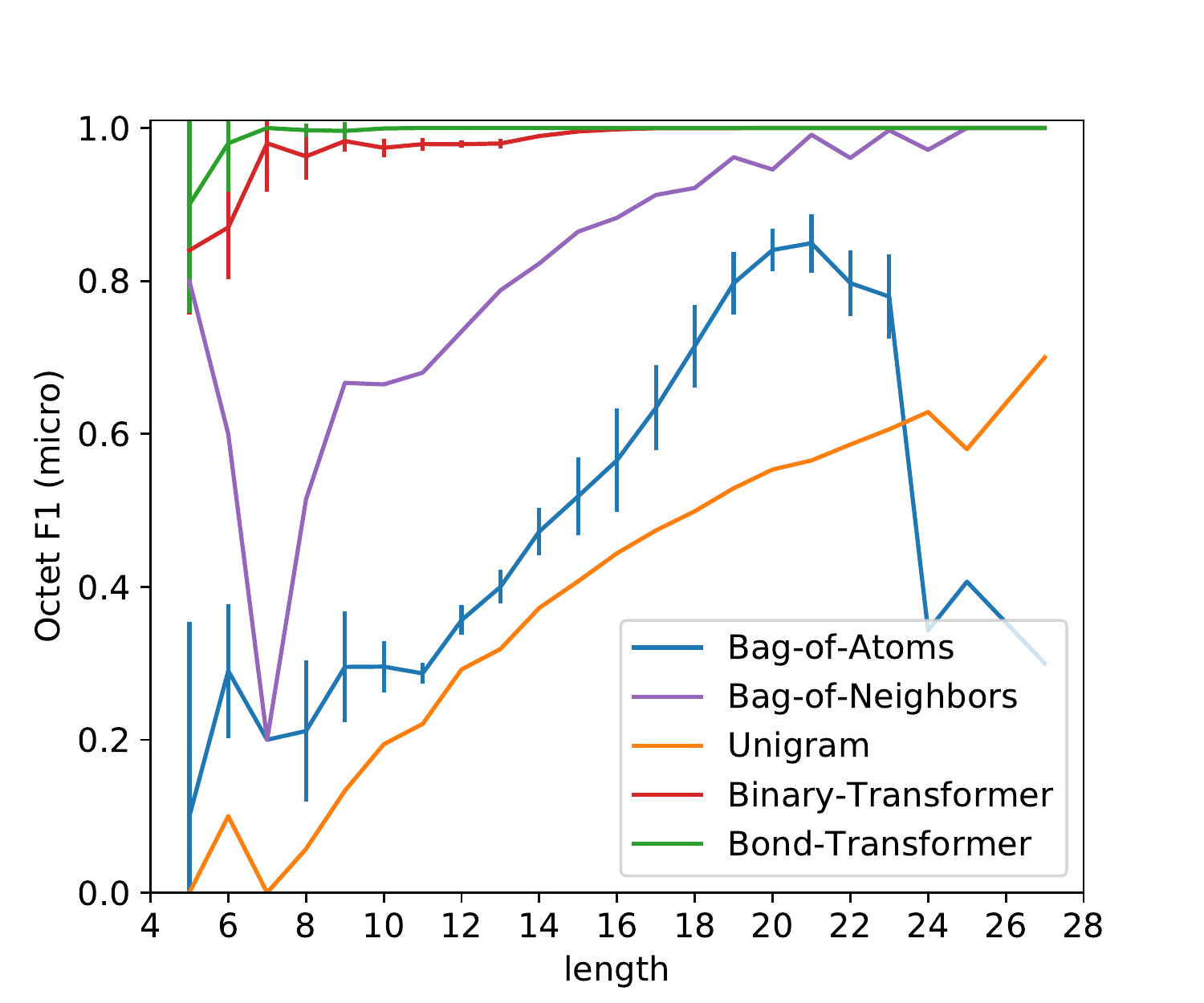}
    \caption{}
    \label{fig:f1_micro_vs_length_qm9}
    \end{subfigure}
    \hfill
    \begin{subfigure}[b]{0.45\textwidth}
    \centering
    \includegraphics[width=\textwidth]{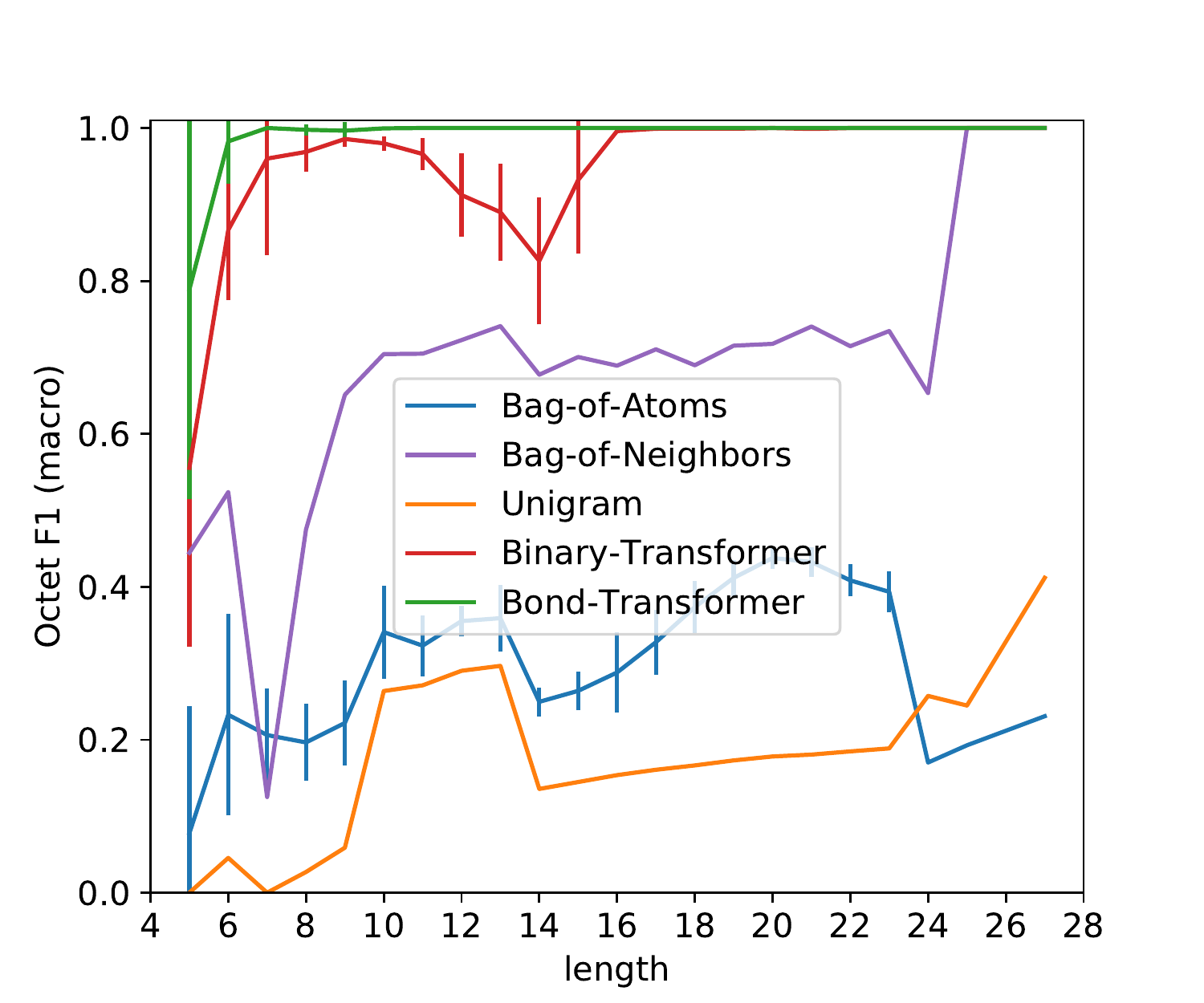}
    \caption{}
    \label{fig:f1_macro_vs_length_qm9}
    \end{subfigure}
    \caption{Octet F1 micro (a) and octet F1 macro (b) evaluated by different number of masked atoms. Octet F1 micro (c) and Octet F1 macro (d) evaluated on molecules of varying size, with 1 atom masked. Error bar corresponds to standard deviation of 10 models trained with different start seed}
    \label{fig:qm9_extended_results}
\end{figure}

From Table \ref{tab:n_mask} and Figure \ref{fig:f1_micro_vs_nmask_qm9},\ref{fig:f1_macro_vs_nmask_qm9} we see that as we mask more atoms per molecule, the \texttt{bond-transformer}, maintains a perfect score, since it can solve the task by only looking at the bonds. The \texttt{Binary-transformer} drops slightly in performance, as we mask more atoms. The \texttt{Bag-of-neighbors} doesn't seem to depend on the number of masked atoms. This indicates that the model most likely, base its predictions on the number of neighbors, which also can be an indication of the number of covalent bonds. As we remove information except compositional, the \texttt{bag-of-atoms} model drops significantly has we mask more atoms, reaching similar performance to the \texttt{Unigram}, as we approach fully masked molecules. This is no surprise, as a fully masked molecule, only gives the model information about the number of atoms, which should not be enough to infer anything.

\begin{table}[H]
\begin{tabular}{|l|l|l|l|l|}
\hline
Model & Metric & $n_{mask}=1$
& $n_{mask}=5$ & all masked\\ \hline
\multirow{3}{*}{\texttt{octet-rule-unigram}}
 & acc & 100  & 100 & 100\\ 
 & f1  & 100  & 100 & 100\\
 & PP  & 1.002 & 1.002 & 1.002 \\ \hline
\multirow{3}{*}{\texttt{bond-transformer}}
 & acc & \textbf{99.99} $\pm$ 0.01 & 99.99 $\pm$ 0.01 & 100.0 $\pm$ 0.0  \\
 & f1  & \textbf{99.99}    $\pm$ 0.01 & 99.99 $\pm$ 0.01 & 100.0 $\pm$ 0.0   \\ 
 & PP  & \textbf{1.002}  $\pm$ 0.001 & 1.002 $\pm$ 0.001  & 1.002 $\pm$ 0.001   \\ \hline
\multirow{3}{*}{\texttt{binary-transformer}}
 & acc & 99.73 $\pm$ 0.06 & 97.91 $\pm$ 0.08 & 95.75 $\pm$ 0.19\\
 & F1  & 99.73 $\pm$ 0.06 & 97.91 $\pm$ 0.08 &  95.75 $\pm$ 0.19\\
 & PP  & 1.009 $\pm$ 0.002 & 1.045 $\pm$ 0.002 & 1.094 $\pm$ 0.004\\ \hline
\multirow{3}{*}{\texttt{bag-of-neighbors}}
 & acc & 90.7 $\pm$ 0.1 & 90.2 $\pm$ 0.1 & 90.8 $\pm$ 0.1 \\ 
 & F1  & 90.7 $\pm$ 0.1 & 90.2 $\pm$ 0.1 & 90.8 $\pm$ 0.1\\
 & PP  & 1.281 $\pm$ 0.004 & 1.299 $\pm$ 0.003 & 1.319 $\pm$ 0.007  \\ \hline
\multirow{3}{*}{\texttt{bag-of-atoms}}
 & acc & 65.8 $\pm$ 4.5 & 54.5 $\pm$ 0.6 & 45.7 $\pm$ 2.2\\ 
 & F1  & 65.8 $\pm$ 4.5 & 54.5 $\pm$ 0.6 & 45.7 $\pm$ 2.2\\
 & PP  & 3.310 $\pm$ 0.478 & 2.895 $\pm$ 0.014 & 2.990 $\pm$ 0.010\\ \hline
\multirow{3}{*}{\texttt{Unigram}}
 & acc & 47.3 & 47.2 & 48.3 \\ 
 & F1  & 47.3  & 47.2 & 48.3\\
 & PP  & 3.104 & 3.113 & 3.038\\ \hline
\end{tabular}
\caption{Performance of our models for 1, 5 and 30 masked atoms per molecule. acc is octet rule accuracy, $F1$ is octet rule F1-micro score and PP is the sample perplexity, each are averaged over the test set. The uncertainty corresponds to the standard deviation of ten models, trained with different start seed.}
\label{tab:n_mask}
\end{table}

The transformer model is very flexible in terms of modeling capability, like any other deep learning model, so to gauge complexity of the task, we evaluate five \texttt{binary-transformer} models of various sizes, which can be seen in Table \ref{tab:transformer_models}.
Here we see that even very small transformer models perform well. As the models increase in number of parameters the performance increases, which however comes at a cost of computation and memory consumption.\\

\begin{table}[H]
\begin{tabular}{|l|l|l|l|l|l|}
\hline
Model & Metric & $n_{mask}=1$ & $n_{mask}=5$ & $t_{train}$ (min) & Parameters\\ \hline

\multirow{3}{*}{\texttt{layers=1, heads=1, $d_{emb}$=4}}
 & acc & 86.0 & 85.8 & \multirow{3}{*}{60} & \multirow{3}{*}{199} \\ 
 & F1  & 86.0  & 85.8 &&\\
 & PP  & 1.426 & 1.441 && \\ \hline
 
 \multirow{3}{*}{\texttt{layers=2, heads=1, $d_{emb}$=4}}
 & acc & 89.9 & 89.8 & \multirow{3}{*}{63} & \multirow{3}{*}{265} \\ 
 & F1  & 89.9  & 89.8 && \\
 & PP  & 1.261 & 1.272 && \\ \hline
 
\multirow{3}{*}{\texttt{layers=2, heads=3, $d_{emb}$=64}}
 & acc & 96.3 & 94.4 & \multirow{3}{*}{77} & \multirow{3}{*}{118149} \\ 
 & F1  & 96.3  & 94.4 &&\\
 & PP  & 1.089 & 1.130 && \\ \hline
 
\multirow{3}{*}{\texttt{layers=4, heads=3, $d_{emb}$=64}}
 & acc & 98.4 & 97.4 & \multirow{3}{*}{82} & \multirow{3}{*}{234885} \\ 
 & F1  & 98.4  & 97.3 && \\
 & PP  & 1.031 & 1.056 && \\ \hline
 
 \multirow{3}{*}{\texttt{layers=8, heads=6, $d_{emb}$=64}}
 & acc & 99.8 & 97.9 & \multirow{3}{*}{110} & \multirow{3}{*}{866181} \\ 
 & F1  & 99.8  & 97.9 && \\
 & PP  & 1.008 & 1.045 && \\ \hline

\end{tabular}
\caption{Performance of \texttt{binary-transformer} models with different number of trainable parameters, for 1 and 5 masked atoms per molecule. $acc$ is octet accuracy, $F1$ is octet F1-score and $PP$ is perplexity, each averaged over the test set. $t_{train}$ is the training time.}
\label{tab:transformer_models}
\end{table}

\section{Zinc extended results}
\label{sec:appendix:zinc}

\begin{figure}[H]
    \centering
    \includegraphics[width=\textwidth]{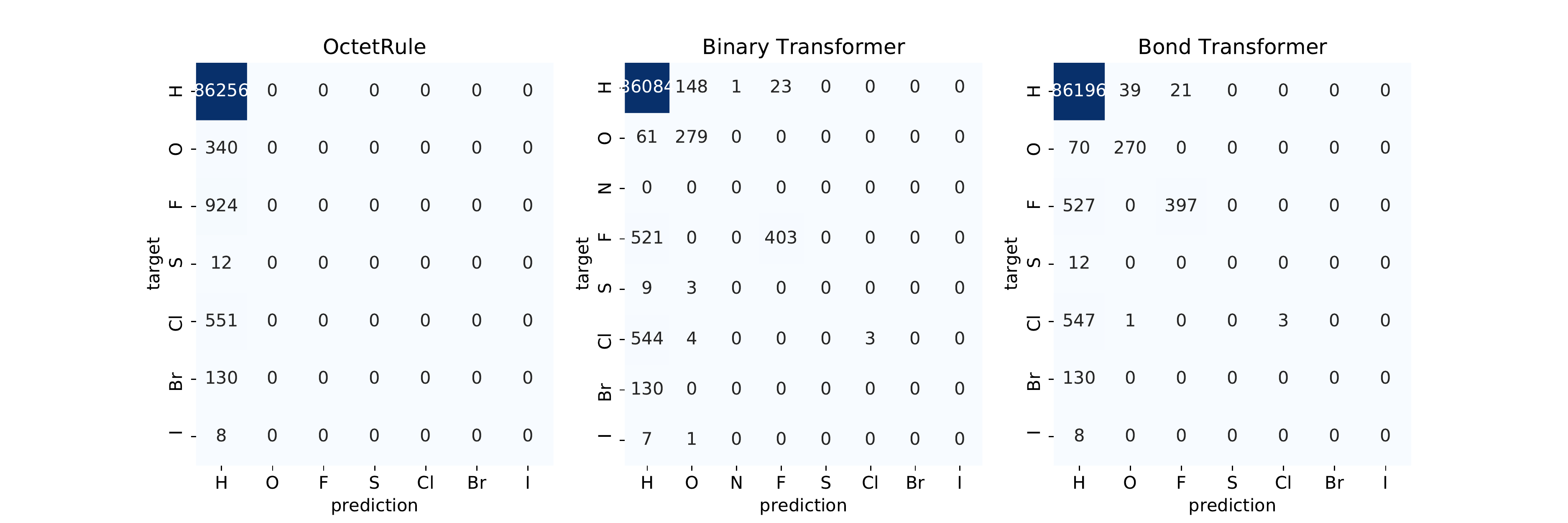}
    \caption{Confusion matrix for cases where the masked atom has one covalent bond.}
    \label{fig:confusion_zinc_1}
\end{figure}

From Figure \ref{fig:confusion_zinc_1} we see that both our transformer models, has learn to discriminate between certain elements, that under the octet-rule should be indistinguishable, like F, but also to allow for ions, in the form of $\text{O}^-$.

\begin{figure}[H]
    \centering
    \includegraphics[width=\textwidth]{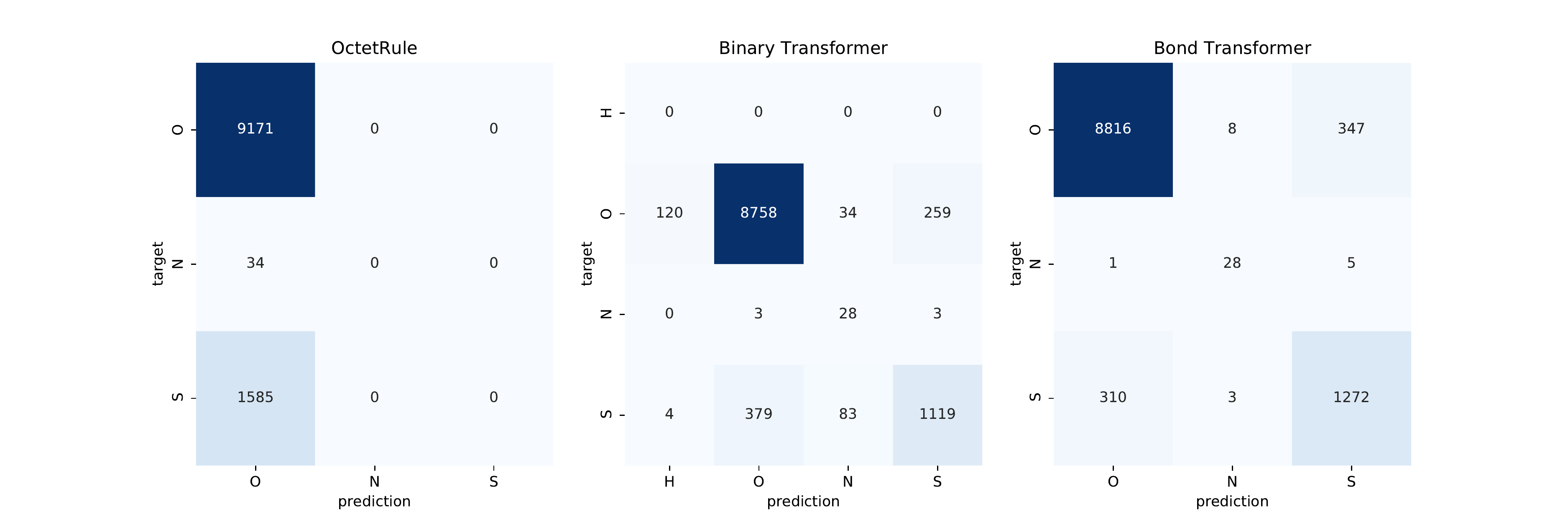}
    \caption{Confusion matrix for cases where the masked atom has two covalent bond.}
    \label{fig:confusion_zinc_2}
\end{figure}

A similar story can be seen in Figure \ref{fig:confusion_zinc_2}, where we have ambiguity between O,S but also $\text{N}^-$ ions.

\begin{figure}[H]
    \centering
    \includegraphics[width=\textwidth]{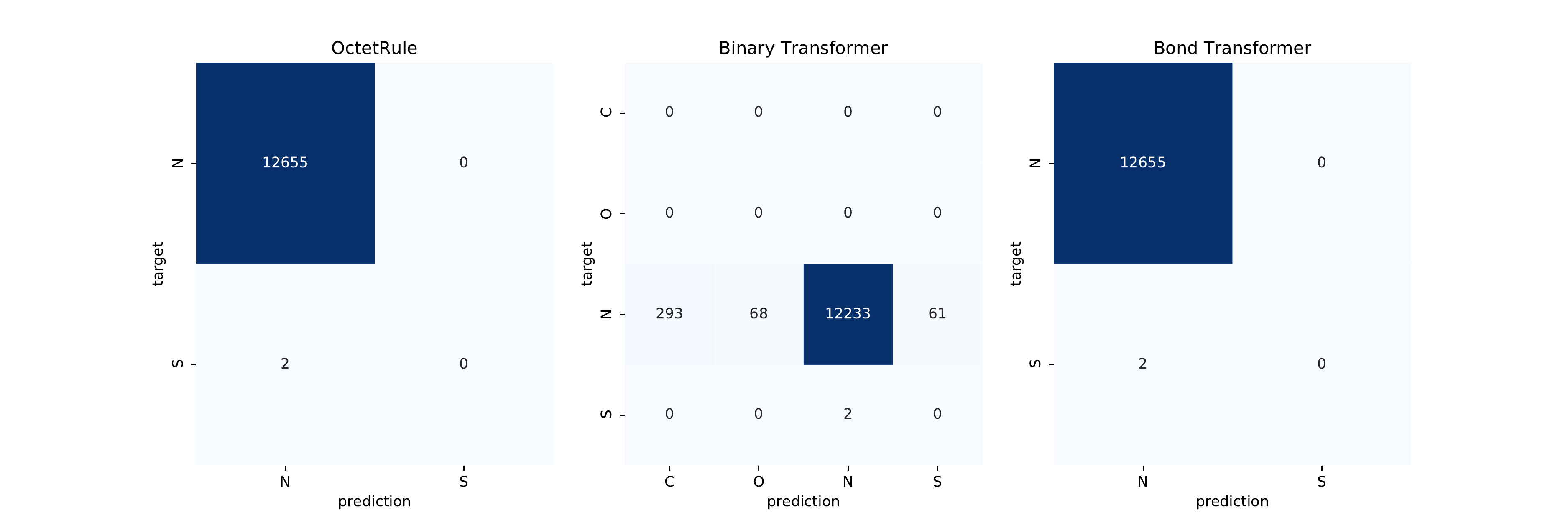}
    \caption{Confusion matrix for cases where the masked atom has three covalent bond.}
    \label{fig:confusion_zinc_3}
\end{figure}

\begin{figure}[H]
    \centering
    \includegraphics[width=\textwidth]{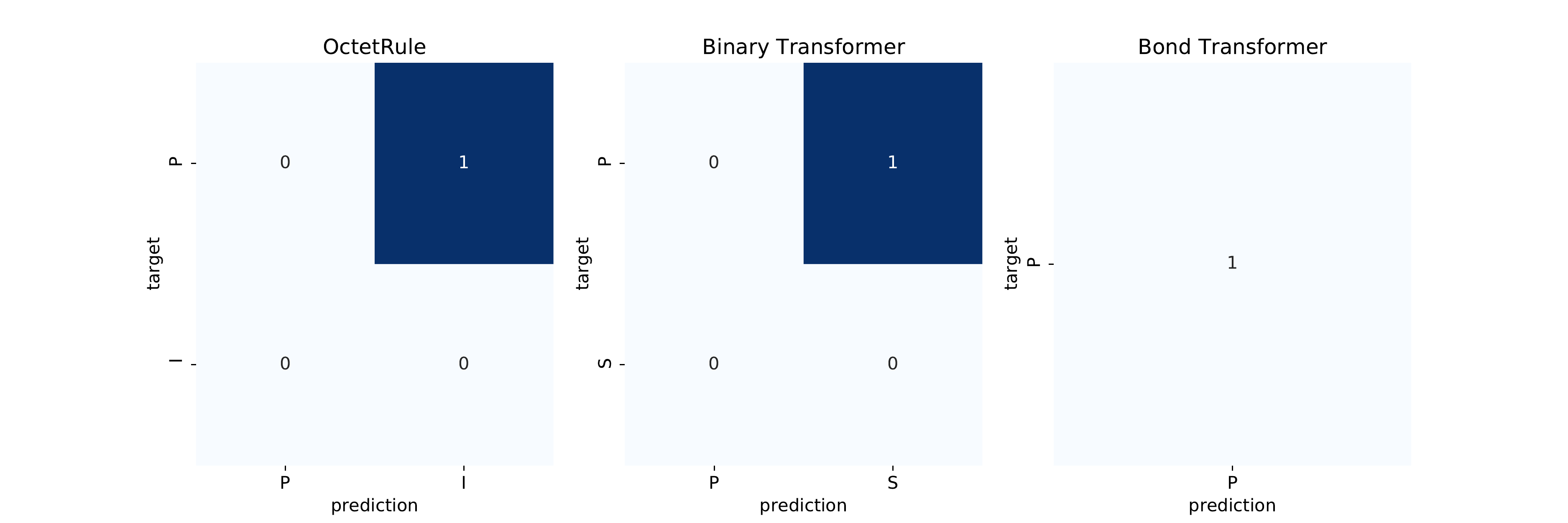}
    \caption{Confusion matrix for cases where the masked atom has five covalent bond.}
    \label{fig:confusion_zinc_5}
\end{figure}

\begin{figure}[H]
    \centering
    \includegraphics[width=\textwidth]{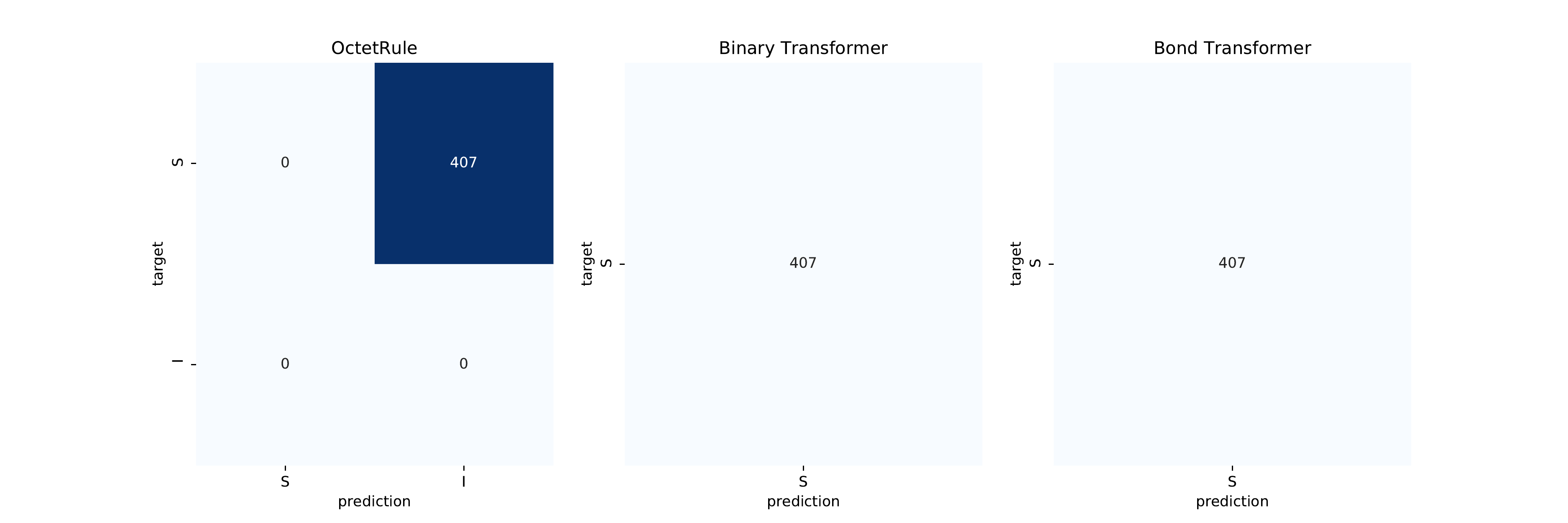}
    \caption{Confusion matrix for cases where the masked atom has six covalent bond.}
    \label{fig:confusion_zinc_6}
\end{figure}

Figure \ref{fig:confusion_zinc_3},\ref{fig:confusion_zinc_5} and \ref{fig:confusion_zinc_6} does not provide any insights, as the dataset is too bias, and almost only contain one type of element for each number of covalent bonds.

\begin{figure}[H]
    \centering
    \includegraphics[width=\textwidth]{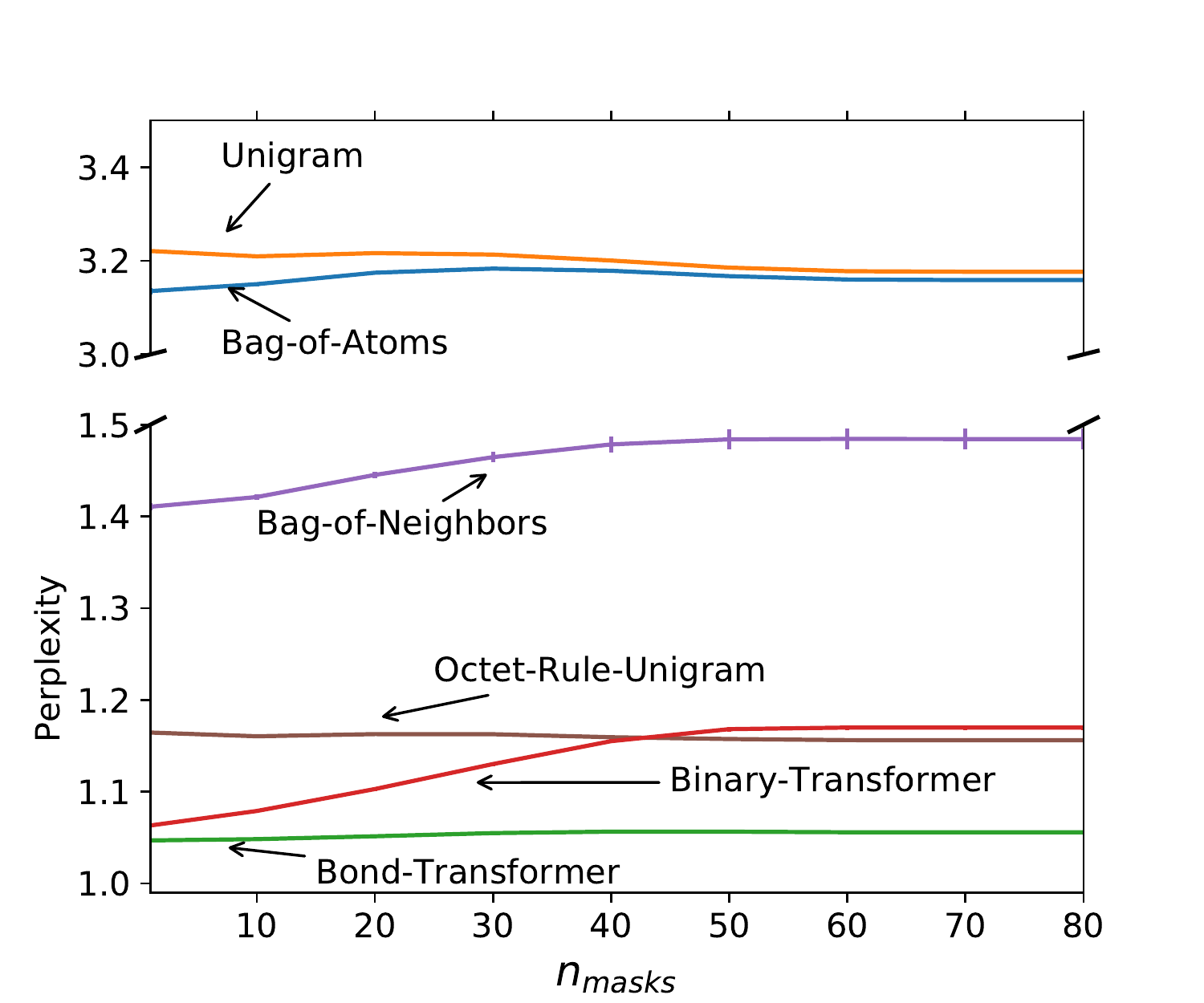}
    \caption{Sample perplexity evaluated by different number of masked atoms. Error bar corresponds to standard deviation of 10 models trained with different start seed}
    \label{fig:perplexity_vs_masks_zinc}
\end{figure}

\begin{figure}[H]
    \centering
    \begin{subfigure}[b]{0.45\textwidth}
    \centering
    \includegraphics[width=\textwidth]{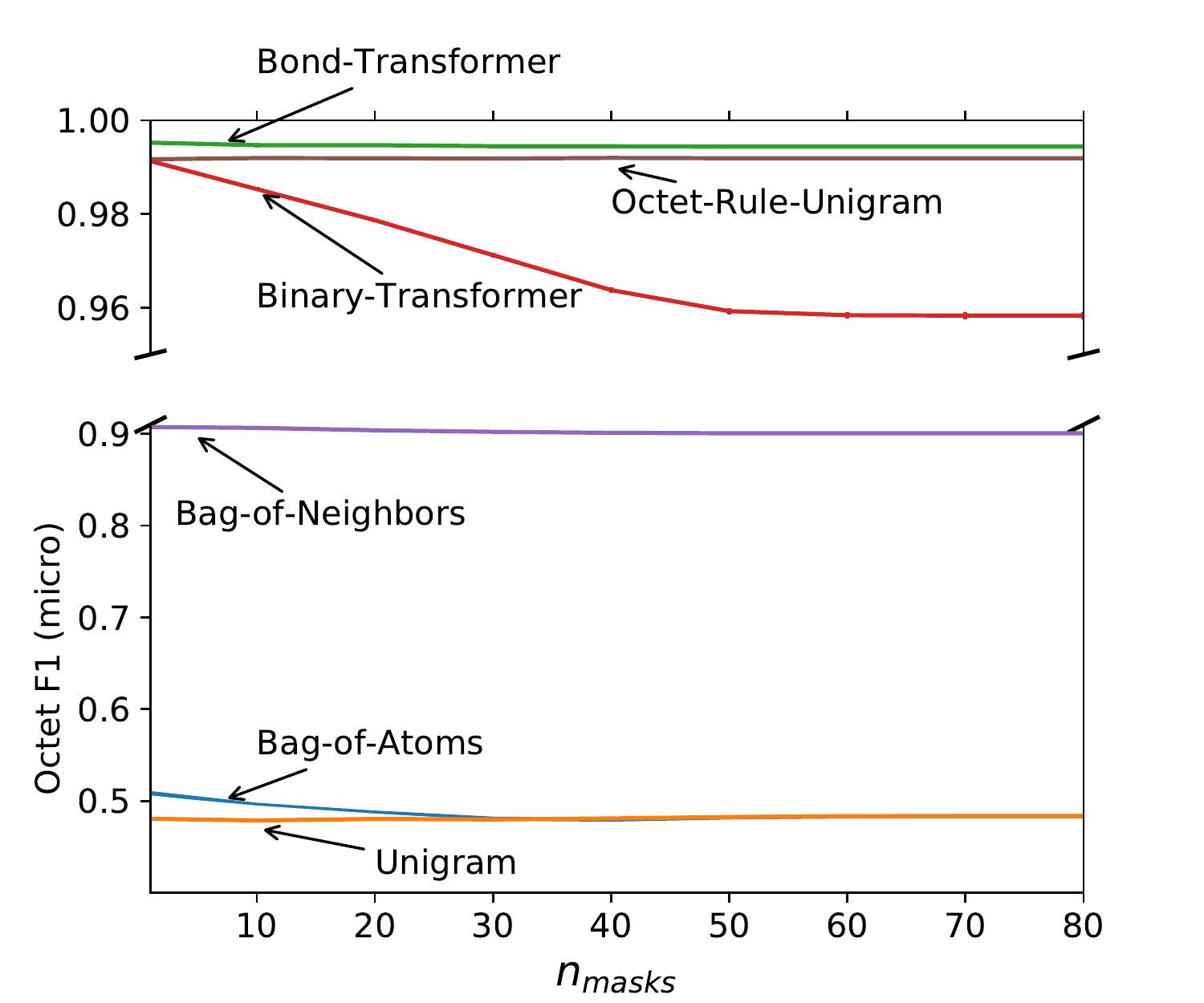}
    \caption{}
    \label{fig:octet_f1_micro_vs_nmask}
    \end{subfigure}
    \hfill
    \begin{subfigure}[b]{0.45\textwidth}
    \centering
    \includegraphics[width=\textwidth]{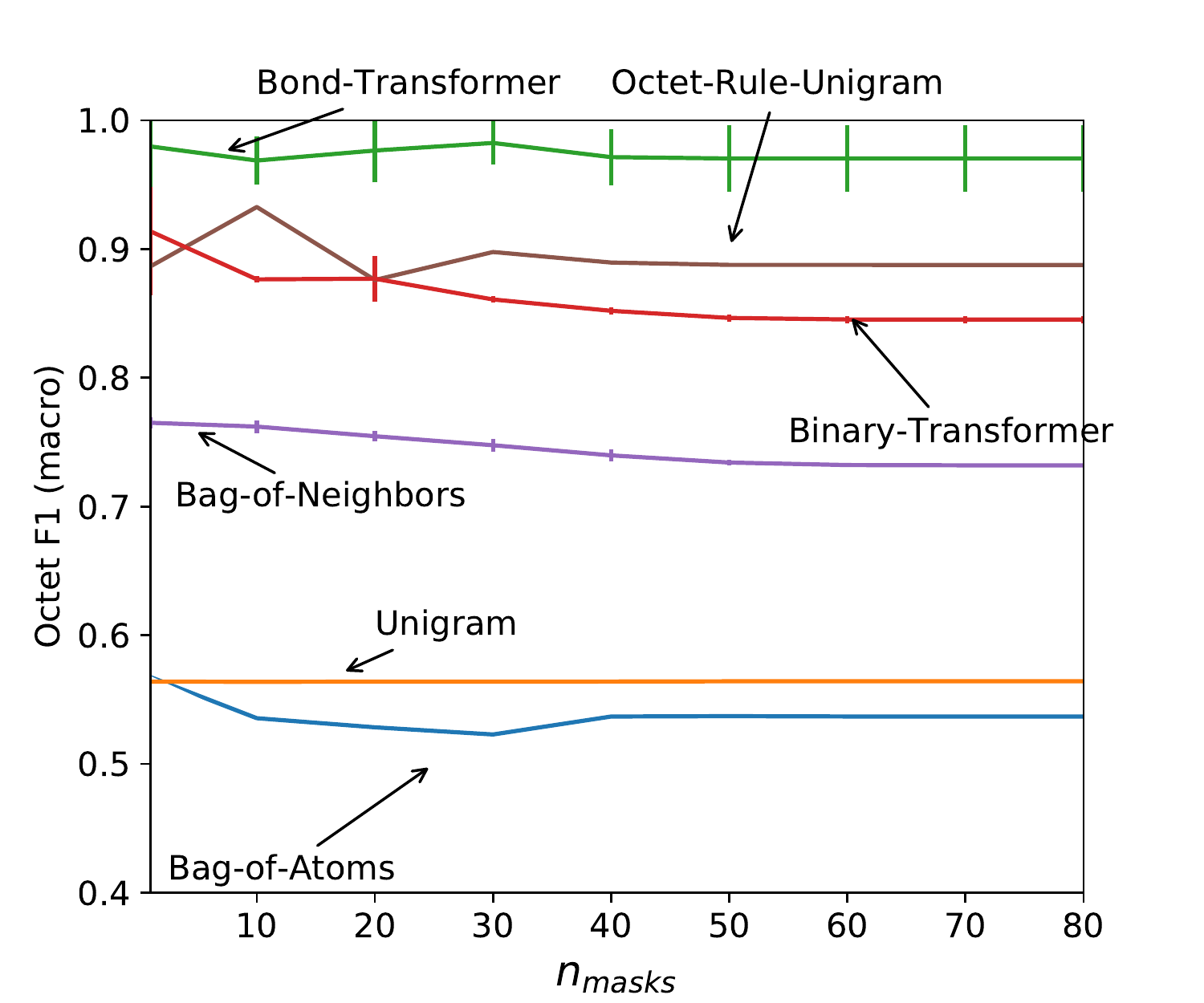}
    \caption{}
    \label{fig:octet_f1_macro_vs_nmask}
    \end{subfigure}
     \centering
    \begin{subfigure}[b]{0.45\textwidth}
    \centering
    \includegraphics[width=\textwidth]{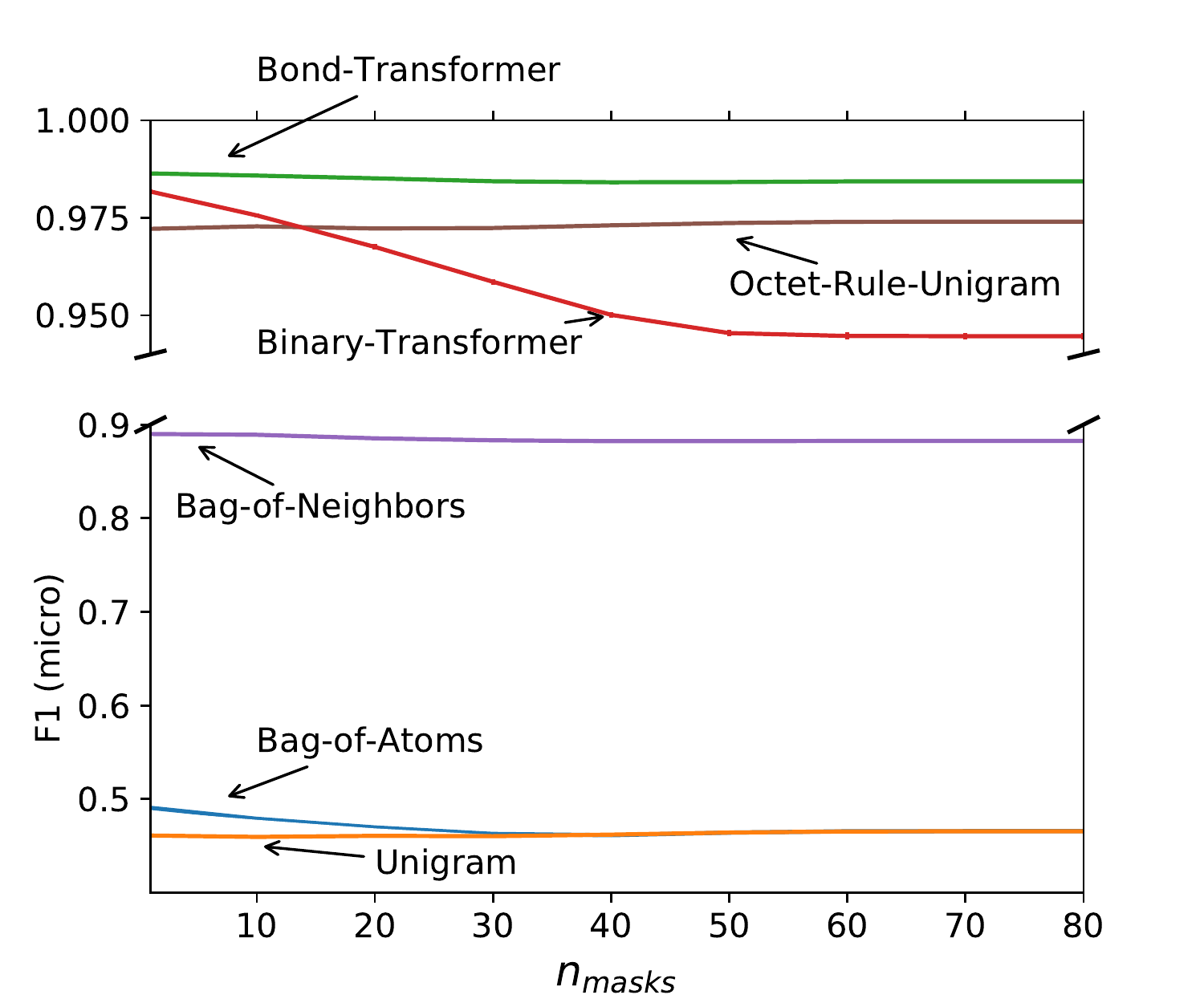}
    \caption{}
    \label{fig:f1_micro_vs_nmask}
    \end{subfigure}
    \hfill
    \begin{subfigure}[b]{0.45\textwidth}
    \centering
    \includegraphics[width=\textwidth]{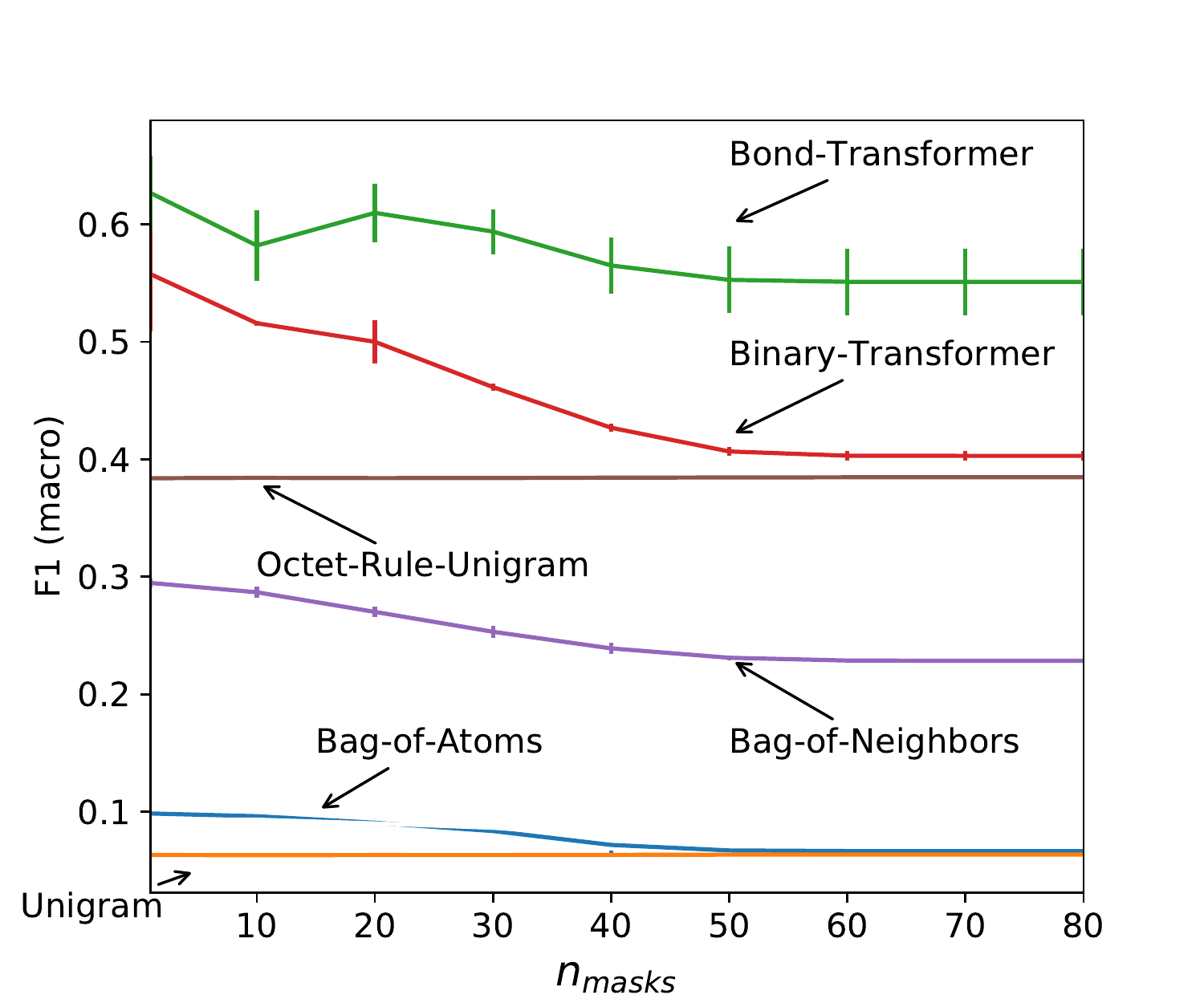}
    \caption{}
    \label{fig:f1_macro_vs_nmask}
    \end{subfigure}
    
    \caption{Octet F1 micro (a), octet F1 macro (b), sample F1 micro (c) and sample F1 macro (d) evaluated by different number of masked atoms. Error bar corresponds to standard deviation of 10 models trained with different start seed}
    \label{fig:metrics_vs_nmask_zinc}
\end{figure}

Figure \ref{fig:perplexity_vs_masks_zinc},\ref{fig:metrics_vs_nmask_zinc} we see the save story as underlined in the main text, namely that the \texttt{Bond-transformer} outperforms and the \texttt{Binary-transformer} also performs similar or better than the \texttt{octet-rule-unigram} model, depending on the number of masks, and metric used to evaluate.

\begin{figure}[H]
    \centering
    \begin{subfigure}[b]{0.45\textwidth}
    \centering
    \includegraphics[width=\textwidth]{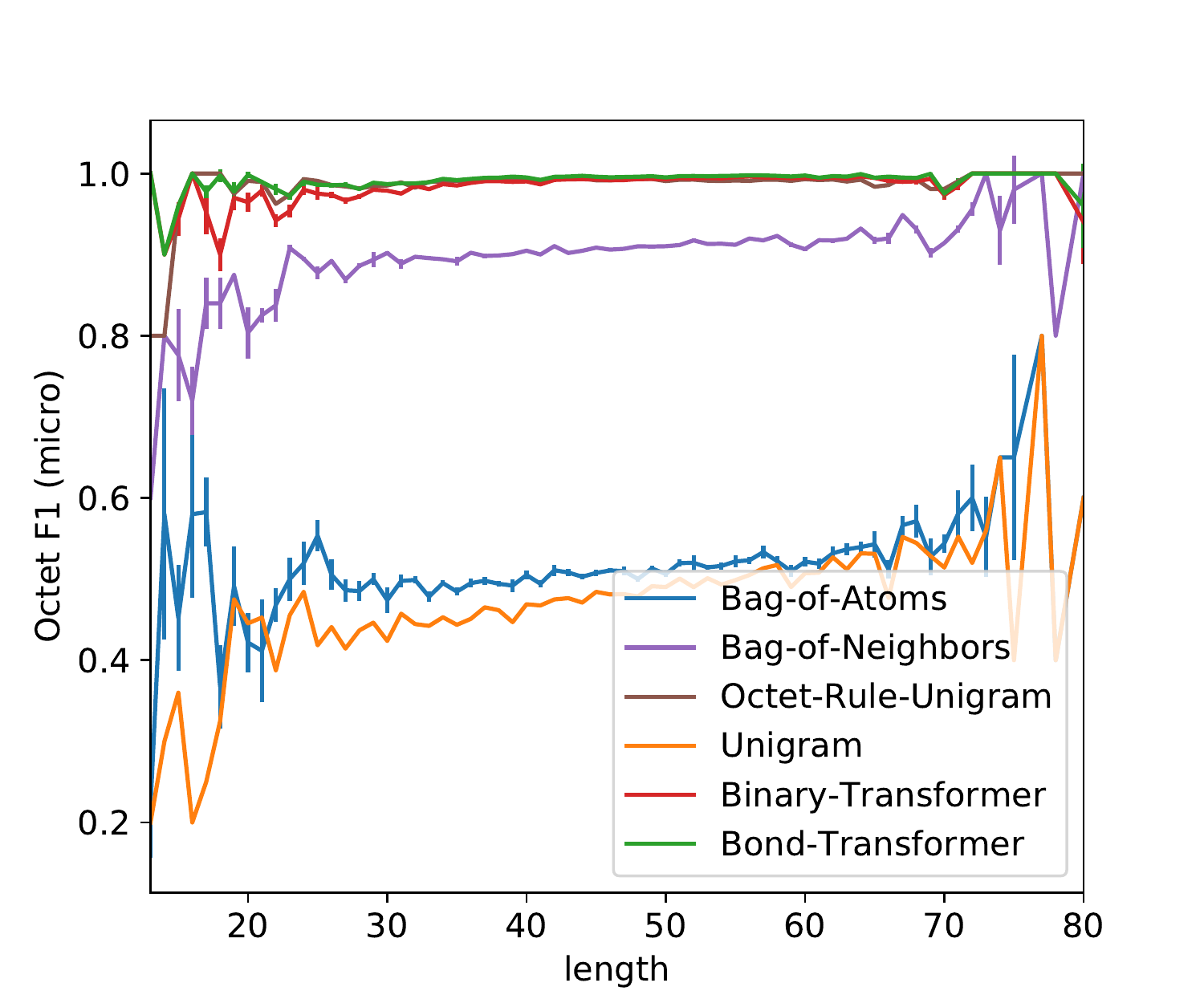}
    \caption{}
    \label{fig:octet_f1_micro_vs_length}
    \end{subfigure}
    \hfill
    \begin{subfigure}[b]{0.45\textwidth}
    \centering
    \includegraphics[width=\textwidth]{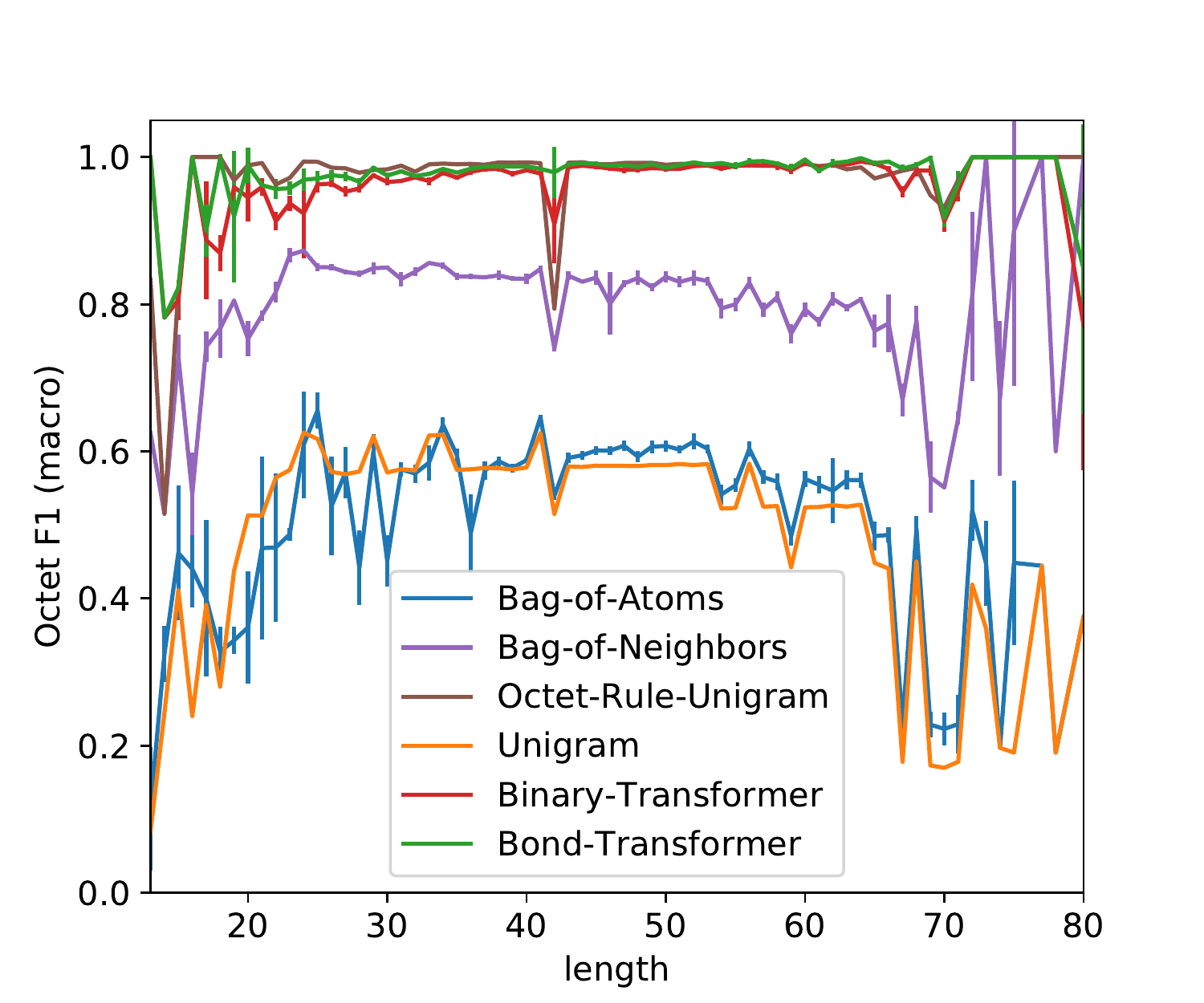}
    \caption{}
    \label{fig:octet_f1_macro_vs_length}
    \end{subfigure}
     \centering
    \begin{subfigure}[b]{0.45\textwidth}
    \centering
    \includegraphics[width=\textwidth]{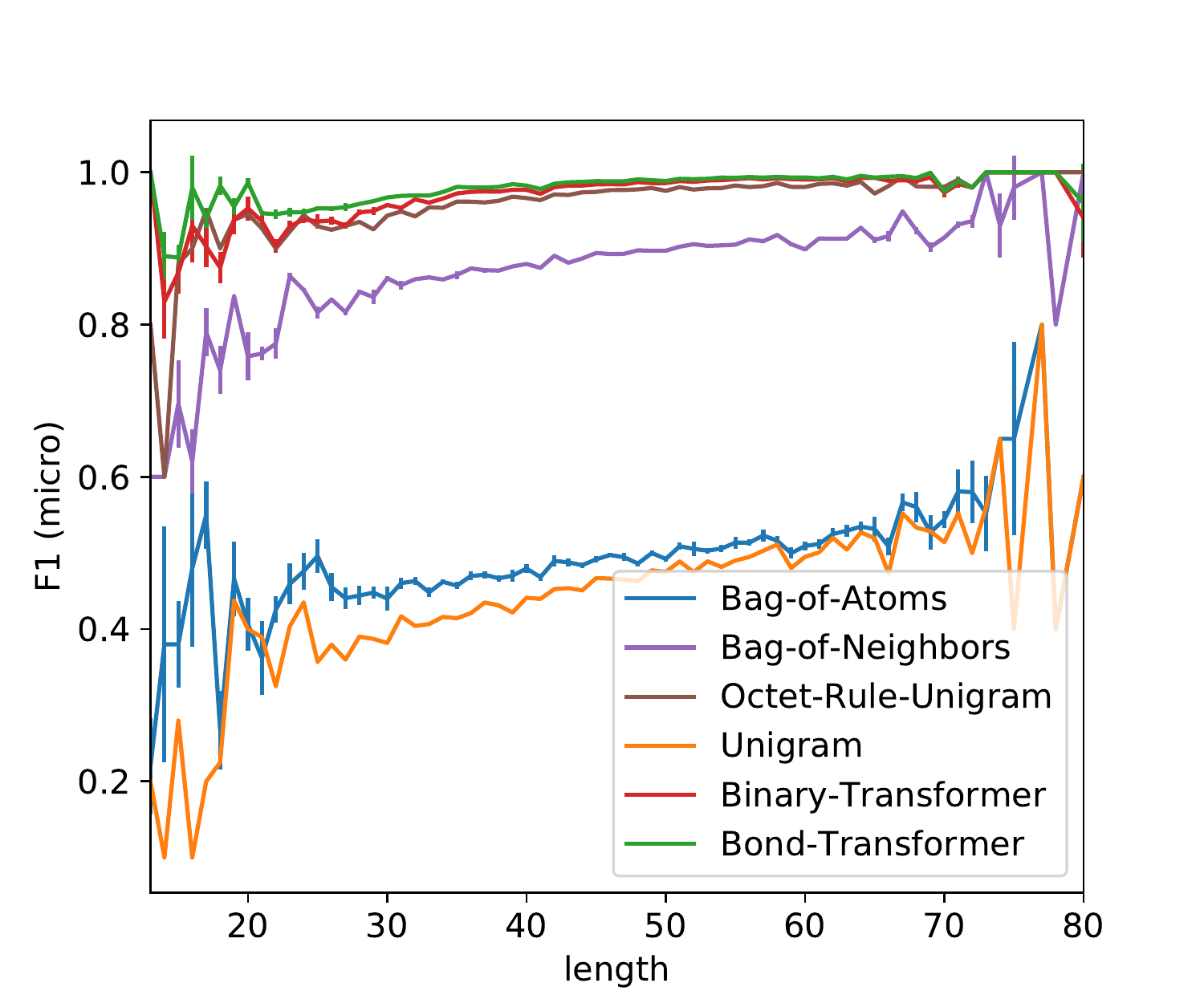}
    \caption{}
    \label{fig:f1_micro_vs_length}
    \end{subfigure}
    \hfill
    \begin{subfigure}[b]{0.45\textwidth}
    \centering
    \includegraphics[width=\textwidth]{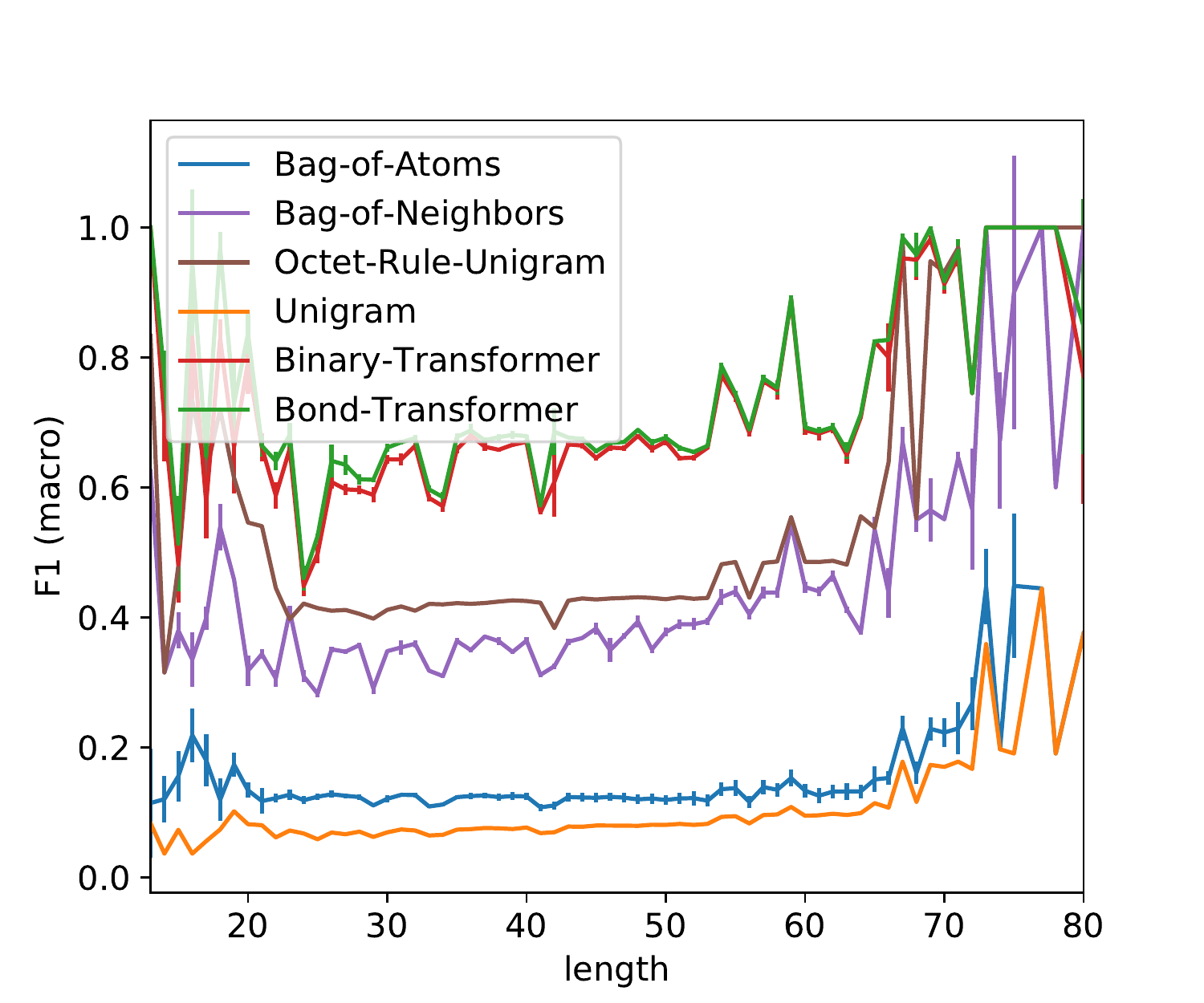}
    \caption{}
    \label{fig:f1_macro_vs_length}
    \end{subfigure}
    \caption{Octet F1 micro (a), octet F1 macro (b), sample F1 micro (c) and sample F1 macro (d) evaluated on molecules of varying size, with 1 atom masked. Error bar corresponds to standard deviation of 10 models trained with different start seed}
\end{figure}

\end{document}